\documentclass[12pt]{article}
\usepackage[round]{natbib}
\usepackage{amsmath}
\usepackage{amssymb}
\usepackage{bbm}
\usepackage{amsthm}
\usepackage{dsfont}
\usepackage[margin=1in]{geometry}
\usepackage{hyperref}
\usepackage{multirow}
\usepackage{color}
\usepackage{tikz}
\usepackage{graphicx}
\usepackage{ulem}
\usepackage{algorithm2e}
\usepackage{float}
\usepackage{lscape}
\usepackage{rotating}
\usepackage{longtable}
\usepackage{subcaption}
\usepackage{etoc}
\usepackage{authblk}
\usepackage{url}
\usepackage{enumerate}

\hypersetup{
    colorlinks=true,
    linkcolor=black,
    filecolor=magenta,
    urlcolor=blue,
    citecolor=black,
}
\usepackage{times}
\def\ChLogical{1}
\def\ChScoring{2}
\def\ChGAM{3}
\def\ChCaseBased{4}
\def\ChSupDis{5}
\def\ChUnsupDis{6}
\def\ChDR{7}
\def\ChPhysics{8}
\def\ChRashomon{9}
\def\ChRL{10}
\def\x{\mathbf{x}}

 \definecolor{dark-green}{rgb}{0.17,0.47,.22}
\title{Interpretable Machine Learning: Fundamental Principles and 10 Grand Challenges
}
\author{Cynthia Rudin$^1$, Chaofan Chen$^2$, Zhi Chen$^1$, Haiyang Huang$^1$, Lesia Semenova$^1$, and Chudi Zhong$^{1}$
\\
$^1$ Duke University\\
$^2$ University of Maine\\}
\date{July, 2021}

\usepackage{natbib}
\usepackage{graphicx}

\begin{document}

\maketitle

\begin{abstract}
    Interpretability in machine learning (ML) is crucial for high stakes decisions and troubleshooting. In this work, we provide fundamental principles for interpretable ML, and dispel common misunderstandings that dilute the importance of this crucial topic. We also identify 10 technical challenge areas in interpretable machine learning and provide history and background on each problem. Some of these problems are classically important, and some are recent problems that have arisen in the last few years. These problems are: (\ChLogical) Optimizing sparse logical models such as decision trees; (\ChScoring) Optimization of scoring systems; (\ChGAM) Placing constraints into generalized additive models to encourage sparsity and better interpretability; (\ChCaseBased) Modern case-based reasoning, including neural networks and matching for causal inference;  (\ChSupDis) Complete supervised disentanglement of neural networks; (\ChUnsupDis) Complete or even partial unsupervised  disentanglement of neural networks; (\ChDR) Dimensionality reduction for data visualization; (\ChPhysics) Machine learning models that can incorporate physics and other generative or causal constraints; (\ChRashomon) Characterization of the ``Rashomon set'' of good models; and (\ChRL) Interpretable reinforcement learning. This survey is suitable as a starting point for statisticians and computer scientists interested in working in interpretable machine learning.\footnote{Equal contribution from C. Chen, Z. Chen, H. Huang, L. Semenova, and C. Zhong}
\end{abstract}

\section*{Introduction}
With widespread use of machine learning (ML), the importance of interpretability has become clear in avoiding catastrophic consequences. Black box predictive models, which by definition are inscrutable, have led to serious societal problems that deeply affect health, freedom, racial bias, and safety. Interpretable predictive models, which are constrained so that their reasoning processes are more understandable to humans, are much easier to troubleshoot and to use in practice. It is universally agreed that interpretability is a key element of trust for AI models \citep{Wagstaff12, RudinWa14,LoPiano20,ashoori2019ai,Thiebes20,Spiegelhalter2020Should,brundage2020trustworthy}. In this survey, we provide fundamental principles, as well as 10 technical challenges in the design of inherently interpretable machine learning models.

Let us provide some background.
A black box machine learning model is a formula that is either too complicated for any human to understand, or proprietary, so that one cannot understand its inner workings. Black box models are difficult to troubleshoot, which is particularly problematic for medical data. Black box models often predict the right answer for the wrong reason (the ``Clever Hans'' phenomenon), leading to excellent performance in training but poor performance in practice \citep{Schramowski20,Lapuschkin19,Oconnor21,Zech2018,BadgeleyEtAl2019,Hamamoto20}.  
There are numerous other issues with black box models.  In criminal justice, individuals may have been subjected to years of extra prison time due to typographical errors in black box model inputs \citep{Wexler17} and poorly-designed proprietary models for air quality have had serious consequences for public safety during wildfires \citep{McGough2018}; both of these situations may have been easy to avoid with interpretable models. In cases where the underlying distribution of data changes (called domain shift, which occurs often in practice), problems arise if users cannot troubleshoot the model in real-time, which is much harder with black box models than interpretable models. Determining whether a black box model is fair with respect to gender or racial groups is much more difficult than determining whether an interpretable model has such a bias. In medicine, black box models turn computer-aided decisions into automated decisions, precisely because physicians cannot understand the reasoning processes of black box models. Explaining black boxes, rather than replacing them with interpretable models, can make the problem worse by providing misleading or false characterizations \citep{Rudin19,LaugelEtAl19,LakkarajuBa20}, or adding unnecessary authority to the black box \citep{RudinRadin2019}. There is a clear need for innovative machine learning models that are inherently interpretable.

There is now a vast and confusing literature on some combination of interpretability and explainability. Much literature on explainability confounds it with interpretability/comprehensibility, thus obscuring the arguments (and thus detracting from their precision), and failing to convey the relative importance and use-cases of the two topics in practice. Some of the literature discusses topics in such generality that its lessons have little bearing on any specific problem. Some of it aims to design taxonomies that miss vast topics within interpretable ML. Some of it provides definitions that we disagree with. Some of it even provides guidance that could perpetuate bad practice. Importantly, most of it assumes that one would explain a black box without consideration of whether there is an interpretable model of the same accuracy. In what follows, we provide some simple and general guiding principles of interpretable machine learning. These are not meant to be exhaustive. Instead they aim to help readers avoid common but problematic ways of thinking about interpretability in machine learning. 
%For instance, people commonly think that interpretability creates trust. It does not, it enables decisions about trust. Many works reference ``the accuracy/interpretability tradeoff'' that is assumed to exist. However, there is no scientific evidence for such a statement in reference to machine learning; there is no evidence that an interpretable model \textit{can't} exist that is as accurate as a black box. An important argument against interpretable machine learning is that it is impossible to define interpretability and therefore, it is not possible to create an interpretable model. This defeatist argument is identical to one that claims there is no clear single measure of predictive performance, therefore we should not do machine learning at all; there must thus be a flaw in this argument. We have heard many times that an explained black box is as good as an inherently interpretable model. These statements are all problematic for important reasons.

The major part of this survey outlines a set of important and fundamental technical grand challenges in interpretable machine learning. These are both modern and classical challenges, and some are much harder than others. They are all either hard to solve, or difficult to formulate correctly. While there are numerous sociotechnical challenges about model deployment (that can be much more difficult than technical challenges), human-computer-interaction challenges, and how robustness and fairness interact with interpretability, those topics can be saved for another day. 
%With that background, we proceed to  the challenges. 
%\section*{The Challenges}
We begin with the most classical and most canonical problems in interpretable machine learning: how to build sparse models for tabular data, including decision trees (Challenge \#\ChLogical) and scoring systems (Challenge \#\ChScoring). %\ref{sec:sparse}
 We then delve into a challenge involving additive models (Challenge  \#\ChGAM),
%\ref{sec:gam}
 followed by another in case-based reasoning (Challenge \#\ChCaseBased),
% \ref{sec:case}
 which is another classic topic in interpretable artificial intelligence. We then move to more exotic problems, namely supervised and unsupervised disentanglement of concepts in neural networks (Challenges \#\ChSupDis{} and  \#\ChUnsupDis).
 Back to classical problems, we discuss dimension reduction (Challenge \#\ChDR).
 Then, how to incorporate physics or causal constraints (Challenge \#\ChPhysics).
 Challenge \#\ChRashomon{} involves understanding, exploring, and measuring the Rashomon set of accurate predictive models. Challenge \#\ChRL{} discusses interpretable reinforcement learning.  Table \ref{tab:applicationstable} provides a guideline that may help users to match a dataset to a suitable interpretable supervised learning technique. We will touch on all of these techniques in the challenges.
\begin{table}[th]
    \centering
    %\begin{tabular}{|c|c|}\hline
    \begin{tabular}{|p{0.3\textwidth}|p{0.65\textwidth}|}\hline
    \textbf{Models} & \textbf{Data type}   \\\hline
      decision trees / decision lists (rule lists) / decision sets   & somewhat clean tabular data with interactions, including multiclass problems. Particularly useful for categorical data with complex interactions (i.e., more than quadratic). \\\hline
   scoring systems      &  somewhat clean tabular data, typically used in medicine and criminal justice because they are small enough that they can be memorized by humans.  \\\hline
      generalized additive models (GAMs)   & continuous data with at most quadratic interactions, useful for large-scale medical record data.   \\\hline
      case-based reasoning & any data type (different methods exist for different data types), including multiclass problems. \\\hline
     disentangled neural networks & data with raw inputs (computer vision, time series, textual data), suitable for multiclass problems.\\
     \hline
    \end{tabular}
    \caption{Rule of thumb for the types of data that naturally apply to various supervised learning algorithms. ``Clean'' means that the data do not have too much noise or systematic bias. ``Tabular'' means that the features are categorical or real, and that each feature is a meaningful predictor of the output on its own. ``Raw'' data is unprocessed and has a complex data type, e.g., image data where each pixel is a feature,  medical records, or time series data. }
    \label{tab:applicationstable}
\end{table}

\subsection*{General Principles of Interpretable Machine Learning}

Our first fundamental principle defines interpretable ML, following \citet{Rudin19}:\\

%\vspace*{5pt}
\noindent\textbf{Principle 1} \textit{An interpretable machine learning model obeys a domain-specific set of constraints to allow it (or its predictions, or the data) to be more easily understood by humans. These constraints can differ dramatically depending on the domain.} \\

A typical interpretable supervised learning setup, with data $\{z_i\}_i$, and models chosen from function class $\mathcal{F}$ is:
\begin{equation}\label{eqn:generic}
    \min_{f\in\mathcal{F}} \frac{1}{n}\sum_i\textrm{Loss}(f,z_i) + C\cdot \textrm{InterpretabilityPenalty}(f),\; \textrm{ subject to }\; \textrm{InterpretabilityConstraint}(f),
\end{equation}
where the loss function, as well as soft and hard interpretability constraints, are chosen to match the domain. (For classification $z_i$ might be $(x_i,y_i)$, $x_i\in \mathbb{R}^p, y_i\in\{-1,1\}$.) 
%When these constraints are obeyed, the resulting model $f$ is more likely to be interpretable, 
The goal of these constraints is to make the resulting model $f$ or its predictions more interpretable. While solutions of \eqref{eqn:generic} would not necessarily be sufficiently interpretable to use in practice, the constraints would generally help us find models that would be interpretable (if we design them well), and we might also be willing to consider slightly suboptimal solutions to find a more useful model. The constant $C$ trades off between accuracy and the interpretability penalty, and can be tuned, either by cross-validation or by taking into account the user's desired tradeoff between the two terms. 
%Since not all machine learning problems are supervised, adaptations of \eqref{eqn:generic} can be used, and we will show this for many of the challenges we introduce here. 

Equation \eqref{eqn:generic} can be generalized to unsupervised learning, where the loss term is simply replaced with a loss term for the unsupervised problem, whether it is novelty detection, clustering, dimension reduction, or another task.

Creating interpretable models can sometimes be much more difficult than creating black box models for many different reasons including: (i) Solving the optimization problem may be computationally hard, depending on the choice of constraints and the model class $\mathcal{F}$. (ii) When one does create an interpretable model, one invariably realizes that the data are problematic and require troubleshooting, which slows down deployment (but leads to a better model). 
(iii) It might not be initially clear which definition of interpretability to use. This definition might require refinement, sometimes over multiple iterations with domain experts. There are many papers detailing these issues, the earliest dating from the mid-1990s \citep[e.g.,][]{kodratoff1994comprehensibility}.

%\vspace*{5pt}
%Each domain is different. 
Interpretability differs across domains just as predictive performance metrics vary across domains. Just as we might choose from a variety of performance metrics (e.g., accuracy, weighted accuracy, precision, average precision, precision@N, recall, recall@N, DCG, NCDG, AUC, partial AUC, mean-time-to-failure, etc.), or combinations of these metrics, we might also choose from a combination of interpretability metrics that are specific to the domain. We may not be able to define a single best definition of interpretability; regardless, if our chosen interpretability measure is helpful for the problem at hand, we are better off including it. 
Interpretability penalties or constraints can include sparsity of the model, monotonicity with respect to a variable, decomposibility into sub-models, an ability to perform case-based reasoning or other types of visual comparisons, disentanglement of certain types of information within the model's reasoning process, generative constraints (e.g., laws of physics), preferences among the choice of variables, or any other type of constraint that is relevant to the domain. Just as it would be futile to create a complete list of performance metrics for machine learning, any list of interpretability metrics would be similarly fated.

Our 10 challenges involve how to define some of these interpretability constraints and how to incorporate them into machine learning models. For tabular data, sparsity is usually part of the definition of interpretability, whereas for computer vision of natural images, it generally is not. (Would you find a model for natural image classification interpretable if it uses only a few pixels from the image?) For natural images, we are better off with interpretable neural networks that perform case-based reasoning or disentanglement, and provide us with a visual understanding of intermediate computations; we will describe these in depth. Choices of model form (e.g., the choice to use a decision tree, or a specific neural architecture) are examples of interpretability constraints. For most problems involving tabular data, a fully interpretable model, whose full calculations can be understood by humans such as a sparse decision tree or sparse linear model, is generally more desirable than either a model whose calculations can only be partially understood, or a model whose predictions (but not its model) can be understood. Thus, we make a distinction between \textit{fully interpretable} and \textit{partially interpretable} models, often preferring the former.
%; partially interpretable models, where constraints can help us understand important elements of a model or its predictions, can also be very helpful, for instance, for computer vision problems that use neural networks. 
%; in this survey, we cover both these fully interpretable models as well as partially interpretable models that have less stringent constraints.

Interpretable machine learning models are not needed for all machine learning problems. For low-stakes decisions (e.g., advertising), for decisions where an explanation would be trivial and the model is 100\% reliable (e.g., ``there is no lesion in this mammogram'' where the explanation would be trivial), for decisions where humans can verify or modify the decision afterwards (e.g., segmentation of the chambers of the heart), interpretability is probably not needed.\footnote{Obviously, this document does not apply to black box formulas that not depend on randomness in the data, i.e., a calculation of deterministic function, not machine learning.} On the other hand, for self-driving cars, even if they are very reliable, problems would arise if the car's vision system malfunctions causing a crash and no reason for the crash is available. Lack of interpretability would be problematic in this case.  
% We note that there are some situations in which interpretability is irrelevant (such as image segmentation of chambers of the heart, where a human can easily verify and adjust the result, or cases where the model is 100\% accurate and the explanation would be trivial, such as ``there is no lesion in this image''), but we are concerned with the more difficult cases here. 

Our second fundamental principle concerns trust:\\

%\vspace*{5pt}
\noindent \textbf{Principle 2} \textit{Despite common rhetoric, interpretable models do not necessarily create or enable trust -- they could also enable {\underline{\rm dis}}trust. They simply allow users to \underline{\textrm{\rm decide}} whether to trust them. In other words, they permit a decision of trust, rather than trust itself.} \\

%\vspace*{5pt}
With black boxes, one needs to make a decision about trust with much less information; without knowledge about the reasoning process of the model, it is much more difficult to detect whether it might generalize beyond the dataset. As stated by \citet{AfnanEtAl2021} with respect to medical decisions, while interpretable AI is an enhancement of human decision making, black box AI is a replacement of it.

\begin{figure}[th]
    \centering
    \includegraphics[scale=.2]{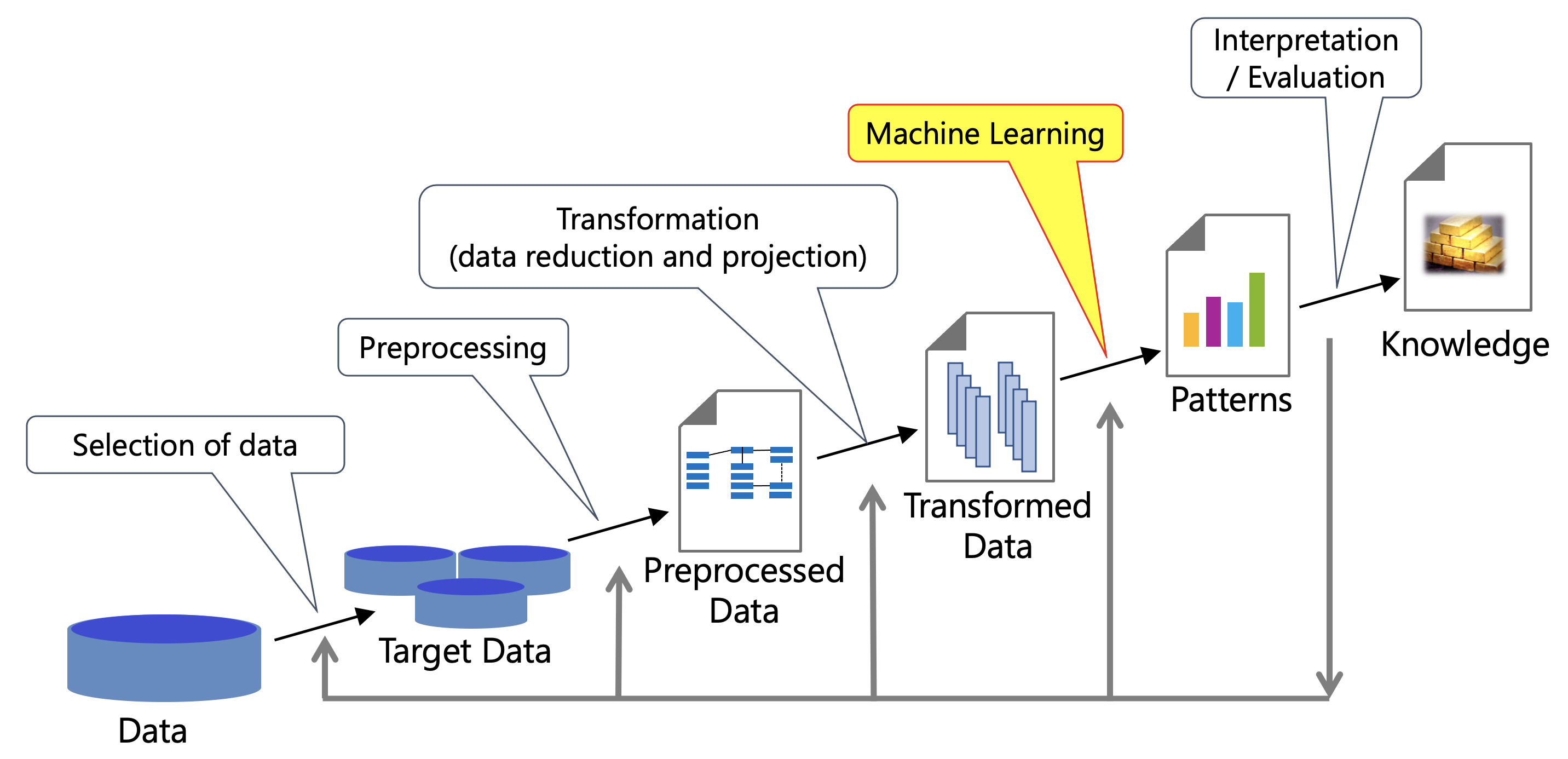}
    \caption{Knowledge Discovery Process \citep[figure adapted from][]{Fayyad96fromdata}}.
    \label{fig:kddprocess}
\end{figure}

An important point about interpretable machine learning models is that \textit{there is no scientific evidence for a general tradeoff between accuracy and interpretability} when one considers the full data science process for turning data into knowledge. (Examples of such pipelines include KDD, CRISP-DM, or the CCC Big Data Pipelines; see Figure \ref{fig:kddprocess}, or \citealt{Fayyad96fromdata, crispdm, bigdata}.) 
In real problems, interpretability is useful for troubleshooting, which leads to better accuracy, not worse. In that sense, we have the third principle:\\

\noindent \textbf{Principle 3} \textit{It is important not to assume that one needs to make a sacrifice in accuracy in order to gain interpretability. In fact, interpretability often begets accuracy, and not the reverse. Interpretability versus accuracy is, in general, a false dichotomy in machine learning.}\\

Interpretability has traditionally been associated with complexity, and specifically, sparsity, but model creators generally would not \textit{equate} interpretability with sparsity. Sparsity is often one component of interpretability, and a model that is sufficiently sparse but has other desirable properties is more typical. While there is almost always a tradeoff of accuracy with sparsity (particularly for extremely small models), there is no evidence of a general tradeoff of accuracy with interpretability. Let us consider both (1) development and use of ML models in practice, and (2) experiments with static datasets; in neither case have interpretable models proven to be less accurate.\\

\noindent\textit{Development and use of ML models in practice.}
A key example of the pitfalls of black box models when used in practice is the story of COMPAS \citep{compas,RudinWaCo2020}. COMPAS is a black box because it is proprietary: no one outside of its designers knows its secret formula for predicting criminal recidivism, yet it is used widely across the United States and influences parole, bail, and sentencing decisions that deeply affect people's lives \citep{propublica2016}. COMPAS is \textit{error-prone} because it could require over 130 variables, and typographical errors in those variables influence outcomes \citep{Wexler17}. COMPAS has been \textit{explained incorrectly}, where the news organization ProPublica mistakenly assumed that an important variable in an approximation to COMPAS (namely race) was also important to COMPAS itself, and used this faulty logic to conclude that COMPAS depends on race other than through age and criminal history \citep{propublica2016,RudinWaCo2020}. While the racial bias of COMPAS does not seem to be what ProPublica claimed, COMPAS still has \textit{unclear dependence on race}. And worst, COMPAS seems to be \textit{unnecessarily complicated} as it does not seem to be any more accurate than a very sparse decision tree \citep{AngelinoLaAlSeRu17-kdd,AngelinoEtAl2018} involving only a couple of variables. It is a key example simultaneously demonstrating many pitfalls of black boxes in practice. And, it shows an example of when black boxes are unnecessary, but used anyway.

There are other systemic issues that arise when using black box models in practice or when developing them as part of a data science process, that would cause them to be less accurate than an interpretable model. A full list is beyond the scope of this particular survey, but an article that points out many serious issues that arise with black box models in a particular domain is that of \citet{AfnanEtAl2021}, who discuss in vitro fertilization (IVF). In modern IVF, black box models that have not been subjected to randomized controlled trials determine who comes into existence. \citet{AfnanEtAl2021} points out ethical and practical issues, ranging from the inability to perform shared decision-making with patients, to economic consequences of clinics needing to ``buy into'' environments that are similar to those where the black box models are trained so as to avoid distribution shift; this is necessary precisely because errors cannot be detected effectively in real-time using the black box. They also discuss accountability issues (``Who is accountable when the model causes harm?''). Finally, they present a case where a standard performance metric (namely area under the ROC curve -- AUC) has been misconstrued as representing the value of a model in practice, potentially leading to overconfidence in the performance of a black box model. Specifically, a reported AUC was inflated by including many ``obvious'' cases in the sample over which it was computed. If we cannot trust reported numerical performance results, then interpretability would be a crucial remaining ingredient to assess trust.  
%They also discuss the possibility of biased selection of embryos leading to a biased selection of humans that come into the world, where such biases would be more difficult to detect with black box models than interpretable ones.

In a full data science process, like the one shown in Figure \ref{fig:kddprocess}, interpretability plays a key role in determining how to update the other steps of the process for the next iteration. One interprets the results and tunes the processing of the data, the loss function, the evaluation metric, or anything else that is relevant, as shown in the figure. How can one do this without understanding how the model works? It may be possible, but might be much more difficult. 
In essence, the messiness that comes with messy data and complicated black box models causes lower quality decision-making in practice. 

Let's move on to a case where the problem is instead controlled, so that we have a static dataset and a fixed evaluation metric.\\

%but we can give an example of an economic consequence
%Let us say a company provides a black box model for a high stakes decision, such as choosing embryos for implantation for in vitro fertilization. Since the logic for individual predictions cannot be checked while the model is deployed, the company that designed the model must ensure that the setting where the model is trained is essentially identical to the setting where it is deployed to prevent issues with domain adaptation or confounding. To ensure the two settings are similar, the deploying organization needs to buy into the setting, which could be expensive (e.g., reagents, medical scanners, etc.), creating an economic disadvantage for the deploying organization
%Other systemic issues can  arise when a black box model gets lost. (For instance, the company that created it folds without preserving the model.) Then, the organizations that used this model have no true way to actually explain how decisions were made in the past, leaving them vulnerable to possible litigation if the model has a long-term impact on people's lives.

%We formally define an interpretable machine learning model as a machine learning model obeying a domain-specific set of constraints that makes it easier to understand \citep{Rudin19}. 

%Thus, we present our fourth principle.\\

\noindent \textit{Static datasets.} Most benchmarking of algorithms is done on static datasets, where the data and evaluation metric are not cleaned or updated as a result of a run of an algorithm. In other words, these experiments are not done as part of a data science process, they are only designed to compare algorithms in a controlled experimental environment. 

Even with static datasets and fixed evaluation metrics, interpretable models do not generally lead to a loss in accuracy over black box models. Even for deep neural networks for computer vision, even on the most challenging of benchmark datasets, a plethora of machine learning techniques have been designed \textit{that do not sacrifice accuracy but gain substantial interpretability} \citep{zhang2018interpretable,chen2019looks,koh2020concept,chen2020concept,angelov2020towards,nauta2020neural}.

%In fact, even for the most challenging benchmark datasets in computer vision, there are now numerous examples of interpretable neural networks that make no sacrifice in accuracy to satisfy interpretability constraints \cite{}. Thus, even in the restricted setting of a static dataset and static evaluation metric, there has been no sacrifice in accuracy to gain interpretability.
%, and we will discuss this in depth in several of our challenges.

%Let us discuss this last point further, as it is the most important of those on this list. As we have discussed, 

Let us consider two extremes of data types: tabular data, where all variables are real or discrete features, each of which is meaningful (e.g., age, race, sex, number of past strokes, congestive heart failure), and ``raw'' data, such as images, sound files or text, where each pixel, bit, or word is not useful on its own. These types of data have different properties with respect to both machine learning performance and interpretability.
 
For tabular data, most machine learning algorithms tend to perform similarly in terms of prediction accuracy. This means it is often difficult even to beat logistic regression, assuming one is willing to perform minor preprocessing such as creating dummy variables \citep[e.g., see][]{christodoulou_systematic_2019}. In these domains, neural networks generally find no advantage. It has been known for a very long time that very simple models perform surprisingly well for tabular data \citep{Holte93}.
The fact that simple models perform well for tabular data could arise from the Rashomon Effect discussed by Leo Breiman \citep{breiman2001statistical}. Breiman posits the possibility of a large Rashomon set, i.e., a multitude of models with approximately the minumum error rate, for many problems. \citet{semenova2019study} show that as long as a large Rashomon set exists, it is more likely that some of these models are interpretable. 

For raw data, on the other hand, neural networks have an advantage currently over other approaches \citep{alexnet}. In these raw data cases, the definition of interpretability changes; for visual data, one may require visual explanations. In such cases, as discussed earlier and in Challenges 
 \ChCaseBased{} and \ChSupDis{},
 interpretable neural networks suffice, without losing accuracy.

These two data extremes show that in machine learning, the dichotomy between the accurate black box and the less-accurate interpretable model is false. The often-discussed hypothetical choice between the accurate machine-learning-based robotic surgeon and the less-accurate human surgeon is moot once someone builds an interpretable robotic surgeon. Given that even the most difficult computer vision benchmarks can be solved with interpretable models, there is no reason to believe that an interpretable robotic surgeon would be worse than its black box counterpart. The question ultimately becomes whether the Rashomon set should permit such an interpretable robotic surgeon--and all scientific evidence so far (including a large-and-growing number of experimental papers on interpretable deep learning) suggests it would.

Our next principle returns to the data science process. \\

\noindent \textbf{Principle 4 } \textit{As part of the full data science process, one should expect both the performance metric and interpretability metric to be iteratively refined. }\\

The knowledge discovery process in Figure \ref{fig:kddprocess} explicitly shows these important feedback loops. We have found it useful in practice to create many interpretable models (satisfying the known constraints) and have domain experts choose between them. Their rationale for choosing one model over another helps to refine the definition of interpretability. Each problem can thus have its own unique interpretability metrics (or set of metrics).

The fifth principle is as follows:\\

\noindent \textbf{Principle 5 } \textit{For high stakes decisions, interpretable models should be used if possible, rather than ``explained" black box models.}\\

Hence, this survey concerns the former.
This is not a survey on Explainable AI (XAI, where one attempts to explain a black box using an approximation model, derivatives, variable importance measures, or other statistics), it is a survey on \textit{Interpretable Machine Learning} (creating a predictive model that is not a black box).
%We clarify that in this survey, we discuss \textit{interpretable} machine learning (creating a predictive model that is not a black box) rather than \textit{explainable} machine learning (where one attempts to explain a black box).
Unfortunately, these topics are much too often lumped together within the misleading term ``explainable artificial intelligence'' or ``XAI'' despite a chasm separating these two concepts \citep{Rudin19}. 
%Interpretable ML is not contained in XAI, despite the flurry of recent papers that have mistakenly lumped them together. 
%Interpretable ML predates XAI, to the beginning of the field of machine learning. 
Explainability and interpretability techniques are not alternative choices for many real problems, as the recent surveys often imply; one of them (XAI) can be dangerous for high-stakes decisions to a degree that the other is not.

Interpretable ML is not a subset of XAI. The term XAI dates from $\sim$2016, and grew out of work on function approximation; i.e., explaining a black box model by approximating its predictions by a simpler model \citep[e.g.,][]{CravenSh95,Craven1996}, or explaining a black box using local approximations. Interpretable ML also has a (separate) long and rich history, dating back to the days of expert systems in the 1950's, and the early days of decision trees. While these topics may sound similar to some readers, they differ in ways that are important in practice. 
%Reviews that focus on XAI tend to focus on methods that explain black boxes, omitting the important disclaimer that it is generally safer to aim first at inherently interpretable models.

In particular, there are many serious problems with the use of explaining black boxes posthoc, as outlined in several papers that have shown why explaining black boxes can be misleading and why explanations do not generally serve their intended purpose \citep{Rudin19,LaugelEtAl19,LakkarajuBa20}. The most compelling such reasons are:
\begin{itemize}
    \item  Explanations for black boxes are often problematic and misleading, potentially creating misplaced trust in black box models. Such issues with explanations have arisen with  assessment of fairness and variable importance \citep{RudinWaCo2020,Dimanov20} as well as uncertainty bands for variable importance \citep{gosiewska2020trust,fisher2018model}. There is an overall difficulty in troubleshooting the combination of a black box and an explanation model on top of it; if the explanation model is not always correct, it can be difficult to tell whether the black box model is wrong, or if it is right and the explanation model is wrong. 
    Ultimately, posthoc explanations are wrong (or misleading) too often. 
    
    One particular type of posthoc explanation, called saliency maps (also called attention maps) have become particularly popular in radiology and other computer vision domains despite known problems \citep{Adebayo18,chen2019looks,ZhangSoSuTaUd2019}. Saliency maps highlight the pixels of an image that are used for a prediction, but they do not explain how the pixels are used. As an analogy, consider a real estate agent who is pricing a house. A ``black box'' real estate agent would provide the price with no explanation. A ``saliency'' real estate agent would say that the price is determined from the roof and backyard, but doesn't explain how the roof and backyard were used to determine the price. 
    %(Furthermore, the roof and backyard may not actually have been the important information behind the price if the explanation is incorrect.) 
    In contrast, an interpretable agent would explain the calculation in detail, for instance, using ``comps'' or comparable properties to explain how the roof and backyard are comparable between properties, and how these comparisons were used to determine the price. One can see from this real estate example how the saliency agent's explanation is insufficient. 
    
    Saliency maps also tend to be unreliable; researchers often report that different saliency methods provide different results, making it unclear which one (if any) actually represents the network's true attention.\footnote{To clear possible confusion, techniques such as SHAP and LIME are tools for explaining black box models, are not needed for inherently interpretable models, and thus do not belong in this survey. These methods determine how much each variable contributed to a prediction. Interpretable models do not need SHAP values because they already explicitly show what variables they are using and how they are using them. For instance, sparse decision trees and sparse linear models do not need SHAP values because we know exactly what variables are being used and how they are used. Interpretable supervised deep neural networks that use case-based reasoning (Challenge \#\ChCaseBased{}) do not need SHAP values because they explicitly reveal what part of the observation they are paying attention to, and in addition, how that information is being used (e.g., what comparison is being made between part of a test image and part of a training image). Thus, if one creates an interpretable model, one  does not need LIME or SHAP whatsoever.}

    \item Black boxes are generally unnecessary, given that their accuracy is generally not better than a well-designed interpretable model. Thus, explanations that seem reasonable can undermine efforts to find an interpretable model of the same level of accuracy as the black box.  
    \item Explanations for complex models hide the fact that complex models are difficult to use in practice for many different reasons. Typographical errors in input data are a prime example of this issue \citep[as in the use of COMPAS in practice, see][]{Wexler17}. A model with 130 hand-typed inputs is more error-prone than one involving 5 hand-typed inputs. 
%Also the question of how to calibrate outside information into a model that one does not understand.
\end{itemize}

In that sense, \textit{explainability methods are often used as an excuse to use a black box model--whether or not one is actually needed}. Explainability techniques give authority to black box models rather than suggesting the possibility of models that are understandable in the first place \citep{RudinRadin2019}.
%Additional points about the dangers of posthoc counterfactual explanations for black box models are made by \citet{LaugelEtAl19}.
%With this survey, we hope to help remedy serious problems with the guidance that permeates the vast majority of the literature on XAI. In particular, we stress that there is a chasm between Interpretable ML and XAI (explaining black boxes). 
 
%which miss the crucial point: they tend to focus almost entirely on explainable ML, with interpretable ML as an afterthought. They tend to view the field as starting around 2016, despite the fact that interpretable AI has been around since the days of expert systems in the 1950's???. They generally list a small handful of interpretable ML techniques but many variations of methods for explainable AI, indicating that one should consider explainability methods rather than aiming first at inherently interpretable models. 
%What if one's problem could be solved with a simple four-leaf decision tree? Or a three-term linear model? Creating a complex deep neural network model and explaining it afterwards (as XAI might be seen to prescribe) would be marvelously silly.

XAI surveys have (thus far) universally failed to acknowledge the important point that interpretability begets accuracy when considering the full data science process, and not the other way around. Perhaps this point is missed because of the more subtle fact that one does generally lose accuracy when approximating a complicated function with a simpler one, and these imperfect approximations are the foundation of XAI. (Again the approximations must be imperfect, otherwise one would throw out the black box and instead use the explanation as an inherently interpretable model.) But function approximators are not used in interpretable ML; instead of approximating a known function (a black box ML model), interpretable ML can choose from a potential myriad of approximately-equally-good models, which, as we noted earlier, is called ``the Rashomon set'' \citep{breiman2001statistical,fisher2018model,semenova2019study}. We will discuss the study of this set in Challenge \ChRashomon.
% \ref{sec:Rashomon}. 
Thus, \textit{when one explains black boxes, one expects to lose accuracy, whereas when one creates an inherently interpretable ML model, one does not.}

In this survey, we do not aim to provide yet another dull taxonomy of  ``explainability'' terminology. 
%We do not use the misleading terminology from the field of XAI; we do not claim interpretable machine learning is \textit{part of} explainable AI, as it is not. Its origins predate explanations of black boxes by many decades \cite{}.
%We do not invent more terminology, as the XAI surveys do, since 
The ideas of interpretable ML can be stated in just one sentence: an interpretable model is constrained, following a domain-specific set of constraints that make reasoning processes understandable.
 %We do not lament (as many others have done \cite{kodratoff1994comprehensibility}) about the lack of a firm definition of interpretability, as it should be domain dependent.
Instead, we highlight important challenges,
%In this survey, we highlight topics that we think are interesting and useful, 
each of which can serve as a starting point for someone wanting to enter into the field of interpretable ML.

\section{Sparse Logical Models: Decision Trees, Decision Lists, and Decision Sets}\label{subsec:sparselogical}

The first two challenges involve optimization of sparse models. We discuss both sparse logical models in Challenge \#\ChLogical{} and scoring systems (which are sparse linear models with integer coefficients) in Challenge \#\ChScoring. 
Sparsity is often used as a measure of interpretability for tabular data where the features are meaningful. Sparsity is useful because humans can handle only 7$\pm$2 cognitive entities at the same time \citep{miller1956magical}, and sparsity makes it easier to troubleshoot, check for typographical errors, and reason about counterfactuals (e.g., ``How would my prediction change if I changed this specific input?''). Sparsity is rarely the only consideration for interpretability, but if we can design models to be sparse, we can often handle additional constraints. Also, if one can optimize for sparsity, a useful baseline can be established for how sparse a model could be with a particular level of accuracy.

We remark that more sparsity does not always equate to more interpretability. \citet{Elomaa94indefense} and \citet{Freitas:2014ic} make the point that ``Humans by nature are mentally opposed to too simplistic representations of complex relations.'' For instance, in loan decisions, we may choose to have several sparse mini-models for length of credit, history of default, etc., which are then assembled at the end into a larger model composed of the results of the mini-models \citep[see][who attempted this]{chen2018interpretable}. On the other hand, sparsity is necessary for many real applications, particularly in healthcare and criminal justice where the practitioner needs to memorize the model.

Logical models, which consist of logical statements involving ``if-then,'' ``or,'' and ``and'' clauses  are among the most popular algorithms for interpretable machine learning, since their statements provide human-understandable reasons for each prediction. 

When would we use logical models? Logical models are usually an excellent choice for modeling categorical data with potentially complicated interaction terms (e.g., ``IF (female AND high blood pressure AND congenital heart failure), OR (male AND high blood pressure AND either prior stroke OR age $>$ 70) THEN predict Condition 1 = true''). Logical models are also excellent for multiclass problems. Logical models are also known for their robustness to outliers and ease of handling missing data. Logical models can be highly nonlinear, and even classes of sparse nonlinear models can be quite powerful. 

\begin{figure}[htbp]
    \centering
    \includegraphics[scale=0.45]{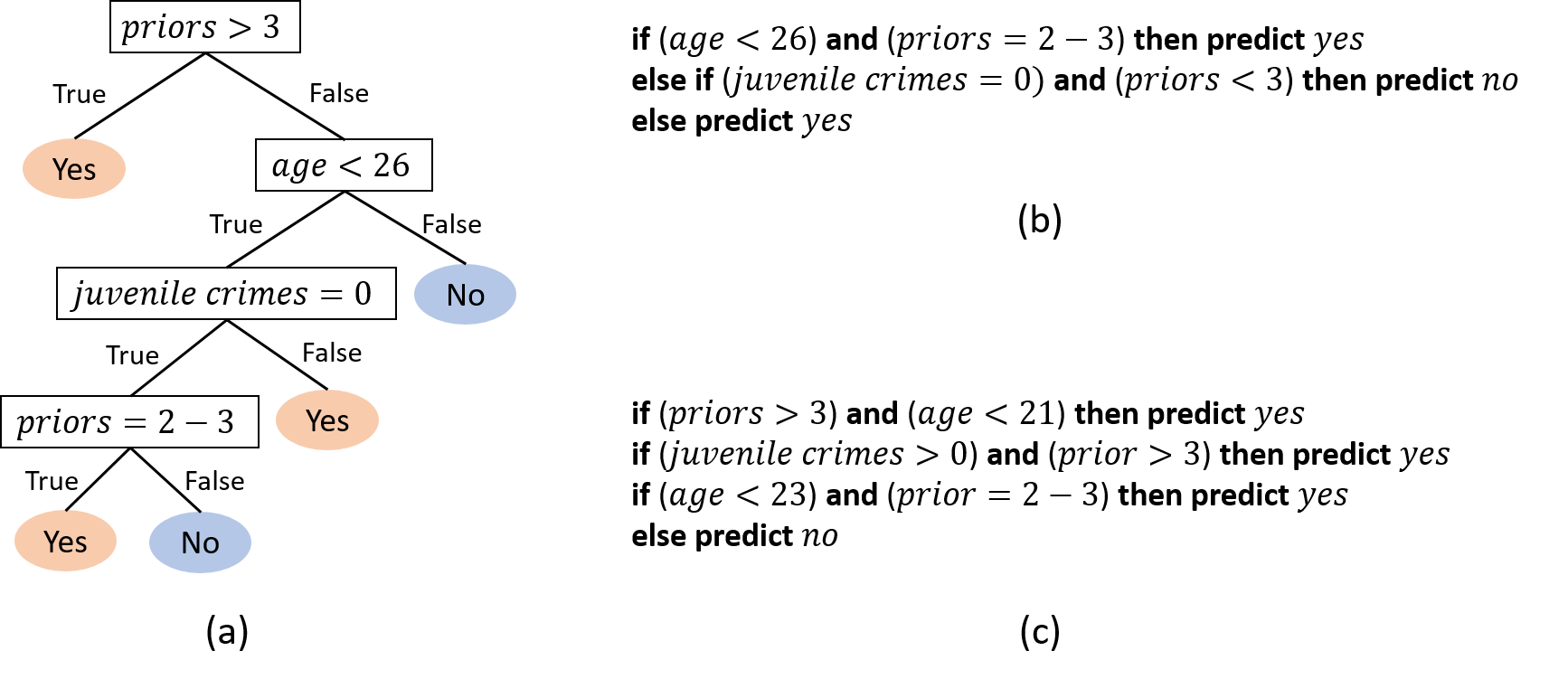}
    \caption{Predicting which individuals are arrested within two years of release by a decision tree (a), a decision list (b), and a decision set (c). The dataset used here is the ProPublica recidivism dataset \citep{propublica2016}}
    \label{fig:example_sparse}
\end{figure}

Figure \ref{fig:example_sparse} visualizes three logical models: a decision tree, a decision list, and a decision set. \textit{Decision trees} are tree-structured predictive models where each branch node tests a condition and each leaf node makes a prediction. 
\textit{Decision lists}, identical to rule lists or one-sided decision trees, are composed of if-then-else statements. The rules are tried in order, and the first rule that is satisfied makes the prediction. Sometimes rule lists have multiple conditions in each split, whereas decision trees typically do not. 
A \textit{decision set}, also known as a ``disjunction of conjunctions,''  ``disjunctive normal form'' (DNF), or  an ``OR of ANDs'' is comprised of an unordered collection of rules, where each rule is a conjunction of conditions. A positive prediction is made if at least one of the rules is satisfied. 
Even though these logical models seem to have very different forms, they are closely related:
every decision list is a (one-sided) decision tree and every decision tree can be expressed as an equivalent decision list (by listing each path to a leaf as a decision rule). The collection of leaves of a decision tree (or a decision list) also forms a decision set.
%Compared with the decision list, rules in a decision set do not connect by ``else'' statement. 
%Figure \ref{fig:example_sparse} also visualizes the dissimilarities between these logical models. For example, decision trees usually use a single condition at each split, while decision lists and sets have multiple conditions at a split. From the interpretability perspective, \cite{lakkaraju2016interpretable} proposed that decision sets are more interpretable than decision lists. 
% Sparsity is often a useful measure of interpretability for logical models (i.e., the number of logical conditions) but it is rarely the only measure one would consider. On the other hand, if one can optimize for sparsity, a useful baseline can be established for how sparse a model could be with a particular level of accuracy. (And, if an algorithm can optimize for sparsity, it is probably capable of optimizing for other objectives too.)
%with some discussions about decision lists and decision sets. 

\begin{figure}[ht]
    \centering
    \begin{subfigure}[]{0.65\linewidth}
     \includegraphics[scale=0.35]{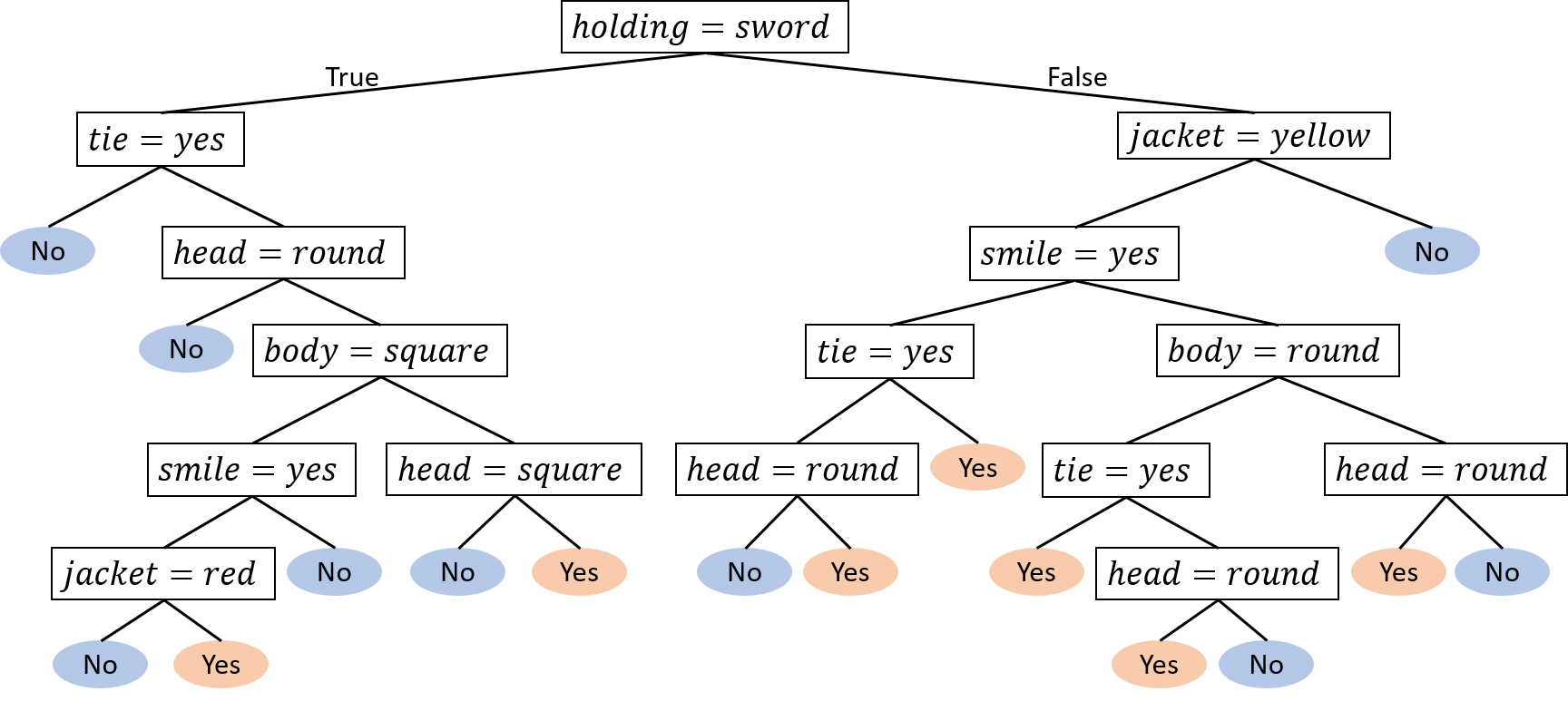}
     \caption{training accuracy: 75.74\%; test accuracy: 69.44\%}
    \end{subfigure}
    \begin{subfigure}[]{0.3\linewidth}
     \includegraphics[scale=0.35]{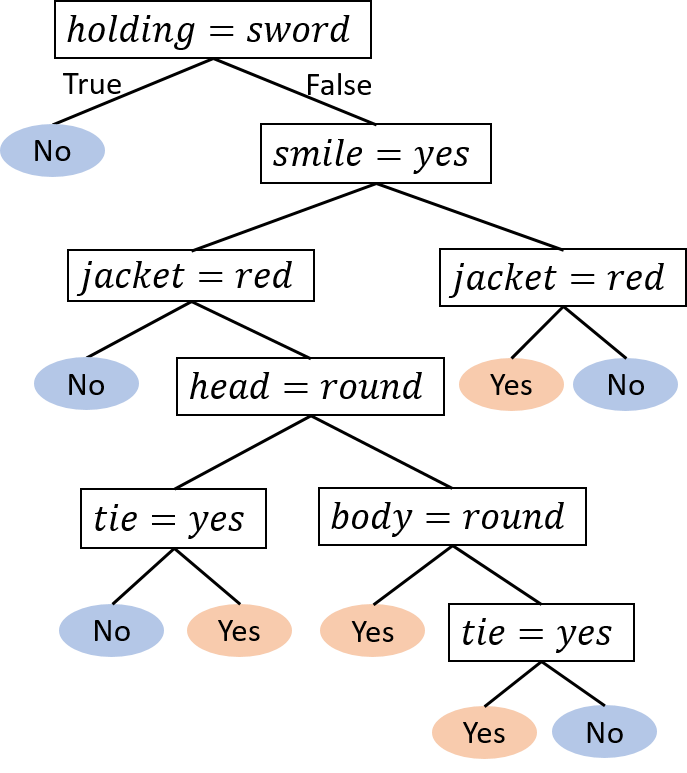}
     \caption{training accuracy: 81.07\%; test accuracy: 73.15\%}
    \end{subfigure}
    \caption{(a) 16-leaf decision tree learned by CART \citep{Breiman84} and (b) 9-leaf decision tree generated by GOSDT \citep{lin2020generalized} for the classic Monk 2 dataset \citep{Dua:2019}. The GOSDT tree is optimal with respect to a balance between accuracy and sparsity.}
    \label{fig:compare_trees}
\end{figure}

Let us provide some background on decision trees. Since \cite{MorganSo1963} developed the first decision tree algorithm, many works have been proposed to build decision trees and improve their performance.
%continuous variables for dec trees 
%dec trees have a major advantage for multiclass
%Note that very sparse models are not always interpretable or desirable. Extra constraints beyond sparsity can make these models more interpretable 
However, learning decision trees with high performance and sparsity is not easy. Full decision tree optimization is known to be an NP-complete problem \citep{laurent1976constructing}, and heuristic greedy splitting and pruning procedures have been the major type of approach since the 1980s to grow decision trees \citep{Breiman84, Quinlan93,loh1997split, mehta1996sliq}. 
%Bayesian tree models \citep{denison1998bayesian, chipman2002bayesian, chipman2010bart} can explore more search space by an MCMC sampling scheme but still build greedily from top-down (meaning that first the split at the top is chosen, then the splits below it, then the splits below them, etc.)
These greedy methods for building decision trees create trees from the top down and prune them back afterwards. They do not go back to fix a bad split if one was made. Consequently, the trees created from these greedy methods tend to be both less accurate and less interpretable than necessary. That is, greedy induction algorithms are not designed to optimize any particular performance metric, leaving a gap between the performance that a decision tree might obtain and the performance that the algorithm's decision tree actually attains, with no way to determine how large the gap is (see Figure \ref{fig:compare_trees} for a case where CART did not obtain an optimal solution, as shown by the better solution from a more recent algorithm called ``GOSDT,'' to the right). This gap can cause a problem in practice because one does not know whether poor performance is due to the choice of model form (the choice to use a decision tree of a specific size) or poor optimization (not fully optimizing over the set of decision trees of that size). 
%If trees are created through greedy induction, they tend to be large and not particularly accurate. 
When fully optimized, single trees can be as accurate as ensembles of trees, or neural networks, for many problems. Thus, it is worthwhile to think carefully about how to optimize them.

GOSDT and related modern decision tree methods solve an optimization problem that is a special case of \eqref{eqn:generic}:
\begin{equation}\label{eqn:trees}
    \min_{f\in\textrm{ set of trees}}\; \frac{1}{n}\sum_i\textrm{Loss}(f,z_i) + C\cdot\textrm{Number of leaves }(f),
\end{equation}
where the user specifies the loss function and the trade-off (regularization) parameter.

Efforts to fully optimize decision trees, solving problems related to \eqref{eqn:trees}, have been made since the 1990s %Since the late 1990s, there has been research on building optimal decision trees using optimization techniques  
\citep{Bennett96optimaldecision, dobkininduction, FarhangfarGZ08,nijssen2007mining, NijssenFromont2010,HuRuSe2019,lin2020generalized}. Many recent papers directly optimize the performance metric (e.g., accuracy) with soft or hard sparsity constraints on the tree size, where sparsity is measured by the number of leaves in the tree. Three major groups of these techniques are (1) mathematical programming, including mixed integer programming (MIP) 
\citep[see the works of ][]{Bennett96optimaldecision,RudinEr18,verwer2019learning, vilas2019, MenickellyGKS18, aghaei2020learning} and SAT solvers \citep{narodytska2018learning, hu2020learning} \citep[see also the review of ][]{carrizosamathematical}, (2) stochastic search through the space of trees  \citep[e.g.,][]{yang2017scalable, gray2008classification, papagelis2000ga}, and (3) customized dynamic programming algorithms that incorporate branch-and-bound techniques for reducing the size of the search space \citep{HuRuSe2019, lin2020generalized,nijssen2020,demirovic2020murtree}.

Decision list and decision set construction lead to the same challenges as decision tree optimization, and have a parallel development path. Dating back to 1980s, decision lists have often been constructed in a top-down greedy fashion. Associative classification methods assemble decision lists or decision sets from a set of pre-mined rules, generally either by greedily adding rules to the model one by one, or simply including all ``top-scoring'' rules into a decision set, where each rule is scored separately according to a scoring function \citep{rivest1987learning, clark1989cn2, liu1998integrating, li2001cmar, yin2003cpar, sokolova2003decision, marchand2005learning, vanhoof2010structure, rudin2013learning,clark1991rule,gaines1995induction,cohen1995fast,frank1998generating, friedman1999bump, marchand2002set, malioutov2013exact}. Sometimes decision lists or decision sets are optimized by sampling \citep{letham2015interpretable, yang2017scalable, wang2015falling}, providing a Bayesian interpretation.  Some recent works can jointly optimize performance metrics and sparsity for decision lists \citep{RudinEr18,yu2020optimal,AngelinoLaAlSeRu17-kdd, AngelinoEtAl2018, aivodji2019learning} and decision sets \citep{wang2015learning, GohRu14, lakkaraju2016interpretable, ignatiev2018sat, DashEtAl18, malioutov2018mlic, ghosh2019imli, dhamnani2019rapid, yu2020computing, cao2020learning}. Some works optimize for individual rules \citep{DashEtAl18,RudinSh2019}

Over the last few years, great progress has been made on optimizing the combination of accuracy and sparsity for logical models, but there are still many challenges that need to be solved. Some of the most important ones are as follows: 
\begin{enumerate}[\thesection .1]
    \item \textbf{Can we improve the scalability of optimal sparse decision trees?} 
    A lofty goal for optimal decision tree methods is to fully optimize trees as fast as CART produces its (non-optimal) trees. Current state-of-the-art optimal decision tree methods can handle medium-sized datasets (thousands of samples, tens of binary variables) in a reasonable amount of time (e.g., within 10 minutes) when appropriate sparsity constraints are used. But how to scale up to deal with large datasets or to reduce the running time remains a challenge. 
    
    These methods often scale exponentially in $p$, the number of dimensions of the data. Developing algorithms that reduce the number of dimensions through variable screening theorems or through other means could be extremely helpful. 
    
For methods that use the mathematical programming solvers, a good formulation is key to reducing training time. For example, MIP solvers use branch-and-bound methods, which partition the search space recursively and solve Linear Programming (LP) relaxations for each partition to produce lower bounds. Small formulations with fewer variables and constraints can enable the LP relaxations to be solved faster, while stronger LP relaxations (which usually involve more variables) can produce high quality lower bounds to prune the search space faster and reduce the number of LPs to be solved. How to formulate the problem to leverage the full power of MIP solvers is an open question. Currently, mathematical programming solvers are not as efficient as the best customized algorithms.

For customized branch-and-bound search algorithms 
such as GOSDT and OSDT \citep{lin2020generalized,HuRuSe2019},
there are several mechanisms to improve scalability: (1) effective lower bounds, which prevent branching into parts of the search space where we can prove there is no optimal solution, (2) effective scheduling policies, which help us search the space to find close-to-optimal solutions quickly, which in turn improves the bounds and again prevents us from branching into irrelevant parts of the space, (3) computational reuse, whereby if a computation involves a sum over (even slightly) expensive computations, and part of that sum has previously been computed and stored, we can reuse the previous computation instead of computing the whole sum over again, (4) efficient data structures to store subproblems that can be referenced later in the computation should that subproblem arise again.

%effective bounds to control the branching of nodes, as well as high quality lower bounds to prune more of the search space, are critical for scalability. Some current methods use sparsity constraints to control “branching” and set tighter lower bounds for each subproblem to improve “pruning.” Novel bounds will be very helpful to further reduce the search space. Meanwhile, an effective scheduling policy for exploring the space could help to speed up the search algorithm. Suppose a subproblem that can tremendously improve the current best tree is solved first. The updated tighter upper bound leads to a higher chance to prune other subproblems. Though \cite{HuRuSe2019, demirovic2020murtree} have proposed some scheduling polices, more heuristics are desired. 

    \item \textbf{Can we efficiently handle continuous variables?} While decision trees handle categorical variables and complicated interactions better than other types of approaches (e.g., linear models), one of the most important challenges for decision tree algorithms is to optimize over continuous features. Many current methods use binary variables as input \citep{HuRuSe2019, lin2020generalized, verwer2019learning, nijssen2020, demirovic2020murtree}, which assumes that continuous variables have been transformed into indicator variables beforehand (e.g., age$>50$). These methods are unable to jointly optimize the selection of variables to split at each internal tree node, the splitting threshold of that variable (if it is continuous), and the tree structure (the overall shape of the tree). \citet{lin2020generalized} preprocesses the data by transforming continuous features into a set of dummy variables, with many different split points; they take split points at the mean values between every ordered pair of unique values present in the training data. Doing this preserves optimality, but creates a huge number of binary features, leading to a dramatic increase in the size of the search space, and the possibility of hitting either time or memory limits. \citet{verwer2019learning, nijssen2020} preprocess the data using an approximation, whereby they consider a much smaller subset of all possible thresholds, potentially sacrificing the optimality of the solution
    %. Though such binary representation for the decision thresholds requires exponentially fewer variables, 
    \citep[see][Section 3, which explains this]{lin2020generalized}. One possible technique to help with this problem is to use \textit{similar support} bounds, identified by \citep{AngelinoEtAl2018}, but in practice these bounds have been hard to implement because checking the bounds repeatedly is computationally expensive, to the point where the bounds have never been used (as far as we know). Future work could go into improving the determination of when to check these bounds, or proving that a subset of all possible dummy variables still preserves closeness to optimality.
    %This unsolved challenge of how to handle continuous variables blocks the optimal decision tree algorithms to be applied to various real datasets. variables could be the major focus of future research. %Moreover, when both categorical and continuous variables are available in the given dataset, how to preserve their properties remains an open problem. The current optimal decision trees, decision lists, and decision sets work for medium-sized datasets, leaving scalability to larger datasets as a future challenge.
    
    %\item Sparse models are generally more interpretable than very complex models. Either soft or hard sparsity constraints are used by current optimal decision trees, decision lists, and decision sets algorithms. For example, soft constraints mean that add one or more terms to the objective function to penalize the large size of the model and let the algorithm learn the balance between loss and sparsity, and hard constraints specify a max depth, the minimum number of samples captured by each node or rule, etc., to directly control the sparsity. Though sparsity constraints help to create very sparse models, these models are not always desirable. \textcolor{red}{example?} Other constraints may make these models more interpretable.
    
    \item \textbf{Can we handle constraints more gracefully?} Particularly for greedy methods that create trees using local decisions, it is difficult to enforce global constraints on the overall tree. Given that domain-specific constraints may be essential for interpretability, an important challenge is to determine how to incorporate such constraints. Optimization approaches (mathematical programming, dynamic programming, branch-and-bound) are more amenable to global constraints, but the constraints can make the problem much more difficult. For instance, \textit{falling constraints} \citep{wang2015falling, chen2018optimization} enforce decreasing probabilities along a rule list, which make the list more interpretable and useful in practice, but make the optimization problem harder, even though the search space itself becomes smaller.
\end{enumerate}
\noindent\textbf{Example:} %Optimal decision tree methods can be applied in high-stake decision-making problems. For example, in criminal justice, optimal decision tree methods can help to predict criminal recidivism, assisting judges to make life-changing decisions in pretrial release, sentencing, and probation. With such interpretable models, both judges and individuals affected by these models can know how these decisions were made, or whether they were made in error. However, due to the scalability issue and limitations on optimizing over continuous variables, current state-of-art algorithms may use a subset of the original dataset and discretize the continuous variables such as age and run the processed dataset. Though the trees learned by these algorithms are usually more generalized and robust with better performance than those by greedy methods, it will be more desirable if we can make use of the full information in the original dataset.
%
%Suppose a financial institution wants to create a risk model for loan default. The dataset used to train the model contains a large number of variables including risk markers, months since the accounts were opened, percent of trades never delinquent, etc., and most of these variables are real-valued. For example, a dataset used for the FICO challenge has 23 continuous variables. Current state-of-art optimal decision tree methods take the binary variables as input. However, binarizing these variables by every threshold leads to more than 1500 variables in the training set, far beyond what current methods can handle (see Question 1 above). Therefore, discretizing continuous variables by some thresholds are commonly used to process the data. But, as we mentioned in the second challenge, such discretization can sacrifice optimality. So how to deal with continuous variables and how to improve scalability are challenges that optimal decision tree methods have to overcome to be widely used for real practice.
Suppose a hospital would like to create a decision tree that will be used to assign medical treatments to patients. A tree is convenient because it corresponds to a set of questions to ask the patient (one for each internal node along a path to a leaf). A tree is also convenient in its handling of multiple medical treatments; each leaf could even represent a different medical treatment. The tree can also handle complex interactions, where patients can be asked multiple questions that build on each other to determine the best medication for the patient. To train this tree, the proper assumptions and data handling were made to allow us to use machine learning to perform causal analysis (in practice these are more difficult than we have room to discuss here). The questions we discussed above arise when the variables are continuous; for instance, if we split on age somewhere in the tree, what is the optimal age to split at in order to create a sparse tree? (See Challenge \ChLogical.2.) If we have many other continuous variables (e.g., blood pressure, weight, body mass index), scalability in how to determine how to split them all becomes an issue. Further, if the hospital has additional preferences, such as  ``falling probabilities,'' where fewer questions should be asked to determine whether a patient is in the most urgent treatment categories, again it could affect our ability to find an optimal tree given limited computational resources (see Challenge \ChLogical.3).

\section{Scoring systems}%\label{subsec:scoring}
\textit{Scoring systems} are linear classification models that require users to add, subtract, and multiply only a few small numbers in order to make a prediction. These models are used to assess the risk of numerous serious medical conditions since they allow quick predictions, without the use of a computer. Such models are also heavily used in criminal justice. 
%Linear models, including scoring systems, do not handle interaction terms, so they should not be used when we suspect that complex interactions between variables are needed. Linear models also do not handle multiclass problems as naturally as, for instance, decision trees or other logical models. Scoring systems have the benefit that they are useful for counterfactual reasoning (e.g., ``How would my predictions change if I changed this feature?''). Scoring systems are extremely popular in medicine, and  doctors find scoring systems intuitive and easy to use.
Table \ref{tab:score_system} shows an example of a scoring system. A doctor can easily determine whether a patient screens positive for obstructive sleep apnea by adding points for the patient's age, whether they have diabetes, body mass index, and sex. If the score is above a threshold, the patient would be recommended to a clinic for a diagnostic test. 

Scoring systems commonly use binary or indicator variables (e.g., age $\geq$ 60) and point values (e.g., Table \ref{tab:score_system}) which make computation of the score easier for humans. 
Linear models do not handle interaction terms between variables like the logical models discussed above in Challenge \ChLogical.1, and they are not particularly useful for multiclass problems, but they are useful for counterfactual reasoning: if we ask ``What changes could keep someone's score low if they developed hypertension?'' the answer would be easy to compute using point scores such as ``4, 4, 2, 2, and -6.''  Such logic is more difficult for humans if the point scores are instead, for instance, 53.2, 41.1, 16.3, 23.6 and -61.8.
%The total comparing the score with the decision threshold. 
\begin{table}[th]
    \centering
    \begin{tabular}{|cc|c|c|}\hline
    \multicolumn{4}{|c|}{Patient screens positive for obstructive sleep apnea if Score $>$1 } \\\hline
        1. & age $\geq 60$ & 4 points & $\hdots\hdots$  \\
        2. & hypertension & 4 points & $+\hdots\hdots$ \\
        3. & body mass index $\geq 30$ & 2 points & $+\hdots\hdots$ \\
        4. & body mass index $\geq 40$ & 2 points & $+\hdots\hdots$ \\
        5. & female & -6 points & $+\hdots\hdots$ \\ \hline
        & Add points from row 1-6 & Score & $=\hdots\hdots$ \\ \hline
    \end{tabular}
    \caption{A scoring system for sleep apnea screening \citep{UstunEtAl2016}. Patients that screen positive may need to come to the clinic to be tested.}
    \label{tab:score_system}
\end{table}

%\begin{figure}
%    \centering
%    \includegraphics[scale=0.5]{example_scoring_system.png}
%    \caption{A scoring system for sleep apnea screening \citep{UstunEtAl2016}. Patients that screen positive may need to come to the clinic to be tested.}
%    \label{fig:score_system}
%\end{figure}

\textit{Risk scores} are scoring systems that have a conversion table to probabilities. For instance, a 1 point total might convert to probability 15\%, 2 points to 33\% and so on. Whereas scoring systems with a threshold (like the one in Table \ref{tab:score_system}) would be measured by false positive and false negative rates, a risk score might be measured using the area under the ROC curve (AUC) and calibration.

The development of scoring systems dates back at least to criminal justice work in the 1920s \citep{burgess1928factors}. Since then, many scoring systems have been designed for healthcare \citep{knaus1981apache,knaus1985apache, knaus1991apache, bone1992definitions, le1993new, antman2000timi, gage2001validation, moreno2005saps, six2008chest, weathers2013ptsd,than2014development}. However, none of the scoring systems mentioned so far was optimized purely using an algorithm applied to data. Each scoring system was created using a different method involving different heuristics. Some of them were built using domain expertise alone without data, and some were created using rounding heuristics for logistic regression coefficients and other manual feature selection approaches to obtain integer-valued point scores \citep[see, e.g., ][]{le1993new}.

Such scoring systems could be optimized using a combination of the user's preferences (and constraints) and data. This optimization should ideally be accomplished by a computer, leaving the domain expert only to specify the problem. However, jointly optimizing for predictive performance, sparsity, and other user constraints may not be an easy computational task. Equation \eqref{eqn:scoring} shows an example of a generic optimization problem for creating a scoring system: 
\begin{eqnarray}  \label{eqn:scoring}
    &\min_{f\in\mathcal{F}}& \frac{1}{n}\sum_i\textrm{Loss}(f,z_i) + C\cdot\textrm{Number of nonzero terms }(f),\; \textrm{ subject to }\\ \nonumber
   && f \textrm{ is a linear model, } f(\mathbf{x})=\sum_{j=1}^p\lambda_jx_j, \\ &&\nonumber
   \textrm{with small integer coefficients, that is, } \forall\;j,\; \lambda_j\in \{-10,-9,..,0,..,9,10\}\\\nonumber
   &&\textrm{and additional user constraints.}
\end{eqnarray}
Ideally, the user would specify the loss function (classification loss, logistic loss, etc.), the tradeoff parameter $C$ between the number of nonzero coefficients and the training loss, and possibly some additional constraints, depending on the domain. 

The integrality constraint on the coefficients makes the optimization problem very difficult. The easiest way to satisfy these constraints is to fit real coefficients (e.g., run logistic regression, perhaps with $\ell_1$ regularization) and then round these real coefficients to integers. However, rounding can go against the loss gradient and ruin predictive performance. Here is an example of a coefficient vector that illustrates why rounding might not work:
\begin{eqnarray*}
[5.3, 6.1, 0.31, 0.30, 0.25, 0.25, 0.24, ..., 0.05] &\rightarrow \textrm{(rounding)}\rightarrow &   [5, 6, 0, 0,  ..., 0].
%\left[0.48, 0.42,0.40,0.38,0.35,0.32, ... ,0.05 \right] &\rightarrow \textrm{(rounding)}\rightarrow & [0,0,0,...,0,0].
\end{eqnarray*}
When rounding, we lose all signal coming from all variables except the first two. The contribution from the eliminated variables may together be significant even if each individual coefficient is small, in which case, we lose predictive performance. 

Compounding the issue with rounding is the fact that $\ell_1$ regularization introduces a strong bias for very sparse problems. To understand why, consider that the regularization parameter must be set to a very large number to get a very sparse solution. In that case, the $\ell_1$ regularization does more than make the solution sparse, it also imposes a strong $\ell_1$ bias. The solutions disintegrate in quality as the solutions become sparser, then rounding to integers only makes the solution worse.

An even bigger problem arises when trying to incorporate additional constraints, as we allude to in \eqref{eqn:scoring}. Even simple constraints such as ``ensure precision is at least 20\%'' when optimizing recall would be very difficult to satisfy manually with rounding. There are four main types of approaches to building scoring systems: i) exact solutions using optimization techniques, ii) approximation algorithms using linear programming, iii) more sophisticated rounding techniques, iv) computer-aided exploration techniques.\\

%However, linear models like Lasso, elastic net, and LARS, though have regularizations to keep the model simple, not directly optimize the $\ell_0$ norm, thereby cannot guarantee an optimal balance between sparsity and performance \citep{tibshirani1996regression,zou2005regularization,efron2004least}. Moreover, the rounding technique is required to transform the learned coefficients to be integers, but rounding is known to provide suboptimal solutions.  

\noindent \textit{Exact solutions.}
There are several methods that can solve \eqref{eqn:scoring} directly \citep{ustun2013supersparse,ustun2016supersparse, ustun2017optimized,ustun2019learning,rudin2018optimized}. To date, the most promising approaches use  mixed-integer linear programming solvers (MIP solvers) which are generic optimization software packages that handle systems of linear equations, where variables can be either linear or integer. Commercial MIP solvers (currently CPLEX and Gurobi) are substantially faster than free MIP solvers, and have free academic licenses. MIP solvers can be used directly when the problem is not too large and when the loss function is discrete or linear (e.g., classification error is discrete, as it takes values either 0 or 1). These solvers are flexible and can handle a huge variety of user-defined constraints easily. However, in the case where the loss function is nonlinear, like the classical logistic loss $\sum_i \log (1+\exp(-y_i f(x_i)))$, MIP solvers cannot be used directly. For this problem using the logistic loss, \citet{ustun2019learning} create a method called RiskSLIM that uses sophisticated optimization tools: cutting planes within a branch-and-bound framework, using ``callback'' functions to a MIP solver. A major benefit of scoring systems is that they can be used as decision aids in very high stakes settings; RiskSLIM has been used to create a model (the 2HELPS2B score) that is used in intensive care units of hospitals to make treatment decisions about critically ill patients \citep{struck2017association}.

While exact optimization approaches provide optimal solutions, they struggle with larger problems. For instance, to handle nonlinearities in continuous covariates, these variables are often discretized to form dummy variables by splitting on all possible values of the covariate (similar to the way continuous variables are handled for logical model construction as discussed above, e.g., create dummy variables for age$<$30, age$<$31, age$<$32, etc.). Obviously doing this can turn a small number of continuous variables into a large number of categorical variables. One way to reduce the size of the problem is to use only a subset of thresholds (e.g., age$<$30, age$<$35, age$<$40, etc.), but it is possible to lose accuracy if not enough thresholds are included. Approximation methods can be valuable in such cases.\\

\noindent \textit{Approximate methods.}
Several works solve approximate versions of \eqref{eqn:scoring}, including works of \citet{SokolovskaEtAl18,sokolovska2017fused,billiet2016interval,billiet2017interval,carrizosastrongly}.
These works generally use a piecewise linear or piecewise constant loss function, and sometimes use the $\ell_1$ norm for regularization as a proxy for the number of terms. This allows for the possibility of solving linear programs using a mathematical programming solver, which is generally computationally efficient since linear programs can be solved much faster than mixed-integer programs.
The main problems with approximation approaches is that it is not clear how well the solution of the approximate optimization problem is to the solution of the desired optimization problem, particularly when user-defined constraints or preferences are required. Some of these constraints may be able to be placed into the mathematical program, but it is still not clear whether the solution of the optimization problem one solves would actually be close to the solution of the optimization problem we actually care about.

It is possible to use sampling to try to find useful scoring systems, which can be useful for Bayesian interpretations, though the space of scoring systems can be quite large and hard to sample \citep{RudinEr18}.\\

\noindent \textit{Sophisticated rounding methods.}
%Methods that create scoring systems without fitting then rounding was proposed in recent years. For example, \citet{ertekin2015bayesian} used a Bayesian method, where a prior is set to favor fewer significant digits, certain values for those coefficients, and sparsity in the number of coefficients and MCMC step is designed to encourage the scale of the coefficients toward their “natural scale.”  directly optimize the performance metrics with $\ell_0$ norm to produce sparse scoring systems and risk scores with optimality guarantees by solving mixed-integer linear and nonlinear programs. These methods are flexible to incorporate more constraints imposed by domain experts. Similarly, MIP solver is also used by \cite{sokolovska2017fused} to minimize the hinge loss with fused lasso penalty, thereby simultaneously learning the thresholds used to discretize the continuous variables and scores for each bin.  
Keeping in mind the disadvantages to rounding discussed above, there are some compelling advantages to sophisticated rounding approaches, namely that they are easy to program and use in practice.
%Rounding methods can be more sophisticated than simply rounding all coefficients to integers at the same time. 
Rounding techniques cannot easily accommodate constraints, but they can be used for problems without constraints, problems where the constraints are weak enough that rounding would not cause us to violate them, or they can be used within the middle of algorithms like RiskSLIM \citep{ustun2019learning} to help find optimal solutions faster. There are several variations of sophisticated rounding.  \citet{chevaleyre2013rounding} propose randomized rounding, where non-integers are rounded up or down randomly. They also propose a greedy method where the sum of coefficients is fixed and coefficients are rounded one at a time. \citet{SokolovskaEtAl18} propose an algorithm that finds a local minimum by improving the solution at each iteration until no further improvements are possible. \citet{ustun2019learning} propose a combination of rounding and ``polishing.'' Their rounding method is called Sequential Rounding. At each iteration, Sequential Rounding chooses a coefficient to round and whether to round it up or down. It makes this choice by evaluating each possible coefficient rounded both up and down, and chooses the option with the best objective. After Sequential Rounding produces an integer coefficient vector, a second algorithm, called Discrete Coordinate Descent (DCD), is used to ``polish'' the rounded solution. At each iteration, DCD chooses a coefficient, and optimizes its value over the set of integers to obtain a feasible integer solution with a better objective. All of these algorithms are easy to program and might be easier to deal with than troubleshooting a MIP or LP solver. \\

\noindent \textit{Computer-aided exploration techniques.} These are design interfaces where domain experts can modify the model itself rather than relying directly on optimization techniques to encode constraints.  \citet{billiet2016interval,billiet2017interval} created a toolbox that allows users to make manual adjustments to the model, which could potentially help users design interpretable models according to certain types of preferences. \citet{Autoscore} also suggest an expert-in-the-loop approach, where heuristics such as the Gini splitting criteria can be used to help discretize continuous variables. Again, with these approaches, the domain expert must know how they want the model to depend on its variables, rather than considering overall performance optimization.

\begin{enumerate}[\thesection .1]
    \item \textbf{Improve the scalability of optimal sparse scoring systems:} 
    As discussed, for scoring systems, the only practical approaches that produce optimal scoring systems require a MIP solver, and these approaches may not be able to scale to large problems, or optimally handle continuous variables. Current state-of-the-art methods for optimal scoring systems (like RiskSLIM) can deal with a dataset with about thousands of samples and tens of variables within an hour. However, an hour is quite long if one wants to adjust constraints and rerun it several times. How to scale up to large datasets or to reduce solving time remains a challenge, particularly when including complicated sets of constraints. 
    
%     \item \textbf{Efficient handling of thresholds for continuous variables:} As with decision trees, optimizing the thresholds comprising the dummy variables is a challenge \cite{carrizosamathematical} 
    
    \item \textbf{Ease of constraint elicitation and handling:} Since domain experts often do not know the full set of constraints they might want to use in advance, and since they also might want to adjust the model manually \citep{billiet2017interval}, a more holistic approach to scoring systems design might be useful. There are many ways to cast the problem of scoring system design with feedback from domain experts. For instance, if we had better ways of representing and exploring the Rashomon set (see Challenge \ChRashomon), domain experts might be able to search within it effectively, without fear of leaving that set and producing a suboptimal model. If we knew domain experts' views about the importance of features, we should be able to incorporate that through regularization \citep{EyeRegulariz18}. Better interfaces might elicit better constraints from domain experts and incorporate such constraints into the models. Faster optimization methods for scoring systems would allow users faster turnaround for creating these models interactively.
\end{enumerate}

\noindent \textbf{Example}: A physician wants to create a scoring system for predicting seizures in critically ill patients \citep[similarly to][]{struck2017association}. The physician has upwards of 70 clinical variables, some of which are continuous (e.g., patient age). The physician creates dummy variables for age and other continuous features, combines them with the variables that are already binarized, runs $\ell_1$-regularized logistic regression and rounds the coefficients. However, the model does not look reasonable, as it uses only one feature, and isn't very accurate. The physician thinks that the model should instead depend heavily on age, and have a false positive rate below 20\% when the true positive rate is above 70\% (see Challenge \ChScoring.2). The physician downloads a piece of software for developing scoring systems. The software reveals that a boosted decision tree model is much more accurate, and that there is a set of scoring systems with approximately the same accuracy as the boosted tree. Some of these models do not use age and violate the physician's other constraint, so these constraints are then added to the system, which restricts its search. The physician uses a built-in tool to look at models provided by the system and manually adjusts the dependence of the model on age and other important variables, in a way that still maintains predictive ability. Ideally, this entire process takes a few hours from start to finish (see Challenge \ChScoring.1). Finally, the physician takes the resulting model into the clinic to perform a validation study on data the model has not been trained with, and the model is adopted into regular use as a decision aid for physicians.

\section{Generalized Additive Models}\label{sec:gam}
\begin{figure}[th]
    \centering
    \includegraphics[scale=0.5]{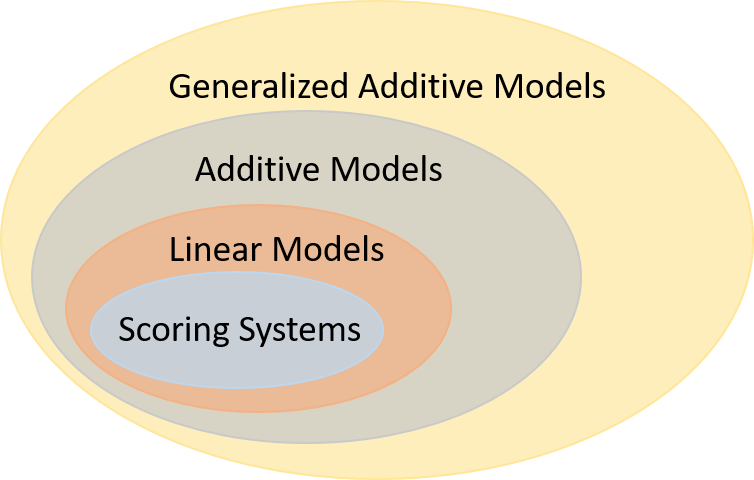}
    \caption{Hierarchical relationships between GAMs, additive models, linear models, and scoring systems.}
    \label{fig:hierarchy}
\end{figure}

Generalized additive models (GAMs) were introduced to present a flexible extension of generalized linear models (GLMs) \citep{nelder1972generalized}, allowing for arbitrary functions for modeling the influence of each feature on a response (\citealt{hastie1990generalized} see also \citealt{wood2017generalized}). 
The set of GAMs includes the set of additive models, which, in turn, includes the set of linear models, which includes scoring systems (and risk scores). Figure \ref{fig:hierarchy} shows these relationships.

\begin{figure}[ht]
    \centering
    \includegraphics[scale=0.31]{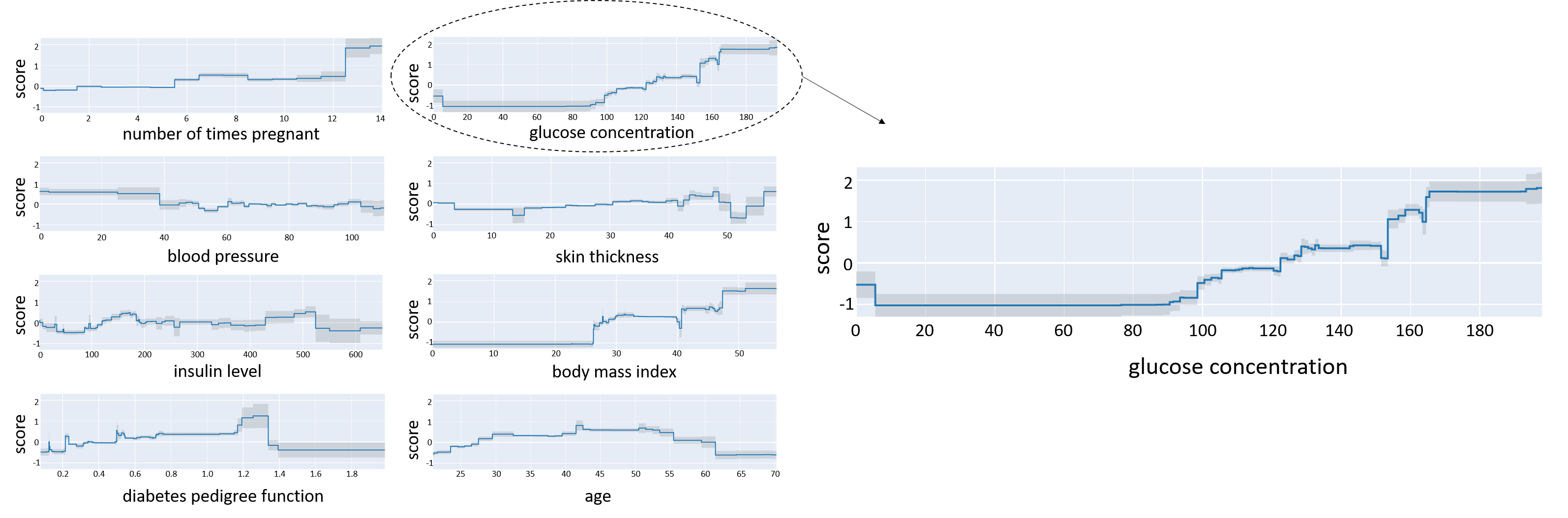}
    \caption{Left: All component functions of a GAM model trained using the \texttt{interpret} package \citep{nori2019interpretml} on a diabetes dataset \citep{Dua:2019}; Right: zoom-in of component function for glucose concentration.}
    \label{fig:example_gams}
\end{figure}

The standard form of a GAM is 
\[g(E[y]) = \beta_0 + f_1(x_{\cdot 1}) + ... + f_p( x_{\cdot p}),\] where $x_{\cdot j}$ indicates the $j$th feature, $g(\cdot)$ is a link function and the $f_i$'s are univariate component functions that are possibly nonlinear; common choices are step functions and splines. 
If the link function $g(\cdot)$ is the identity, the expression describes an additive model such as a regression model; 
if the link function is the logistic function, then the expression describes a generalized additive model that could be used for classification. The standard form of GAMs is interpretable because the model is constrained to be a linear combination of univariate component functions. 
We can plot each component function with $f_j(x_{\cdot j})$ as a function of $x_{\cdot j}$ to see the contribution of a single feature to the prediction. 
The left part of Figure \ref{fig:example_gams} shows all component functions of a GAM model (with no interactions) that predicts whether a patient has diabetes. The right enlarged figure visualizes the relationship between plasma glucose concentration after 2 hours into an oral glucose tolerance test and the risk of having diabetes. 
%Physicians might describe this enlarged figure as: ``Risk is low and constant when the glucose concentration is below 90. Then the variation of risk follows a pattern of alternating a jump and a relatively smooth line (e.g. there is a small jump at glucose concentration 100, followed by a smooth line from glucose concentration 100-120).''

If the features are all binary (or categorical), the GAM becomes a linear model and the visualizations are just step functions. The visualizations become more interesting for continuous variables, like the ones shown in Figure \ref{fig:example_gams}. If a GAM has bivariate component functions (that is, if we choose an $f_j$ to depend on two variables, which permits an interaction between these two variables), a heatmap can be used to visualize the component function on the two dimensional plane and understand the pairwise interactions \citep{lou2013accurate}. As a comparison point with decision trees, GAMs typically do not handle more than a few interaction terms, and all of these would be quadratic (i.e., involve 2 variables); this contrasts with decision trees, which handle complex interactions of categorical variables. GAMs, like other linear models, do not handle multiclass problems in a natural way. GAMs have been particularly successful for dealing with large datasets of medical records that have many continuous variables because they can elucidate complex relationships between, for instance, age, disease and mortality. (Of course, dealing with large raw medical datasets, we would typically encounter serious issues with missing data, or bias in the labels or variables, which would be challenging for any method, including GAMs.)

A component function $f_j$ can take different forms. For example, it can be a weighted sum of indicator functions, that is: 
\begin{equation}\label{eq:gamindicator}
f_j(x_{\cdot j}) = \sum_{\textrm{thresholds} j'} c_{j,j'} \mathbbm{1}[x_{\cdot j} > \theta_{j'}].
\end{equation} 
If the weights on the indicator functions are integers, and only a small set of weights are nonzero, then the GAM becomes a scoring system.
If the indicators are all forced to aim in one direction (e.g., $\mathbbm{1}[x_{\cdot j}>\theta_{j'}]$ for all $j'$, with no indicators in the other direction, $\mathbbm{1}[x_{\cdot j}<\theta_{j'}]$ ) and the coefficients $c_{j,j'}$ are all constrained to be nonnegative, then the function will be monotonic.
%, if a boosted stump is learned at each iteration $t$. 
In the case that splines are used as component functions, the GAM can be a weighted sum of the splines' basis functions, i.e. $f_j(x_{\cdot j}) = \sum_{k=1}^{K_j} \beta_{jk}b_{jk}(x_{\cdot j})$.

There are many different ways to fit GAMs. The traditional way is to use backfitting, where we iteratively train a component function to best fit the residuals from the other (already-chosen) components \citep{hastie1990generalized}. If the model is fitted using boosting methods \citep{freund1997decision, friedman2001greedy, friedman2002stochastic}, we learn a tree on each single feature in each iteration and then aggregate them together \citep{lou2012intelligible}. 
%$f_i$s can be different functions. For example, ... a sum of indicator functions such as $const*1_{[x<b]}$ or $const*1_{[x\geq b]}$ is a piecelinear or ...
%GAMs are traditionally fit using non-parametric smoothers (e.g., running line smoothers or splines) via the backfitting algorithm \citep{hastie1990generalized}. Later, the component functions have been extended to trees on a single variable \citep{lou2012intelligible} (e.g., ....) and trend filters \citep{sadhanala2019additive}, and the fitting procedures start to include penalized iteratively re-weighted least squares \citep{wood2017generalized} and boosting \citep{freund1997decision, friedman2001greedy, friedman2002stochastic}. 
Among different estimations of component functions and fitting procedures, \cite{binder2008comparison} found that boosting performed particularly well in high-dimensional settings, and \cite{lou2012intelligible} found that using a shallow bagged ensemble of trees on a single feature in each step of stochastic gradient boosting generally achieved better performance. 
%Other recent works also extend GAMs to include a small number of pairwise interaction terms to improve the performance while maintaining interpretability \citep{lou2013accurate} or to study the interpretability when GAMs are applied for multiclass classification problems \citep{zhang2019axiomatic}.  

We remark that GAMs have the advantage that they are very powerful, particularly if they are trained as boosted stumps or trees, which are reliable out-of-the-box machine learning techniques. The AdaBoost algorithm also has the advantage that it maximizes convex proxies for both classification error and area under the ROC curve (AUC) simultaneously \citep{RudinSc09,ErtekinRu11}. This connection explains why boosted models tend to have both high AUC and accuracy. 
However, boosted models are not naturally sparse, and issues with bias arise under $\ell_1$ regularization, as discussed in the scoring systems section. 

We present two interesting challenges involving GAMs:

\begin{enumerate}[\thesection .1]

    \item \textbf{How to control the simplicity and interpretability for GAMs}: 
    The simplicity of GAMs arises in at least two ways: sparsity in the number of component functions and smoothness of the component functions. Imposing monotonicity of the component functions also helps with interpretability when we have prior knowledge, e.g., that risk increases with age.  
    %In general, we want component functions to be smooth to prevent overfitting the data and we often prefer sparser models since it they generally more interpretable. 
    In the case when component functions are estimated by splines, many works apply convex regularizers (e.g., $\ell_1$) to control both smoothness and sparsity \citep{lin2006component, ravikumar2009sparse, meier2009high,yin2017convex, petersen2016fused, lou2016sparse, sadhanala2019additive, haris2019generalized}. For example, they add ``roughness'' penalties and lasso type penalties on the $f_j$'s in the objective function to control both the smoothness of component functions and sparsity of the model. Similarly, if the $f_j$'s are sums of indicators, as in \eqref{eq:gamindicator}, we could regularize to reduce the $c_{j,j'}$ coefficients to induce smoothness.
    %Specifically, they control the smoothness of component functions by specifying the function class or adding roughness penalties in the objective function and impose the sparsity by adding lasso type penalties on $f_j$s. 
    These penalties are usually convex, therefore, when combined with convex loss functions, convex optimization algorithms minimize their (regularized) objectives. There could be some disadvantages to this setup: (1) as we know, $\ell_1$ regularization imposes a strong bias on very sparse solutions. (2)  \citet{lou2012intelligible} find that imposing smoothness may come at the expense of accuracy, (3) imposing smoothness may miss important naturally-occurring patterns like a jump in a component function; in fact, they found such a jump in mortality as a function of age that seems to occur around retirement age. In that case, it might be more interpretable to include a smooth increasing function of age plus an indicator function around retirement age. At the moment, these types of choices are hand-designed, rather than automated.
    
    As mentioned earlier, boosting can be used to train GAMs to produce accurate models. However, sparsity and smoothness are hard to control with AdaBoost since it adds a new term to the model at each iteration.
    
    \item \textbf{How to use GAMs to troubleshoot complex datasets?} GAMs are often used on raw medical records or other complex data types, and these datasets are likely to benefit from troubleshooting. Using a GAM, we might find counterintuitive patterns; e.g., as shown in \cite{caruana2015intelligible}, asthma patients fared better than non-asthma patients in a health outcomes study. \cite{caruana2015intelligible} provides a possible reason for this finding, which is that asthma patients are at higher natural risk, and are thus given better care, leading to lower observed risk. Medical records are notorious for missing important information or providing biased information such as billing codes. Could a GAM help us to identify important missing confounders, such as retirement effects or special treatment for asthma patients? Could GAMs help us reconcile medical records from multiple data storage environments? These data quality issues can be really important. 
    
 %   On the other hand, when the model is fitted by boosting with trees estimating each component function, we usually obtain better performance than using splines, but smoothness and sparsity are hard to control. \cite{lou2012intelligible} requires trees to contain a fixed number of leaves or capture a specific number of samples. However, each component function is finally a sum of indicator functions generated in all iterations, which might not be smooth. Moreover, it is not natural for boosting to do feature selection and this group of methods is not necessarily optimal. 
\end{enumerate}

%To Be MOVED Because of that, missing data becomes an issue (imagine if 10 out of 90 features in the model are missing). Also uncertainty quantification becomes more difficult with complex models.

\noindent\textbf{Example}: Suppose a medical researcher has a stack of raw medical records and would like to predict the mortality risk for pneumonia patients.  
The data are challenging, including missing measurements (structural missingness as well as data missing not at random, and unobserved variables), insurance codes that do not convey exactly what happened to the patient, nor what their state was. However, the researcher decides that there is enough signal in the data that it could be useful in prediction, given a powerful machine learning method, such as a GAM trained with boosted trees. There are also several important continuous variables, such as age, that could be visualized.
%The researcher hopes to produce predictions with uncertainty quantification.
A GAM with a small number of component functions might be appropriate since the doctor can visualize each component function. If there are too many component functions (GAM without sparsity control), analyzing contributions from all of them could be overwhelming (see Challenge \ChGAM.1). If the researcher could control the sparsity, smoothness, and monotonicity of the component functions, she might be able to design a model that not only predicts well, but also reveals interesting relationships between observed variables and outcomes. This model could also help us to determine whether important variables were missing, recorded inconsistently or incorrectly, and could help identify key risk factors (see Challenge \ChGAM.2).

From there, the researcher might want to develop even simpler models, such as decision trees or scoring systems, for use in the clinic (see Challenges \ChLogical{} and \ChScoring).
%Suppose the doctor wants to know not only whether a patient has liver fibrosis but also the level of liver fibrosis a patient might have. Then it is a multi-class classification problem with five different labels (no fibrosis, portal fibrosis, few septa, many septa, cirrhosis) and the first challenge we discussed above arise. 

\section{Modern case-based reasoning}\label{sec:case}
%not just knn, reason about partial cases. Chaofan.  Also cite Been Kim's Bayesian Case Model paper\\
%Cite Tong's new paper and other citations to protopnet if possible: https://arxiv.org/abs/2007.01777\\

% What is case-based reasoning and why it is interesting in interpretable ML
Case-based reasoning is a paradigm that involves solving a new problem using known solutions to similar past problems \citep{aamodt1994case}. It is a problem-solving strategy that we humans use naturally in our decision-making processes \citep{newell1972human}. For example, when ornithologists classify a bird, they will look for specific features or patterns on the bird and compare them with those from known bird species to decide which species the bird belongs to. The interesting question is: can a machine learning algorithm emulate the case-based reasoning process that we humans are accustomed to? A model that performs case-based reasoning is appealing, because by emulating how humans reason, the model can explain its decision-making process in an interpretable way. 

The potential uses for case-based reasoning are incredibly broad: whereas the earlier challenges apply only to tabular data, case-based reasoning applies to both tabular and raw data, including computer vision. For computer vision and other raw data problems, we distinguish between the feature extraction and prediction steps. While a human may not be able to understand the full mapping from the original image to the feature (or concept) space, the human may be able to visually verify that a particular interpretable concept/feature has been extracted from an image. After this step, the features/concepts are combined to form a prediction, which is usually done via a sparse linear combination, or with another calculation that a human can understand. 

\begin{figure}[ht]
    \centering
    \includegraphics[width=.5\linewidth]{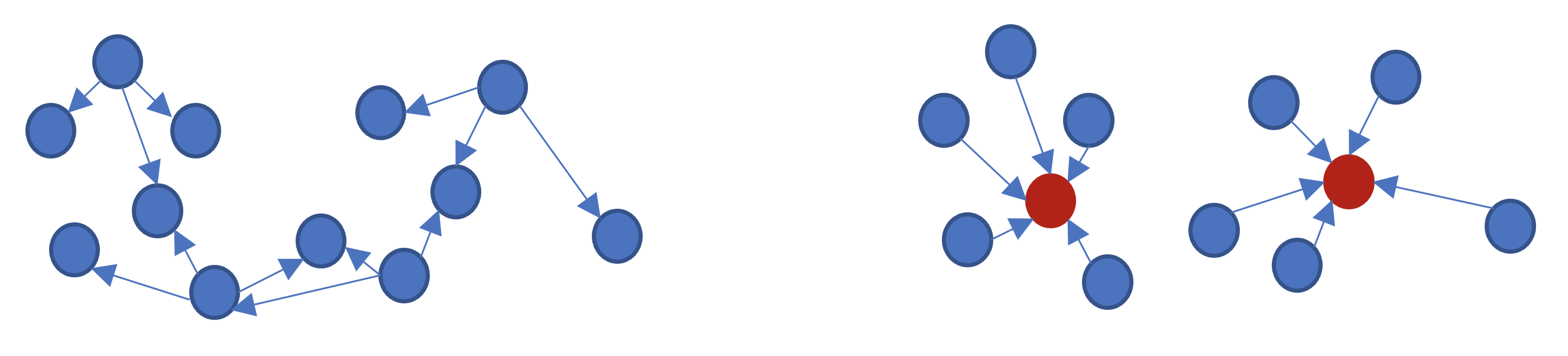}
    \caption{Case-based reasoning types. \textit{Left:} Nearest neighbors (just some arrows are shown for 3-nearest neighbors). \textit{Right:} Prototype-based reasoning, shown with two prototypes.}
    \label{fig:CBR1}
\end{figure}

% Two types of case-based reasoning: kNN and prototype-based models
Case-based reasoning has long been a subject of interest in the artificial intelligence (AI) community \citep{riesbeck1989inside,kolodner1988proceedings,hammond1989proceedings}. There are, in general, two types of case-based reasoning techniques: (i) nearest neighbor-based techniques, and (ii) prototype-based techniques, illustrated in Figure \ref{fig:CBR1}. There are many variations of each of these two types. 
%Both types are special cases of \eqref{eqn:generic}, where the models are constrained to reason about an observation based on its similarity to other observations.

% Nearest neighbor-based techniques: history and recent work (deep kNN, deep weighted averaging classifiers)
\textbf{
Nearest neighbor-based techniques.} These techniques make a decision for a previously unseen test instance, by finding training instances that most closely resemble the particular test instance (i.e., the training instances that have the smallest ``distance'' or the largest ``similarity'' measures to the test instance). A classic example of nearest neighbor-based techniques is $k$-nearest neighbors ($k$NN) \citep{fix1951discriminatory,cover1967nearest}. Traditionally, a $k$NN classifier is non-parametric and requires no training at all -- given a previously unseen test instance, a $k$NN classifier finds the $k$ training instances that have the smallest $\ell^2$ distances to the test instance, and the class label of the test instance is predicted to be the majority label of those $k$ training instances. Many variants of the original $k$NN classification scheme have been developed in the 1970s and 80s \citep{dudani1976distance,fukunaga1973optimization,keller1985fuzzy,bezdek1986generalized}. Some later papers on nearest neighbor-based techniques focused on the problem of ``adaptive $k$NN,'' where the goal is to learn a suitable distance metric to quantify the dissimilarity between any pair of input instances (instead of using a pre-determined distance metric such as the Euclidean $\ell^2$ distance measure), to improve the performance of nearest neighbor-based techniques. For example, \citet{weinberger2009distance} proposed a method to learn a parametrized distance metric (such as the matrix in a Mahalanobis distance measure) for $k$NN. Their method involves minimizing a loss function on the training data, such that for every training instance, the distance between the training instance and its ``target'' neighbors (of the same class) will be minimized while the distance between the training instance and its ``imposter'' neighbors (from other classes) will be maximized (until those neighbors are at least $1$ distance unit away from the training instance). More recently, there are works that began to focus on ``performing $k$NN in a learned latent space,'' where the latent space is often learned using a deep neural network. For example, \cite{salakhutdinov2007learning} proposed to learn a nonlinear transformation, using a deep neural network that transforms the input space into a feature space where a $k$NN classifier will perform well (i.e., deep $k$NN). \cite{papernot2018deep} proposed an algorithm for deep $k$NN classification, which uses the $k$-nearest neighbors of a test instance from every hidden layer of a trained neural network. \cite{card2019deep} introduced a deep weighted averaging classifier, which classifies an input based on its latent-space distances to other training examples.

We remark that the notion of ``adaptive $k$NN'' %(where the goal is to learn a distance metric so that $k$NN performs well)
is mathematically the same as ``performing $k$NN in a learned latent space.'' Let us show why that is. In adaptive kNN, we would learn a distance metric $d(\cdot,\cdot)$ such that k-nearest neighbors tends to be an accurate classifier. The distance between points $x_1$ and $x_2$ would be $d(x_1,x_2)$. For latent space nearest neighbor classification, we would learn a mapping $\phi: x \rightarrow \phi(x)$  from our original space to the latent space, and then compute $\ell_2$ distances in the latent space. That is, $\|\phi(x_1),\phi(x_2)\|_2$. Equating these two perspectives, we have $d(x_1,x_2)=\|\phi(x_1),\phi(x_2)\|_2$. That is, the latent space mapping acts as a transformation of the original metric so that $\ell_2$ distance works well for $k$NN.

As an aside, the problem of \textit{matching in observational causal inference} also can use case-based reasoning. Here treatment units are matched to similar control units to estimate treatment effects. We refer readers to recent work on this topic for more details \citep{FLAME,MaltsObsStudies}.

% Prototype-based techniques: history and recent work
\textbf{Prototype-based techniques.} Despite the popularity of nearest-neighbor techniques, those techniques often require a substantial amount of distance computations (e.g., to find out the nearest neighbors of a test input), which can be slow in practice. Also, it is possible that the nearest neighbors may not be particularly good representatives of a class, so that reasoning about nearest neighbors may not be interpretable. Prototype-based techniques are an alternative to nearest-neighbor techniques that have neither of these disadvantages. Prototype-based techniques learn, from the training data, a set of \textit{prototypical cases} for comparison. Given a previously unseen test instance, they make a decision by finding prototypical cases (instead of training instances from the entire training set) that most closely resemble the particular test instance. One of the earliest prototype learning techniques is learning vector quantization (LVQ) \citep{kohonen1995learning}. In LVQ, each class is represented by one or more prototypes, and points are assigned to the nearest prototype. During training, if the training example's class agrees with the nearest prototype's class, then the prototype is moved closer to the training example; otherwise the prototype is moved further away from the training example. In more recent works, prototype learning is also achieved by solving a discrete optimization program, which selects the ``best'' prototypes from a set of training instances according to some training objective. For example, \cite{bien2011prototype} formulated the prototype learning problem as a set-cover integer program (an NP-complete problem), which can be solved using standard approximation algorithms such as relaxation-and-rounding and greedy algorithms. \cite{kim2016examples} formulated the prototype learning problem as an optimization program that minimizes the squared maximum mean discrepancy, which is a submodular optimization problem and can be solved approximately using a greedy algorithm. 

\textbf{Part-based prototypes.} One issue that arises with both nearest neighbor and prototype techniques is the comparison of a \textit{whole} observation to another \textit{whole} observation. This makes little sense, for instance, with images, where some aspects resemble a known past image, but other aspects resemble a different image. For example, we consider architecture of buildings: while some architectural elements of a building may resemble one style, other elements resemble another. Another example is recipes. A recipe for a cheesecake with strawberry topping may call for part of a strawberry pancake recipe according to the typical preparation for a strawberry sauce, while the cheesecake part of the recipe could follow a traditional plain cheesecake recipe. In that case, it makes more sense to compare the strawberry cheesecake recipe to both the pancake recipe and the plain cheesecake recipe. Thus, some newer case-based reasoning methods have been comparing \textit{parts} of observations to \textit{parts} of other observations, by creating comparisons on subsets of features. This allows case-based reasoning techniques both more flexibility and more interpretability.

\citet{kim2014bayesian} formulated a prototype-parts learning problem for structured (tabular) data using a Bayesian generative framework. They considered the example (discussed above) of recipes in their experiments. \citet{wu2017prototypal} used a convex combination of training instances to represent a prototype, where the prototype does not necessarily need to be a member of the training set. Using a convex combination of training examples as a prototype would be suitable for some data types (e.g., tabular data, where a convex combination of real training examples might resemble a realistic observation), but for images, averaging the latent positions of units in latent space may not correspond to a realistic-looking image, which means the prototype may not look like a real image, which could be a disadvantage to this type of approach.

\begin{figure}[t]
  \centering
    \includegraphics[scale=0.65,trim={0 1.5cm 0 0},clip]{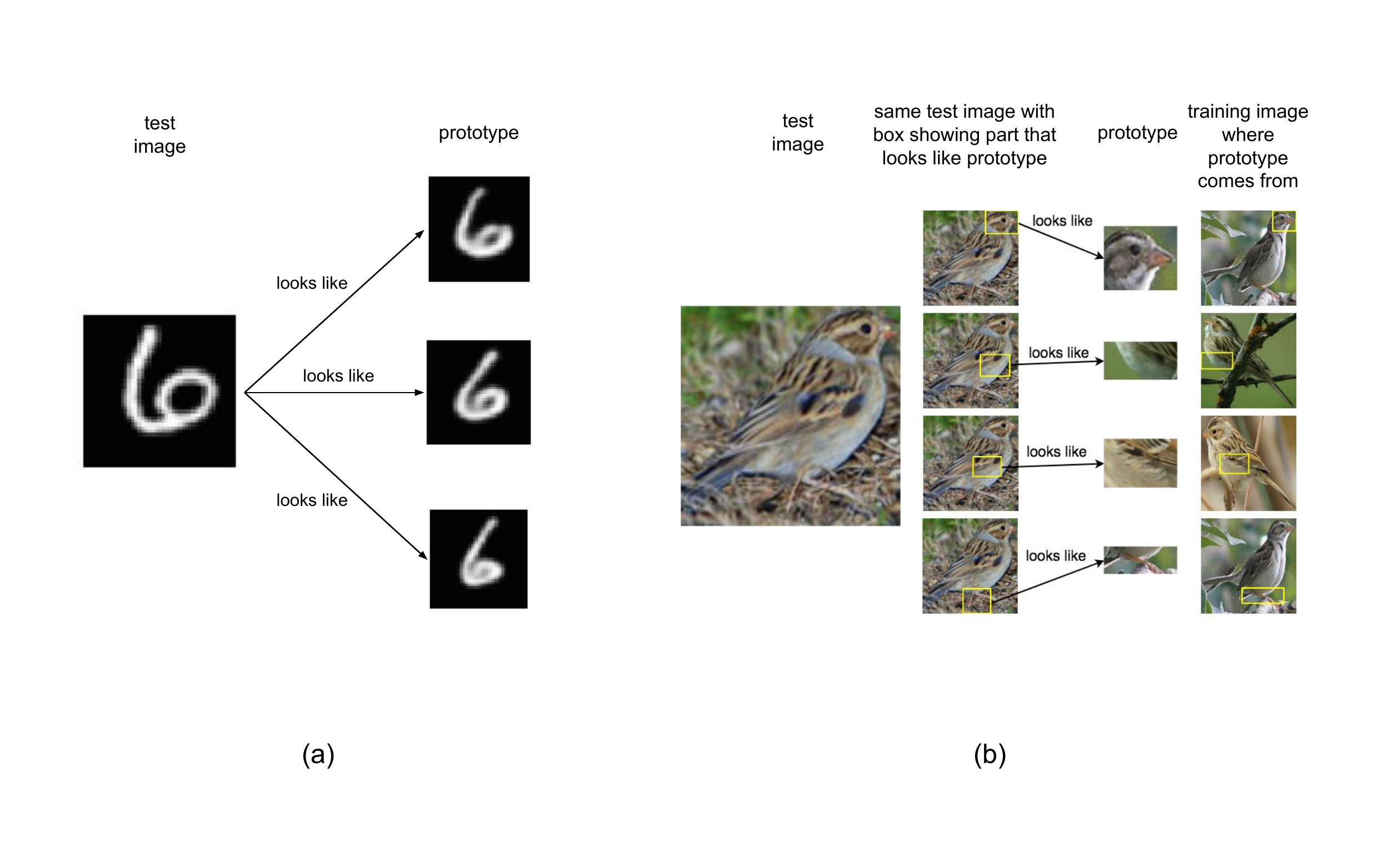}
  \caption{(a) Prototype-based classification of the network in \cite{LiLiuChenRudin}. The network compares a previously unseen image of ``6'' with 15 prototypes of handwritten digits, learned from the training set, and classifies the image as a 6 because it looks like the three prototypes of handwritten 6's, which have been visualized by passing them through a decoder from latent space into image space.
  %The prototypes are visualized by passing them through the decoder, and the number below each visualized prototype indicates the $\ell^2$ distance between that prototype and the test image in the encoding space. 
  (b) Part-based prototype classification of a ProtoPNet in \cite{chen2019looks}. The ProtoPNet compares a previously unseen image of a bird with prototypical parts of a clay colored sparrow, which are learned from the training set. It classifies the image as a clay colored sparrow because (the network thinks that) its head looks like a prototypical head from a clay-colored sparrow, its wing bars look like prototypical wing bars from a clay-colored sparrow, and so on. Here, the prototypes do not need to be passed through a decoder, they are images from the training set.}
  \label{fig:deep-CBR}
\end{figure}

Recently, there are works that integrate deep learning with prototype- and prototype-parts-based classification. This idea was first explored by \citet{LiLiuChenRudin} for image classification. In this work, \citeauthor{LiLiuChenRudin} created a neural network architecture that contains an autoencoder and a special prototype layer, where each unit of that layer (i.e., a prototype) stores a weight vector that resembles some encoded training input. Given an input instance, the network compares the encoded input instance with the learned prototypes (stored in the prototype layer), which can be visualized using the decoder. The prediction of the network is based on the $\ell^2$ distances between the encoded input instance and the learned prototypes. The authors applied the network to handwritten digit recognition \citep[MNIST,][]{lecun2010mnist}, and the network was able to learn prototypical cases of how humans write digits by hand. Given a test image of a handwritten digit, the network was also able to find prototypical cases that are similar to the test digit (Figure \ref{fig:deep-CBR}(a)).

More recently, \citet{chen2019looks} extended the work of \citet{LiLiuChenRudin} to create a \textit{prototypical part network} (ProtoPNet) whose prototype layer stores \textit{prototypical parts} of encoded training images. The prototypical parts are patches of convolutional-neural-network-encoded training images, and represent typical features observed for various image classes. Given an input instance, the network compares an encoded input image with each of the learned prototypical parts, and generates a \textit{prototype activation map} that indicates both the location and the degree of the image patch most similar to that prototypical part. The authors applied the network to the benchmark Caltech-UCSD Birds-200-2011 (CUB-200-2011) dataset \citep{CUB_200_2011} for bird recognition, and the ProtoPNet was able to learn prototypical parts of $200$ classes of birds, and use these prototypes to classify birds with an accuracy comparable to non-interpretable black-box models. Given a test image of a bird, the ProtoPNet was able to find prototypical parts that are similar to various parts of the test image, and was able to provide an explanation for its prediction, such as ``this bird is a clay-colored sparrow, because its head looks like that prototypical head from a clay-colored sparrow, and its wing bars look like those prototypical wing bars from a clay-colored sparrow'' (Figure \ref{fig:deep-CBR}(b)). In their work, \citeauthor{chen2019looks} also removed the decoder, and instead introduced \textit{prototype projection}, which pushes every prototypical part to the nearest encoded training patch of the same class for visualization. This improved the visualization quality of the learned prototypes (in comparison to the approach of \citealt{LiLiuChenRudin} which used a decoder).

The works of \cite{LiLiuChenRudin} and \cite{chen2019looks} have been extended in the domain of deep case-based reasoning and deep prototype learning. In the image recognition domain, \cite{nauta2020looks} proposed a method for explaining what visual characteristics a prototype (in a trained ProtoPNet) is looking for; \cite{nauta2020neural} proposed a method for learning neural prototype trees based on a prototype layer; \cite{rymarczyk2020protopshare} proposed data-dependent merge-pruning of the prototypes in a ProtoPNet, to allow prototypes that activate on similarly looking parts from various classes to be pruned and shared among those classes. In the sequence modeling domain (such as natural language processing), \cite{ming2019interpretable} and \cite{hong2020interpretable} took the concepts in \cite{LiLiuChenRudin} and \cite{chen2019looks}, and integrated prototype learning into recurrent neural networks for modeling sequential data. \citet{BarnettEtAl2021} extended the ideas of \citet{chen2019looks} and developed an application to interpretable computer-aided digital mammography.

% Limitations and future directions?
Despite the recent progress, many challenges still exist in the domain of case-based reasoning, including:

\begin{enumerate}[\thesection .1]
    \item \textbf{How to extend the existing case-based reasoning approach to handling more complex data, such as video?}
    Currently, case-based reasoning has been used for structured (tabular) data, static images, and simple sequences such as text. It remains a challenge to extend the existing case-based reasoning approach to handling more complex data, such as video data, which are sequences of static images. While \cite{trinh2021interpretable} recently extended the ideas in \cite{chen2019looks} and developed a dynamic prototype network (DPNet) which learns prototypical video patches from deep-faked and real videos, how to efficiently compare various videos and find similar videos (for either nearest-neighbor or prototype-based classification) remains an open challenge for case-based video classification tasks. Performing case-based reasoning on video data is technically challenging because of the high dimensionality of the input data. On the other hand, a video is an ordered combination of frames and we can take advantage of the sequential nature of the data. For example, current algorithms can be refined or improved with more information that comes from neighboring video frames; could prototypes be designed from neighboring frames or parts of the frames?
    
    \item \textbf{How to integrate prior knowledge or human supervision into prototype learning?}
    Current approaches to prototype learning do not take into account prior knowledge or expert opinions. At times, it may be advantageous to develop prototype-learning algorithms that collaborate with human experts in choosing prototypical cases or prototypical features. For example, in healthcare, it would be beneficial if a prototype-based classifier learns, under the supervision of human doctors, prototypical signs of cancerous growth. For example, the doctors might prune prototypes, design them, or specify a region of interest where the prototypes should focus. Such human-machine collaboration would improve the classifier's accuracy and interpretability, and would potentially reduce the amount of data needed to train the model. However, human-machine collaboration is rarely explored in the context of prototype learning, or case-based reasoning in general.
    
    \item \textbf{How to troubleshoot a trained prototype-based model to improve the quality of prototypes?} Prototype-based models make decisions based on similarities with learned prototypes. However, sometimes, the prototypes may not be learned well enough, in the sense that they may not capture the most representative features of a class. This is especially problematic for part-based prototype models, because these models reason by comparing subsets of features they deem important. In the case of a (part-based) prototype model with ``invalid'' prototypes that capture non-representative or undesirable features (e.g., a prototype of text in medical images, which represents information that improves training accuracy but not test accuracy), one way to ``fix'' the model is to get rid of the invalid prototypes, but this may lead to an imbalance in the number of prototypes among different classes and a bias for some classes that are over-represented with abundant prototypes. An alternative solution is to replace each undesirable prototype with a different prototype that does not involve the undesirable features. The challenge here lies in how we can replace undesirable prototypes systematically without harming the model performance, and at the same time improve the given model incrementally without retraining the model from scratch.
    %Zhi: how to troubleshoot prototype based methods -> find an invalid prototype, e.g. a text prototype in the medical imaging -> how to fix the model when we know this problem.
\end{enumerate}

\noindent \textbf{Example:} Suppose that a doctor wants to evaluate the risk of breast cancer among patients \citep[see][]{BarnettEtAl2021}. The dataset used to predict malignancy of breast cancer usually contains a set of mammograms, and a set of patient features (e.g., age). Given a particular patient, how do we characterize prototypical signs of cancerous growth from a sequence of mammograms taken at various times (this is similar to a video, see Challenge \ChCaseBased.1)? How can a doctor supervise prototype learning by telling a prototype-based model what (image/patient) features are typical of breast cancer (Challenge \ChCaseBased.2)? If a trained model contains a prototype of ``text'' in the mammograms (such text could be doctor's notes or patient IDs left on some training images), how can we replace the unwanted prototype with another prototype that is more medically relevant, without retraining the model from scratch (Challenge \ChCaseBased.3)?

%%%%%%%%%%%%%%%%%%%%%%%%%%%
\section{Complete supervised disentanglement of neural networks}\label{sec:superviseddis}

Deep neural networks (DNNs) have achieved state-of-the-art predictive performance for many important tasks \citep{lecun2015deep}. DNNs are the quintessential ``black box'' because the computations  within its hidden layers are typically inscrutable. 
% The interpretability of DNNs could be significantly improved if one can clearly identify what is learned by the neurons in their hidden layers, i.e. their latent space.
As a result, there have been many works that have attempted to ``disentangle'' DNNs in various ways so that information flow through the network is easier to understand. ``Disentanglement'' here refers to the way information travels through the network: we would perhaps prefer that all information about a specific concept (say ``lamps'') traverse through one part of the network  while information about another concept (e.g., ``airplane'') traverse through a separate part. Therefore, disentanglement is an interpretability constraint on the neurons. For example, suppose we hope neuron ``$\textrm{neur}$'' in layer $l$ is aligned with concept $c$, in which case, the disentanglement constraint is
\begin{equation}
    \text{DisentanglementConstraint}(c,\textrm{neur},l): \text{Signal}(c, \textrm{neur}) = \text{Signal}(c, l),\label{eqn:disentanglement}
\end{equation}
where $\text{Signal}(c,x)$ means the signal of concept $c$ that passes through $x$, which measures similarity between its two arguments. This constraint means \textit{all} signal about concept $c$ in layer $l$ will \textit{only} pass through neuron $\textrm{neur}$.
In other words, using these constraints, we could constrain our network to have the kind of ``Grandmother node'' that scientists have been searching for in both real and artificial convolutional neural networks  \citep{GrossGrandmother2002}.

In this challenge, we consider the possibility of fully disentangling a DNN so that each neuron in a piece of the network represents a human-interpretable concept.

% , an ideal interpretable latent space could be made of disjoint data generation factors. This is also known as a disentangled latent space \citep{bengio2009learning,higgins2018towards}.
% since humans attempt to understand and reason about many aspects of the world using concepts.
% \textcolor{red}{reorganize, use figure as the example to explain why it is interpretable}

\begin{figure}[ht]
    \centering
    \includegraphics[scale=0.5]{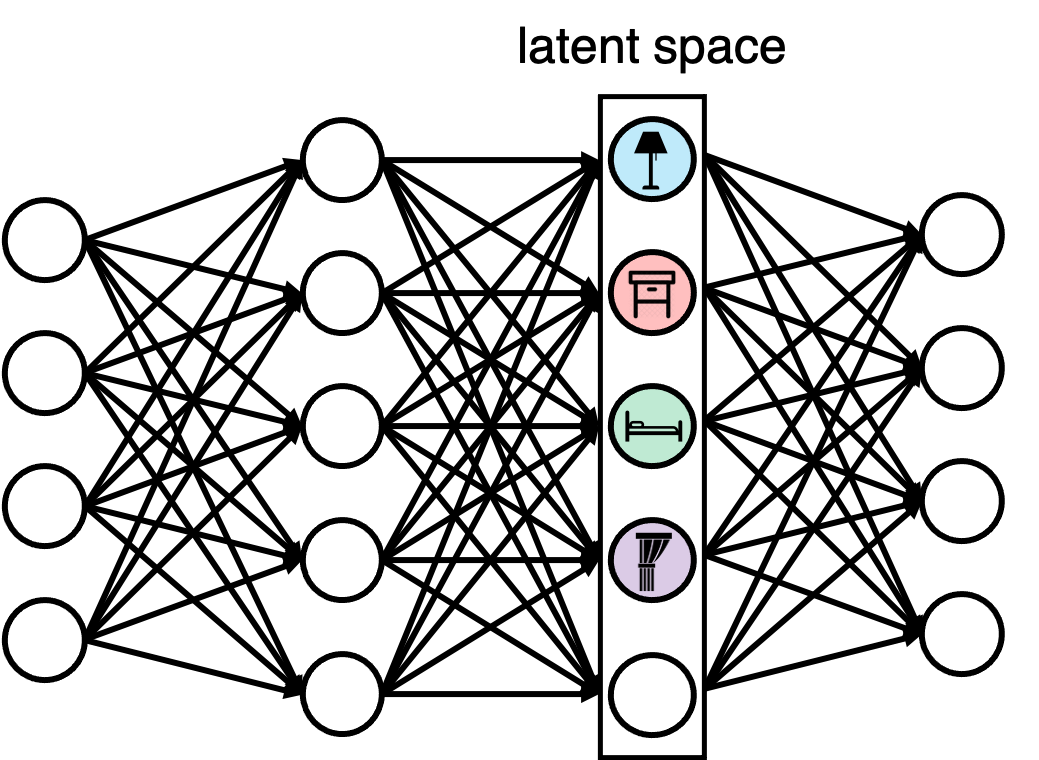}
    \caption{Disentangled latent space. The axes (neurons) of the latent space are aligned with supervised concepts, e.g. “lamp,” “bed,” “nightstand,” “curtain.” All information about the concept up to that point in the network travels through that concept’s corresponding neuron.}
    \label{fig:disentangled}
\end{figure}
The vector space whose axes are formed by activation values on the hidden layer's neurons is known as the latent space of a DNN.
Figure \ref{fig:disentangled} shows an example of what an ideal interpretable latent space might look like. The axes of the latent space are aligned with individual visual concepts, such as ``lamp,'' ``bed,'' ``nightstand,'' ``curtain.'' Note that the ``visual concepts'' are not restricted to objects but can also be things such as weather or materials in a scene. We hope all information about the concept travels through that concept’s corresponding neuron on the way to the final prediction. For example, the ``lamp'' neuron will be activated if and only if the network thinks that the input image contains information about lamps. This kind of representation makes the reasoning process of the DNN much easier to understand: the image is classified as ``bedroom'' because it contains information about ``bed'' and ``lamp.''

Such a latent space, made of disjoint data generation factors, is a disentangled latent space \citep{bengio2009learning,higgins2018towards}. An easy way to create a disentangled latent space is just to create a classifier for each concept (e.g., create a lamp classifier), but this is not a good strategy: it might be that the network only requires the light of the lamp rather than the actual lamp body, so creating a lamp classifier could actually reduce performance. Instead, we would want to encourage the information that is used about a concept to go along one path through the network.

\begin{figure}[ht]
    \centering
    \includegraphics[scale=0.5]{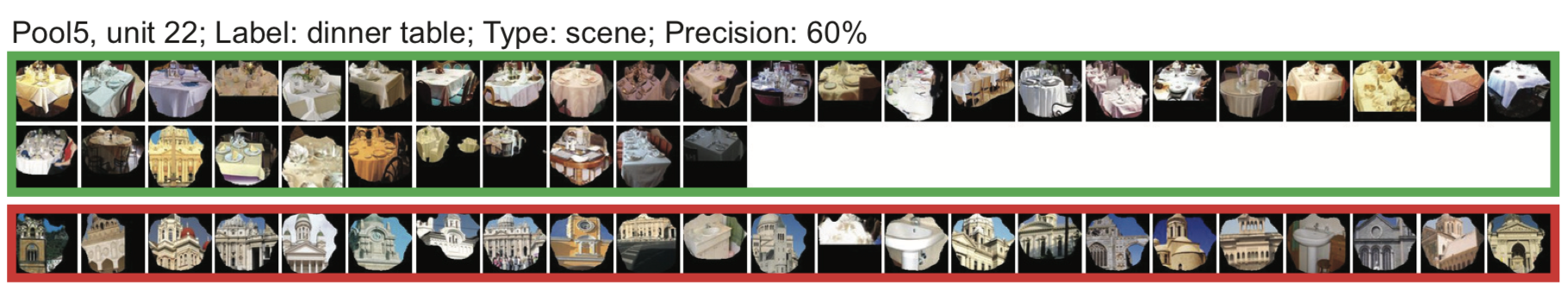}
    \caption{Example of an impure neuron in standard neural network from \citet{zhou2014object}. This figure shows images that highly activate this neuron. Both dining tables (green) and Greek-style buildings (red) are highly activated on the neuron, even though these two concepts are unrelated.}
    \label{fig:example_impure}
\end{figure}

Disentanglement is not guaranteed in standard neural networks. In fact, information about any concept could be scattered throughout the latent space of a standard DNN. For example, post hoc analyses on neurons of standard convolutional neural networks \citep{zhou2018interpreting,zhou2014object} show that concepts that are completely unrelated could be activated on the same axis, as shown in Figure \ref{fig:example_impure}. Even if we create a vector in the latent space that is aimed towards a single concept \citep[as is done in][]{kim2018interpretability,zhou2018interpretable}, that vector could activate highly on multiple concepts, which means the signal for the two concepts is not  disentangled. In that sense, vectors in the latent space are ``impure'' in that they do not naturally represent single concepts \citep[see][for a detailed discussion]{chen2020concept}.
% \textcolor{red}{add more explanations}

 This challenge focuses on supervised disentanglement of neural networks, i.e., the researcher specifies which concepts to disentangle in the latent space. (In the next section we will discuss unsupervised disentanglement.) 
 %is frequently mentioned in the deep generative models as a high-level goal of unsupervised learning \citep{chen2016infogan,higgins2016beta}, which discover the disentangled factors to generate data without supervision. We will talk about them in the next section. 
 Earlier work in this domain disentangles the latent space for specific applications, such as face recognition, where we might aim to separate identity and pose \citep{zhu2014multi}. Recent work in this area aims to disentangle the latent space with respect to a collection of predefined concepts \citep{chen2020concept,koh2020concept,losch2019interpretability,adel2018discovering}. For example, \citet{chen2020concept} adds constraints to the latent space to decorrelate the concepts and align them with axes in the latent space; this is called ``concept whitening.'' (One could think of this as analogous to a type of principal components analysis for neural networks.) To define the concepts, a separate dataset is labeled with concept information (or the main training dataset could also serve as this separate dataset as long as it has labels for each concept.) These concept datasets are used to align the network's axes along the concepts. This type of method can create disentangled latent spaces where concepts are more ``pure'' than those of a standard DNN, without hurting accuracy. With the disentangled latent space, one can answer questions about how the network gradually learns concepts over layers \citep{chen2020concept}, or one could interact with the model by intervening on the activations of concept neurons \citep{koh2020concept}. Many challenges still exist for supervised disentanglement of DNNs, including:

\begin{enumerate}[\thesection .1]
    \item \textbf{How to make a whole layer of a DNN interpretable?} Current methods can make neurons in a single layer interpretable but all of them have limitations. \cite{koh2020concept}, \cite{losch2019interpretability} and \cite{adel2018discovering} try to directly learn a concept classifier in the latent space. 
    % Training of the concepts in these methods is easy since one can build a concept loader very similar to the data loader of the main dataset. 
    However, as discussed by \cite{chen2020concept}, good discriminative power on concepts does not guarantee concept separation and disentanglement. That is, we could easily have our network classify all the concepts correctly, in addition to the overall classification task, but this would not mean that the information flow concerning each concept goes only though that concept's designated path through the network; the activations of the different concept nodes would likely still be correlated. The method discussed earlier, namely that of 
    \citet{chen2020concept}, can fully disentangle the latent space with respect to a few pre-defined concepts. But it also has two disadvantages if we want to disentangle all neurons in a layer, namely: (i) this method currently has a group of unconstrained neurons to handle residual information, which means it does not disentangle all neurons in a layer, just some of them, and
    (ii) it requires loading samples of all target concepts to train the latent space. If the layer to be disentangled contains 500 neurons, we would need to load samples for all 500 concepts, which takes a long time. 
   % It might be possible to combine the advantages of several of these methods, i.e., to develop a method capable of disentangling original layers in the DNN while all neurons in that layer are aligned with pre-defined concepts. This is challenging because (a) removing the residual information that could be carried by unconstrained neurons might sacrifice accuracy; (b) loading data from many auxiliary concept datasets takes a long time: regular convolutional layers usually have (at least) hundreds of filters, thus one would need to load samples of hundreds of concepts at the same time. Perhaps this 
    Perhaps the latter problem could be solved by training with a small random subset of concepts at each iteration; but this has not been tried in current methods at the time of this writing. The first problem might be solved by using an unsupervised concept detector from Challenge \ChUnsupDis{} and having a human interact with the concepts to determine whether each one is interpretable.
    
    \item  \textbf{How to disentangle all neurons of a DNN simultaneously?} Current methods try to disentangle at most a single layer in the DNN (that is, they attempt only the problem discussed above). This means one could only interpret neurons in that specific layer, while semantic meanings of neurons in other layers remain unknown. Ideally, we want to be able to completely understand and modify the information flowing through all neurons in the network. This is a challenging task for many obvious reasons, the first one being that it is hard practically to define what all these concepts could possibly be. We would need a comprehensive set of human-interpretable concepts, which is hard to locate, create, or even to parameterize.
    %First, state-of-the-art neural network architectures contain an enormous amount of neurons. 
    Even if we had this complete set,
    we would probably not want to manually specify exactly what part of the network would be disentangled with respect to each of these numerous concepts. For instance,
    if we tried to disentangle the same set of concepts in all layers, it would be immediately problematic because
    %It is hard practically to pre-specify enough concepts to be aligned with these neurons. One may use the same set of concepts in all layers to reduce the quantity of new definitions. 
    %However, learning the same set of concepts in all layers is problematic since complicated concepts, like objects and weather of the scene, cannot be learned in earlier layers of DNN. Second, even if we can find enough concepts, placing them in the correct layer is also challenging. 
    DNNs are naturally hierarchical: high-level concepts (objects, weather, etc.) are learned in deeper layers, and the deeper layers leverage low-level concepts (color, texture, object parts, etc.) learned in lower layers. Clearly, complex concepts like ``weather outside" could not be learned well in earlier layers of the network, so higher-level concepts might be reserved for deeper layers. 
    Hence, we also need to know the hierarchy of the concepts to place them in the correct layer. 
    Defining the concept hierarchy manually is almost impossible, since there could be thousands of concepts. But how to automate it is also a challenge.

    \item \textbf{How to choose good concepts to learn for 
    disentanglement?} In supervised disentanglement, the concepts are chosen manually. To gain useful insights from the model, we need good concepts. But what are good concepts in specific application domains? For example, in medical applications, past works mostly use clinical attributes that already exist in the datasets. However, \cite{chen2020concept} found that attributes in the ISIC dataset might be missing the key concept used by the model to classify lesion malignancy. Active learning approaches could be incredibly helpful in interfacing with domain experts to create and refine concepts.
    
    Moreover, it is challenging to learn concepts with continuous values. These concepts might be important in specific applications, e.g., age of the patient and size of tumors in medical applications. Current methods either define a concept by using a set of representative samples or treat the concept as a binary variable, where both are discrete. Therefore, for continuous concepts, a challenge is how to choose good thresholds to transform the continuous concept into one or multiple binary variables.
    \item \textbf{How to make the mapping from the disentangled layer to the output layer interpretable?} The decision process of current disentangled neural networks contains two parts, $x\rightarrow c$ mapping the input $x$ to the disentangled representation (concepts) $c$, and $c\rightarrow y$ mapping the disentangled representation $c$ to the output $y$. \citep[The notations are adopted from][]{koh2020concept}. All current methods on neural disentanglement aim at making the $c$ interpretable, i.e., making the neurons in the latent space aligned with human understandable concepts, but how these concepts combined to make the final prediction, i.e., $c\rightarrow y$, often remains a black box. This leaves a gap between the interpretability of the latent space and the interpretability of the entire model. Current methods either rely on variable importance methods to explain $c\rightarrow y$ posthoc \citep{chen2020concept}, or simply make $c\rightarrow y$ a linear layer \citep{koh2020concept}. However, a linear layer might not be expressive enough to learn $c\rightarrow y$. \cite{koh2020concept} also shows that a linear function $c\rightarrow y$ is less effective than nonlinear counterparts when the user wants to intervene in developing the disentangled representation, e.g., replacing predicted concept values $\hat{c}_j$ with true concept values $c_j$. Neural networks like neural additive models \citep{agarwal2020neural} and neural decision trees \citep{yang2018deep} could potentially be used to model $c\rightarrow y$, since they are both differentiable, nonlinear, and intrinsically interpretable once the input features are interpretable. %However, this idea hasn't been studied yet.

\end{enumerate}

\noindent \textbf{Example:} Suppose machine learning practitioners and doctors want to build a supervised disentangled DNN on X-ray data to detect and predict arthritis. %They may encounter several challenges. 
First, they aim to choose a set of relevant concepts that have been assessed by doctors (Challenge \ChSupDis.3). They should also choose thresholds to turn continuous concepts (e.g., age) into binary variables to create the concept datasets (Challenge \ChSupDis.3). Using the concept datasets, they can use supervised disentanglement methods like concept whitening \citep{chen2020concept} to build a disentangled DNN. However, if they choose too many concepts to disentangle in the neural network, loading samples from all of the concept datasets may take a very long time (Challenge \ChSupDis.1). Moreover, doctors may have chosen different levels of concepts, such as bone spur (high-level) and shape of joint (low-level), and they would like the low-level concepts to be disentangled by neurons in earlier layers, and high-level concepts to be disentangled in deeper layers, since these concepts have a hierarchy according to medical knowledge. However, current methods only allow placing them in the same layer (Challenge \ChSupDis.2). Finally, all previous steps can only make neurons in the DNN latent space aligned with medical concepts, while the way in which these concepts combine to predict arthritis remains uninterpretable (Challenge \ChSupDis.4).

\section{Unsupervised disentanglement of neural networks}\label{sec:unsupdis}
\begin{figure}[ht]
    \centering
    \includegraphics[scale=0.4]{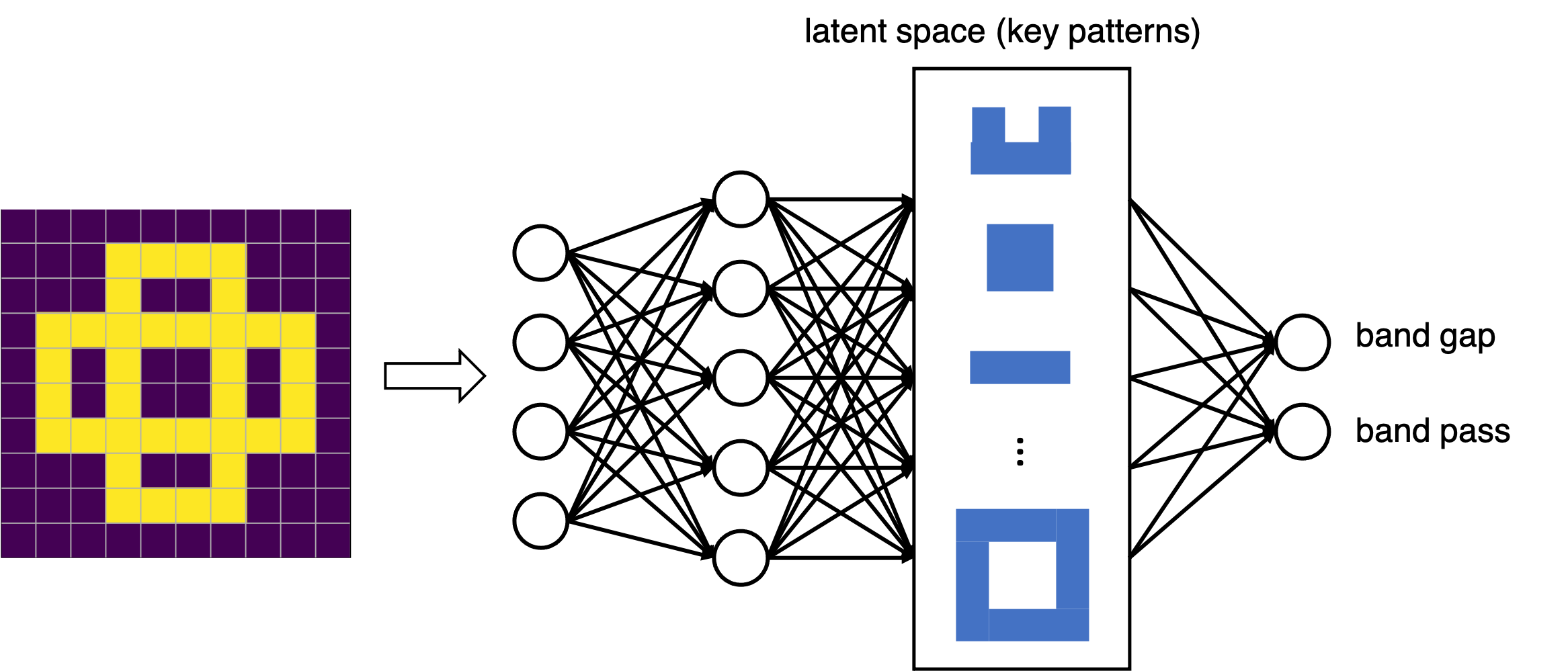}
    \caption{Neural network with unsupervised disentanglement in the latent space. The input (on the left) is a unit cell of metameterial, made of stiff (yellow) and soft (purple) constituent materials. The target of the neural network is predicting whether the material with unit cell on the left supports the formation of forbidden frequency bands of propagation, i.e. existence of ``band gaps.'' Key patterns related to the band gap (illustrated in blue) have been discovered without supervision. Neurons in the latent space are aligned with these patterns. Figure adapted from \citet{chenmetamine}. }
    \label{fig:unsupdisentangled}
\end{figure}

% we want to do what we did in section 4 but in many cases we don't know the concepts... e.g. material science
The major motivation of unsupervised disentanglement is the same as Challenge \ChSupDis, i.e., making information flow through the network easier to understand and interact with. But in the case where we do not know the concepts, or in the case where the concepts are numerous and we do not know how to parameterize them, we cannot use the techniques from Challenge \ChSupDis. In other words, the concept $c$ in Constraint (\ref{eqn:disentanglement}) is no longer a concept we predefine, but it must still be an actual concept in the existing universe of concepts. There are situations where concepts are actually unknown; for example, in materials science, the concepts, such as key geometric patterns in the unit cells of materials, have generally not been previously defined. There are also situations where concepts are generally known but too numerous to handle; a key example is computer vision for natural images. Even though we could put a name to many concepts that exist in natural scenes, labeled datasets for computer vision have a severe labeling bias: we tend only to label entities in images that are useful for a specific task (e.g., object detection), thus ignoring much of the information found in images. If we could effectively perform unsupervised disentanglement, we can rectify problems caused by human bias, and potentially make scientific discoveries in uncharted domains. For example, an unsupervised disentangled neural network can be used to discover key patterns in materials and characterize their relation to the physical properties of the material (e.g., ``will the material allow light to pass through it?''). Figure \ref{fig:unsupdisentangled} shows such a neural network, with a latent space completely disentangled and aligned with the key patterns discovered without supervision: in the latent space, each neuron corresponds to a key pattern and all information about the pattern flows through the corresponding neuron. Analyzing these patterns' contribution to the prediction of a desired physical property could help material scientists understand what correlates with the physical properties and could provide insight into the design of new materials.

Multiple branches of works are related to unsupervised disentanglement of deep neural networks, and we will describe some of them.

\textit{Disentangled representations in deep generative models:} 
Disentanglement of generative models have long been studied \citep{schmidhuber1992learning,desjardins2012disentangling}. Deep generative models such as generative adversarial networks \citep[GANs,][]{goodfellow2014gan} and variational auto-encoders \citep[VAE,][]{kingma2013auto} try to learn a mapping from points in one probability  distribution to points in another. The first distribution is over points in the latent space, and these points are chosen i.i.d$.$ according to a zero-centered Gaussian distribution. The second distribution is over the space from which the data are drawn (e.g., random natural images in the space of natural images). 
% This idea is very natural because most deep generative models such as generative adversarial networks \citep[GANs, ][]{goodfellow2014gan} and variational auto-encoders  \citep[VAE][]{kingma2013auto} ensure the latent features follow independent distributions (usually i.i.d. normal).
The statistical independence between the latent features makes it easy to disentangle the representation \citep{bengio2013representation}: a disentangled representation simply guarantees that knowing or changing one latent feature (and its corresponding concept) does not affect the distribution of any other. Results on simple imagery datasets show that these generative models can decompose the data generation process into disjoint factors (for instance, age, pose, identity of a person) and explicitly represent them in the latent space, without any supervision on these factors; this happens based purely on statistical independence of these factors in the data. Recently, the quality of disentanglement in deep generative models has been improved \citep{chen2016infogan,higgins2016beta}. These methods achieve full disentanglement without supervision by maximizing the mutual information between latent variables and the observations. However, these methods only work for relatively simple imagery data, such as faces or single 3D objects (that is, in the image, there is only one object, not a whole scene). Learning the decomposition of a scene into groups of objects in the latent space, for example, is not yet achievable by these deep generative models. One reason for the failure of these approaches might be that the occurrence of objects may not be statistically independent; for example, some objects such as ``bed'' and ``lamp'' tend to co-occur in the same scene. Also, the same type of object may occur multiple times in the scene, which cannot be easily encoded by a single continuous latent feature.
% Neural network disentanglement has long been studied under the notion of \textit{disentangled representation} \citep{bengio2013representation, higgins2018towards} in generative models, meaning decomposing the data generation process into disjoint factors (for instance, age, pose, identity of a person). Such factors could be also interpreted as abstract concepts in the data. The idea of learning disjoint factors that control data generation has been studied in both supervised learning \citep{tran2017disentangled,reed2014learning,tenenbaum2000separating,zhu2014multi} and unsupervised learning \citep{schmidhuber1992learning,desjardins2012disentangling}. Recently, our ability to disentangle in deep neural networks has improved, due to advances in deep generative models \citep{chen2016infogan,higgins2016beta}. These methods achieve  full disentanglement without supervision by maximizing the mutual information between latent variables and the observations. However, these methods only work for relatively simple imagery data, such as faces or single 3D objects (in the image, there is only one object). Learning the decomposition of a scene into groups of objects in the latent space, for example, is not achievable by these deep generative models.

\begin{figure}[ht]
    \centering
    \includegraphics[scale=0.5]{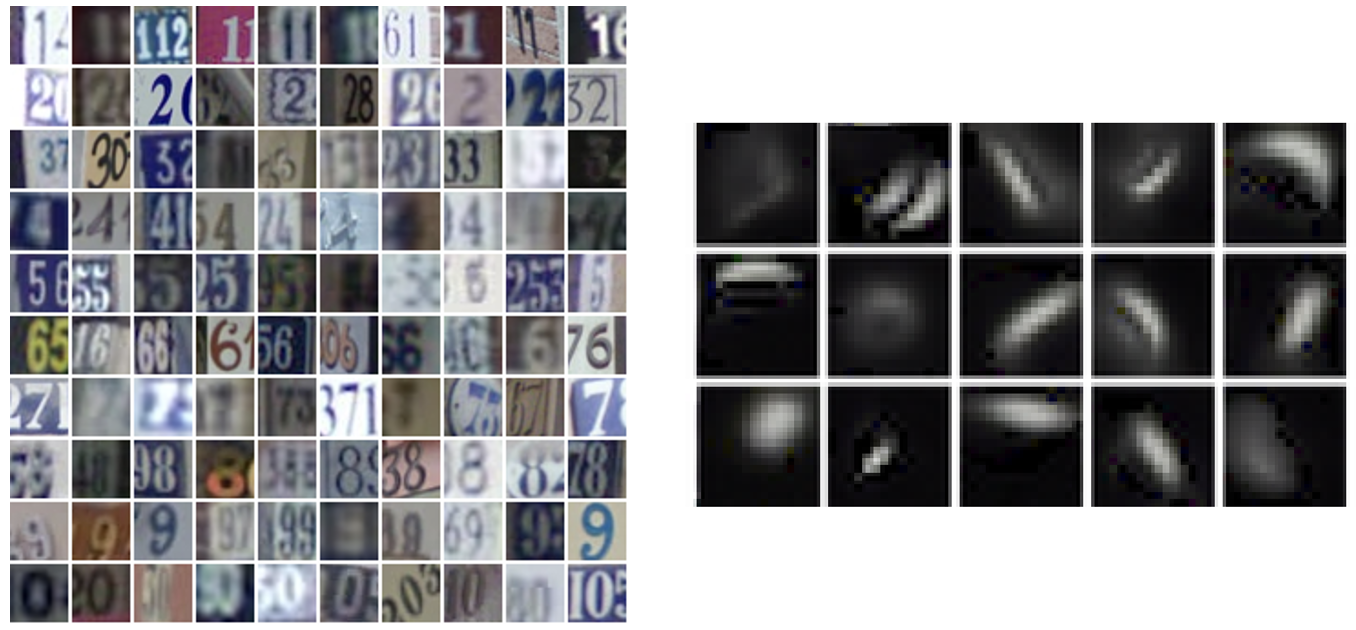}
    \caption{Left: Sample Images collected from the SVHN dataset \citep{svhn}. Right: Stroke-like templates discovered by the SCAE. Although the objects in the dataset are mostly digits, current capsule networks are unable to discover them as concepts without supervision.}
    \label{fig:svhn_scae}
\end{figure}

\textit{Neural networks that incorporate compositional inductive bias:} 
Another line of work designs neural networks that directly build compositional structure into a neural architecture. Compositional structure occurs naturally in computer vision data, as objects in the natural world are made of parts. In areas that have been widely studied beyond computer vision, including speech recognition, researchers have already summarized a series of compositional hypotheses, and incorporated them into machine learning frameworks. For example, in computer vision, the ``vision as inverse graphics'' paradigm \citep{baumgart1974geometric} tries to treat vision tasks as the inverse of the computer graphics rendering process. In other words, it tries to decode images into a combination of features that might control rendering of a scene, such as object position, orientation, texture and lighting. Many studies on unsupervised disentanglement have been focused on creating this type of representation because it is intrinsically disentangled.
%This type of representation of the data is already disentangled without requiring further preprocessing, therefore has attracted attention from the researchers studying unsupervised disentanglement.
Early approaches toward this goal includes DC-IGN \citep{kulkarni2015deep} and Spatial Transformers \citep{jaderberg2015spatial}. Recently, Capsule Networks \citep{hinton2011transforming,sabour2017dynamic} have provided a new way to incorporate compositional assumptions into neural networks. Instead of using neurons as the building blocks, Capsule Networks combine sets of neurons into larger units called ``Capsules,’’ and force them to represent information such as pose, color and location of either a particular part or a complete object. This method was later combined with generative models in the Stack Capsule Autoencoder (SCAE) \citep{kosiorek2019stacked}. With the help of the Set Transformer \citep{lee2019set} in combining information between layers, SCAE discovers constituents of the image and organizes them into a smaller set of objects. 
Slot Attention modules \citep{locatello2020object} further use an iterative attention mechanism to control information flow between layers, and achieve better results on unsupervised object discovery. Nevertheless, similar to the generative models, these networks perform poorly when aiming to discover concepts on more realistic datasets. For example, SCAE can only discover stroke-like structures that are uninterpretable to humans on the Street View House Numbers (SVHN) dataset \citep{svhn} (see Figure~\ref{fig:svhn_scae}). The reason is that the SCAE can only discover visual structures that appear frequently in the dataset, but in reality the appearance of objects that belong to the same category can vary a lot. There have been proposals \citep[e.g., GLOM][]{ hinton2021represent} on how a neural network with a fixed architecture could potentially parse an image into a part-whole hierarchy. Although the idea seems to be promising, no working system has been developed yet. There is still a lot of room for development for this type of method.

\textit{Works that perform unsupervised disentanglement implicitly}: Many other interpretable neural networks can also learn disjoint concepts in the latent space although the idea of ``disentanglement'' is not explicitly mentioned in these papers. For example, as mentioned in Section \ref{sec:case}, \cite{chen2019looks} propose to create a protopype layer, storing prototypical parts of training images, to do case based reasoning. When classifying birds, the prototypical parts are usually object parts such as head and wings of birds. Interestingly, a separation cost is applied to the  learned prototypes to encourage diversity of the prototypical parts, which is very similar to the idea of disentanglement. 
% \cite{zhang2018interpretable} proposes to design a mutual information loss to push a filter in high conv-layers toward the representation of an object part without supervision. Here, disentanglement is encouraged by forcing the filter to be exclusively activated in a single region of the feature map. 
\cite{zhang2018interpretable} propose to maximize mutual information between the output of convolution filters and predefined part templates (feature masks with positive values in a localized region and negative in other places). The mutual information regularization term is essentially the sum of two disentanglement losses, (a) an inter-category entropy loss that encourages each filter to be exclusively activated by images of one category and not activated on other categories; (b) a spatial entropy loss that encourages each filter to be activated only on a local region of the image. Results show the convolutional filters with the mutual information loss tend to be activated only on a specific part of the object. These two methods are capable of learning single objects or parts in the latent space but have not yet been generalized to handle more comprehensive concepts (such as properties of scenes, including style of an indoor room - e.g., cozy, modern, etc., weather for outdoor scenes, etc.), since these concepts are not localized in the images.

Despite multiple branches of related work targeting concept discovery in different way, many challenges still exist:
\begin{enumerate}[\thesection .1]
    \item \textbf{How to quantitatively evaluate unsupervised disentanglement?}
    
    Recall that unsupervised disentanglement is desired (but challenging) in two different domains: (a) domains in which we do not know what the concepts are (e.g., materials science); (b) domains in which concepts are known but labeling biases exist. 
    
    Let us start with a version of case (b) where some concepts are known, such as objects in natural images. Let us say we are just missing a subset of concept labels in the dataset, which is a simple type of missing data bias. If we build a neural network that disentangles the latent space, we can quantitatively evaluate the quality of disentanglement. 
    For instance, let us say we know a collection of concepts that we would like disentangled (say, a collection of specific objects that appear in natural images). If we use an unsupervised algorithm for disentangling the space, and then evaluate whether it indeed disentangled the known concepts, then we have a useful quantitative evaluation
    %with known concepts in the images, we can easily evaluate it using true labels of these concepts in the datasets. Even if the network learns concepts without supervision, we can at least evaluate the disentanglement by measuring alignment between the discovered concepts and existing annotations in imagery datasets
    \citep{higgins2016beta,chenisolating,kim2018disentangling,eastwood2018framework,do2019theory}.

    If we are working in a domain where we do not know the concepts to disentangle (case (a)), we also might not know what regularization (or other type of inductive bias) to add to the algorithm so that it can discover concepts that we would find interpretable. In that case, it becomes difficult to evaluate the results quantitatively.
    %If we are working in a domain where we do not know the concepts, we also may not know the best inductive biases. 
    % First, for domain with unknown concepts, they would not (in general) require the same types of inductive biases as the domains that we could generate labels for. 
    %Not knowing the useful inductive biases can cause a problem for evaluation. 
    % Previous neural networks using inductive biases of images may not work for other fields like materials science. 
    % The inductive biases would be different in these fields: for images, we know scenes consist of objects and the objects interact with each other, whereas in materials we are looking for key shapes or patterns that have no formal name or relationship.
    Materials science, as discussed above, is an example of one of these domains, where we cannot find ground-truth labels of key patterns, nor can we send queries to human evaluators (even material scientists do not know the key patterns in many cases). Evaluation metrics from natural images do not work. Evaluating disentanglement in these domains is thus a challenge.
    %might require a domain specific definition of disentanglement.
    %and a quantitative metric based on its definition.
    
    % Even for imagery data, relying only on human annotations to evaluate disentanglement might be problematic. This is because supervised annotations may suffer from human bias in what is labeled, and could be misleading. 
    
    Going back to case (b), in situations where labeling biases exist (i.e., when only some concepts are labeled, and the set of labels are biased), current evaluation metrics of disentanglement that rely only on human annotations can be problematic. For instance, most annotations in current imagery datasets concern only objects in images but ignore useful information such as lighting and style of furniture. An unsupervised neural network may discover important types of concepts that do not exist in the annotations. Medical images may be an important domain here, where some concepts are clearly known to radiologists, but where labels in available datasets are extremely limited. 

    \item \textbf{How to adapt neural network architectures designed with compositional constraints to other domains?}
    
    Incorporating compositional assumptions into network architecture design is a way to create intrinsically interpretable disentangled representations. That said, such assumptions are usually modality- and task-specific, severely limiting the general applicability for such designs. Let us take Capsule Networks \citep{sabour2017dynamic} and the Slot Attention module \citep{locatello2020object} in the computer vision field as examples: these modules try to create object-centric and part-centric representations inside the latent space of the network. These ideas are based on the assumption that an image can be understood as a composition of different objects, and the interference between objects is negligible. Nevertheless, such an assumption cannot necessarily be applied to materials science, in which a material's physical properties could depend jointly, in a complex way, on the patterns of constituent materials within it.  
    %be applied to natural language processing tasks, in which the meaning of individual words is more often affected by the context. %The object-based representation also creates obstacles for applying the design to other computer vision tasks, owing to the fact that it ignores information such as commonalities between objects, which is important for style-transfer. 
    How would we redefine the compositional constraints derived in computer vision for natural images to work for other domains such as materials science?
    %How to improve generalizability of network architectures designed with combinatorial constraints?
    % Can we improve the generalizability of this type of network structure?  
    
    \item \textbf{How to learn part-whole disentanglement for more complicated patterns in large vision datasets?}
    A specific area within unsupervised disentanglement in computer vision is to semantically segment the image into different parts, where each part represents an object or a part of an object \citep{crawford2019spatially,sabour2017dynamic,kosiorek2019stacked,locatello2020object}. Current networks can achieve convincing results over simple, synthetic datasets (where objects consist of simple geometric shapes or digits). (Such datasets include CLEVR6, \citealt{johnson2017clevr}, d-Sprite, \citealt{dsprites17}, and Objects Room and Tetrominoes, \citealt{multiobjectdatasets19}.) However, no work has successfully learned part-whole relationships on more realistic datasets such as ImageNet. New techniques may be needed to handle various interactions between different objects in a real-world dataset.
    
    % \item \textbf{How to achieve disentanglement by weak supervision?}  Weakly supervised information? VQA? general principle, common knowledge about concepts/data generation factors in the world

\end{enumerate}

\noindent\textbf{Example:} 
Suppose material scientists want to build a classification model that can predict whether the designs of metameterials support existence of band gaps (same example as Figure \ref{fig:unsupdisentangled}). Because unit cells of metamaterials are usually represented as a matrix/tensor of what constituent materials are placed at each location, the researchers plan to build a deep neural network, since neural networks excel at extracting useful information from raw inputs. 
% They hope the neural network can not only predict band gap existence but also disentangle the key patterns in the latent space. Knowing these key patterns can help them understand the formation of band gaps and provide insights in the design of new materials.
They might encounter several challenges when building the disentangled neural network. First, they need to identify the disentanglement constraints to build into the network architecture. Architectures that work well for imagery data may not work for unit cells of metamaterials, since key patterns in unit cells can be completely different from objects/concepts in imagery data (Challenge \ChUnsupDis.2 above). Moreover, evaluation of disentanglement can be challenging as the material patterns have no ground truth labels (Challenge \ChUnsupDis.1 above).

% Suppose a self-driving car company wants to build an vision system for their cars based on deep neural networks. They hope the system could be interpretable which can help to detect the system malfunctions causing a crash. Therefore, they plan to build a disentangled neural network but the disentanglement has to be unsupervised since the real world contains too many concepts to be labeled. They might encounter the following challenges when building the vision system. First, they need to identify what is the disentanglement constraint they need for their vision system (Challenge \ChScoring above). Part-whole disentanglement might be a good choice for the constraint, as scenes on the road can be decomposed into objects (e.g., cars and traffic signs), and the objects can be decomposed into parts (e.g., wheels). Unfortunately, current part-whole disentanglement methods like Capsule Net cannot deal with large real-world datasets (Challenge 3 above). After building the vision system, they need to evaluate the performance of disentanglement. But without true labels, quantitative evaluation can be very hard (Challenge 1 above).

\section{Dimension Reduction for Data Visualization}\label{sec:DR}
Even in data science, a picture is worth a thousand words. Dimension reduction (DR) techniques take, as input, high-dimensional data and project it down to a lower-dimensional space (usually 2D or 3D) so that a human can better comprehend it. 
Data visualization can provide an intuitive understanding of the underlying structure of the dataset. DR can help us gain insight and build hypotheses. DR can help us design features so that we can build an interpretable supervised model, allowing us to work with high-dimensional data in a way we would not otherwise be able to. With DR, biases or pervasive noise in the data may be illuminated, allowing us to be better data caretakers. However, with the wrong DR method, information about the high-dimensional relationships between points can be lost when projecting onto a 2D or 3D space. 

DR methods are unsupervised ML methods that are constrained to be interpretable. Referring back to our generic interpretable ML formulation \eqref{eqn:generic}, DR methods produce a function mapping data $\x_1,
\x_2, ..., \x_n$ from $p$-dimensional space to $\mathbf{y}_1$, $\mathbf{y}_2$, ... $\mathbf{y}_n$ in a low dimensional space (usually 2D). The constraint of mapping to 2 dimensions is an interpretability constraint. DR methods typically have a loss function that aims to preserve a combination of distance information between points and neighborhood information around each point when projecting to 2D. Each DR algorithm chooses the loss function differently. Each algorithm also chooses its own distance or neighborhood information to preserve.

\begin{figure}[th]
    \centering
    \includegraphics[scale=0.4]{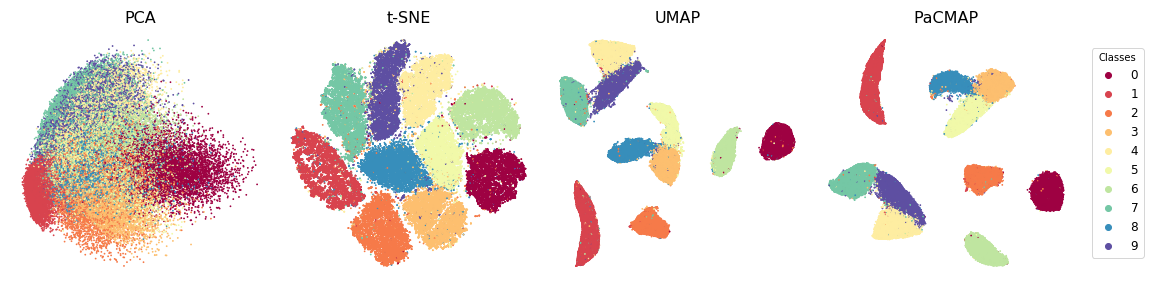}
    \caption{Visualization of the MNIST dataset \citep{lecun2010mnist} using different kinds of DR methods: PCA \citep{Pearson01}, t-SNE \citep{vandermatten08,Linderman19FItSNE,openTSNE}, UMAP \citep{UMAP}, and PaCMAP \citep{pacmap}. The axes are not quantified because these are projections into an abstract 2D space.}
    \label{fig:mnist_dr}
\end{figure}

Generally speaking, there are two primary types of approaches to DR for visualization, commonly referred to as local and global methods. Global methods aim mainly to preserve distances between any pair of points (rather than neighborhoods), while the local methods emphasize preservation of local neighborhoods (that is, which points are nearest neighbors). As a result, local methods can preserve the local cluster structure better, while failing to preserve the overall layout of clusters in the space, and vice versa. Figure~\ref{fig:mnist_dr} demonstrates the difference between the two kinds of algorithms over the MNIST handwritten figure dataset \citep{lecun2010mnist}, which is a dataset where local structure tends to be more important than global structure. The only global method here, PCA \citep{Pearson01}, fails to separate different digits into clusters, but it gives a sense of which digits are different from each other. 
%For example, all the 1s are quite similar, and 0 is a different digit compared to all the others. 
t-SNE \citep{vandermatten08}, which is a local method, successfully separated all the digits, but could not also keep the scale information that is preserved in the PCA embedding. More recent methods, such as UMAP \citep{UMAP} and PaCMAP \citep{pacmap} also separated the digits while preserving some of the global information.

Early approaches toward this problem, including Principal Component Analysis (PCA) \citep{Pearson01} and Multidimensional Scaling (MDS) \citep{Torgerson52}, mostly fall into the global category. They aim to preserve as much information as possible from the high-dimensional space, including the distances or rank information between pairs of points. These methods usually apply matrix decomposition over the data or pairwise distance matrix, and are widely used for data preprocessing. These methods usually fail to preserve local structure, including cluster structure.
%However, their stronger focus over the global structure makes them fail to elucidate the local structure in visualization, such as rendering out cluster structures.

To solve these problems, researchers later (early 2000s) developed methods that aimed at local structure, because they had knowledge that high dimensional data lie along low-dimensional manifolds. Here, it was important to preserve the local information along the manifolds. Isomap \citep{Tenenbaum00}, Local Linear Embedding (LLE) \citep{Roweis00}, Hessian Local Linear Embedding \citep{Donoho03}, and Laplacian Eigenmaps \citep{Belkin01} all try to preserve exact local Euclidean distances from the original space when creating low-dimensional embeddings. But distances between points behave differently in high dimensions than in low dimensions, leading to problems preserving the distances. In particular, these methods tended to exhibit what is called the ``crowding problem,'' where samples in the low-dimensional space crowd together, devastating the local neighborhood structure and losing information. 
t-SNE \citep{vandermatten08} was able to handle this problem by transforming the high-dimensional distances between each pair of points into probabilities of whether the two points should be neighbors in the low-dimensional space. Doing this aims to ensure that local neighborhood structure is preserved. Then, during the projection to low dimensions, t-SNE enforces that the distribution over distances between points in the low-dimensional space is
%conditional probabilities over distances 
a specific transformation of the distribution over distances between points in the high-dimensional space. Forcing distances in the low-dimensional space to follow a specific distribution avoids the crowding problem, which is when a large proportion of the distances are almost zero.
%With a long-tailed t-distribution of distances in the low-dimensional space, t-SNE can separate samples in the low-dimensional space, therefore avoid the ``crowding problem.''

\begin{figure}[th]
    \centering
    \includegraphics[scale=0.4]{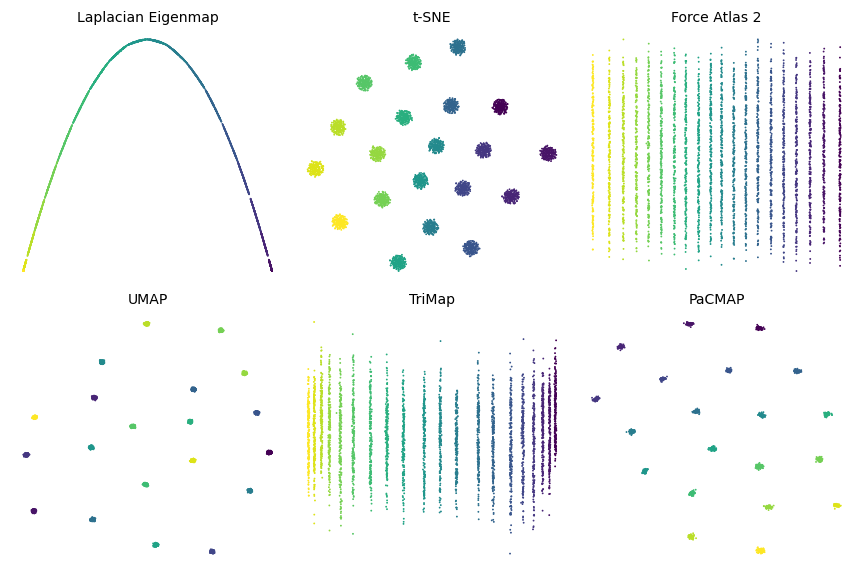}
    \caption{Visualization of 4000 points sampled from 20 isotropic Gaussians using Laplacian Eigenmap \citep{Belkin01, sklearn}, t-SNE \citep{vandermatten08,Linderman19FItSNE,openTSNE}, ForceAtlas2 \citep{ForceAtlas2}, UMAP \citep{UMAP}, PaCMAP \citep{pacmap} and TriMap \citep{2019TRIMAP}. The 20 Gaussians are equally spaced on an axis in 50-dimensional space, labelled by the gradient colors. The best results are arguably those of t-SNE and PaCMAP in this figure, which preserve clusters compactly and their relative placement (yellow on the left, purple on the right).}
    \label{fig:gaussian_dr}
\end{figure}

Though more successful than previous methods, t-SNE still suffers from slow running times, sensitivity to hyperparameter changes, and lack of preservation of global structure. A series of t-SNE variants aim to improve on these shortcomings.
The most famous among them are BH t-SNE \citep{Maaten2014AcceleratingTU} and FIt-SNE  \citep{Linderman19FItSNE}. BH t-SNE constructs a $k$-nearest neighbor graph over the high dimensional space to record local neighborhood structure, and utilizes the Barnes-Hut force calculation algorithm \citep[which is typically used for multi-body simulation; ][]{barnes1986hierarchical} to accelerate optimization of the low-dimensional embedding. These choices reduced the complexity for each step in the optimization from $O(n^2)$, where $n$ is the number of samples in the dataset, to a smaller $O(n\log n)$. FIt-SNE further accelerates the rendering phase with a Fast Fourier Transform, and brings down the complexity to $O(n)$. Besides these variants, multiple new algorithms have been created based on the framework of t-SNE. Prominent examples include LargeVis \citep{Tang16}, which is widely used in network analysis, and UMAP \citep{UMAP}, which is widely used in computational biology. With a better initialization created by unsupervised machine learning algorithms (such as spectral clustering) and better loss functions, these algorithms improve global structure preservation and run-time efficiency. 
%Following the idea of t-SNE, computational biologists are also using graph layout algorithms such as ForceAtlas 2 \citep{ForceAtlas2} over the $k$-nearest-neighbor graph for visualization. These algorithms usually improve running speed and preservation of both local and global structure by using a different loss function.
Figure~\ref{fig:gaussian_dr} demonstrates differences in results from DR methods on a dataset of isotropic Gaussians. As the figure shows, the results of different DR techniques look quite different from each other.
Recent studies on DR algorithms shed light on how the loss function affects the rendering of local structure \citep{Unify}, and provide guidance on how to design good loss functions so that the local and global structure can both be preserved simultaneously \citep{pacmap}. 
Nevertheless, several challenges still exists for DR methods:

\begin{enumerate}[\thesection .1]
    \item \textbf{How to capture information from the high dimensional space more accurately?}
    
    Most of the recent DR methods 
    %\citep{vandermatten08, Tang16, UMAP, AtSNE, Belkina19OptSNE, pacmap} 
    capture information in the high dimensional space mainly from the $k$-nearest-neighbors and their relative distances, at the expense of information from points that are more distant, which would allow the preservation of more global structure.
    %graph, where each node represents a sample, and each edge represents the nearest neighbor relationship between two samples, weighted by the distance between them. However, the $k$-nearest-neighbors graph cannot capture non-local information in the high dimensional space. 
    \citet{usetsne, useumap} discuss possible pitfalls in data analysis created by t-SNE and UMAP due to loss of non-local information. Recent methods mitigate the loss of global information by using global-aware initialization \citep{UMAP, artoftsne} (that is, initializing the distances in the low-dimensional space using PCA) and/or selectively preserving distances between non-neighbor samples \citep{AtSNE, pacmap}. Nevertheless, these methods are still designed and optimized under the assumption that the nearest neighbors, defined by the given metric (usually Euclidean distance in the high dimensional space), can accurately depict the relationships between samples. This assumption may not hold true for some data, for instance, Euclidean distance may not be suitable for measuring distances between weights (or  activations) of a neural network  \citep[see][for a detailed example of such a failure]{tsnecuda}. We would like DR methods to better capture information from the high dimensional space to avoid such pitfalls. 
    
\begin{figure}[th]
    \centering
    \includegraphics[scale=0.45]{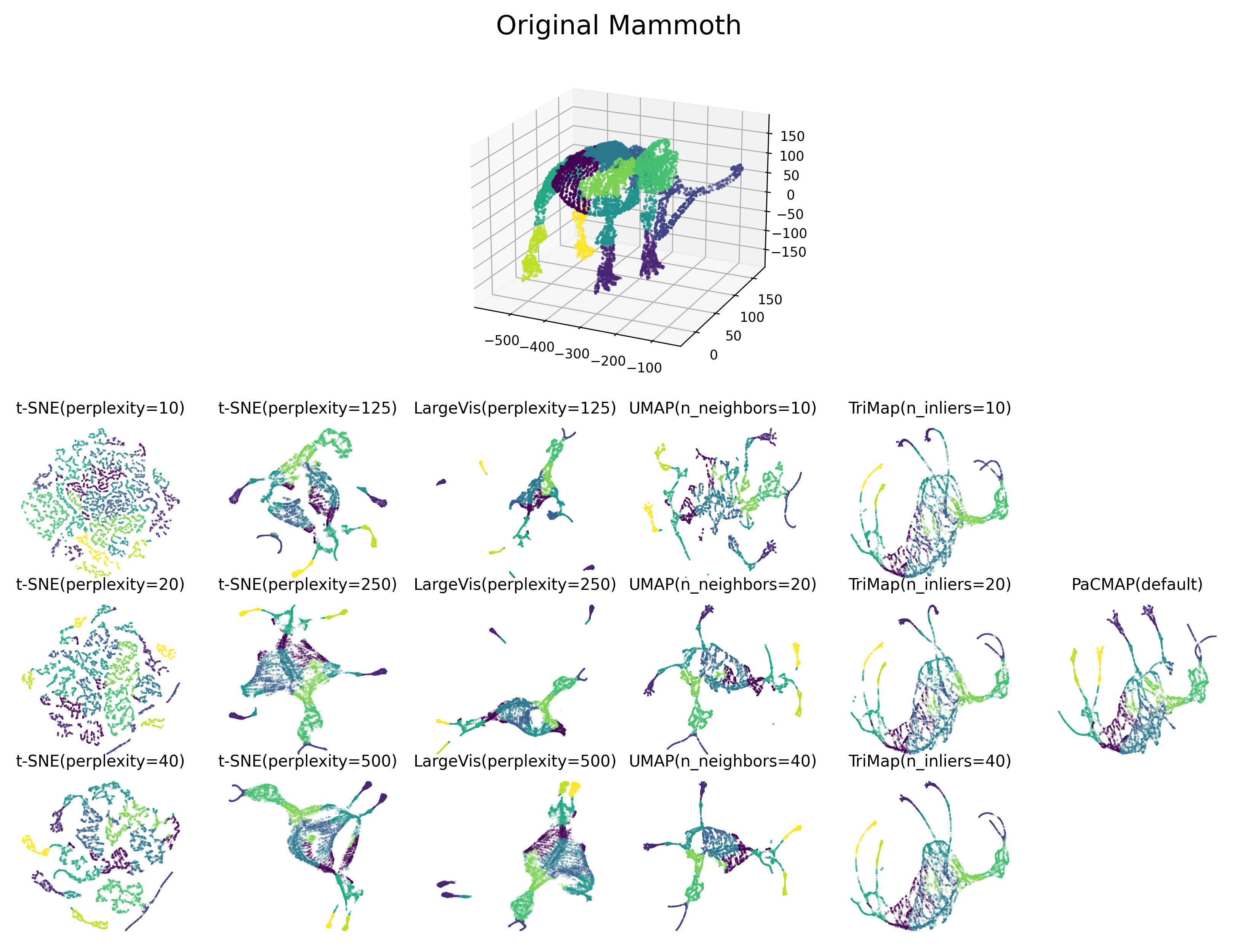}
    \caption{Projection from \citet{pacmap} of the Mammoth dataset into 2D using t-SNE  \citep{vandermatten08,Linderman19FItSNE,openTSNE}, LargeVis \citep{Tang16}, UMAP \citep{UMAP}, TriMap \citep{2019TRIMAP} and PaCMAP \citep{pacmap}. Incorrectly-chosen hyperparameters will lead to misleading results even in a simple dataset. This issue is particularly visible for t-SNE (first two columns) and UMAP (fourth column). The original dataset is 3 dimensional and is shown at the top.}
    \label{fig:mammoth_dr}
\end{figure}

    \item \textbf{How should we select hyperparameters for DR?}
    
    Modern DR methods, due to their multi-stage characteristics, involve a large number of hyperparameters, including the number of high-dimensional nearest neighbors to be preserved in the low-dimensional space and the learning rate used to optimize the low-dimensional embedding. There are often dozens of hyperparameters in any given DR method, and since DR methods are unsupervised and we do not already know the structure of the high-dimensional data, it is difficult to tune them. A poor choice of hyperparameters may lead to disappointing (or even misleading) DR results. Fig~\ref{fig:mammoth_dr} shows some DR results for the Mammoth dataset \citep{SmithsonianMammoth, useumap} using t-SNE, LargeVis, UMAP, TriMAP and PaCMAP with different sets of reasonable hyperparameters. When their perplexity parameter or the number of nearest neighbors is not chosen carefully, algorithms can fail to preserve the global structure of the mammoth (specifically, the overall placement of the mammoth's parts), and they create spurious clusters (losing connectivity between parts of the mammoth) and lose details (such as the toes on the feet of the mammoth). For more detailed discussions about the effect of different hyperparameters, see \cite{usetsne, useumap}. Multiple works  \citep[for example][]{Belkina19OptSNE, pacmap} aimed to alleviate this problem for the most influential hyperparameters, but the problem still exists, and the set of hyperparameters remains data-dependent. 
    %Users have to do multiple experiments to find the correct set of hyperparameters. 
    The tuning process, which sometimes involves many runs of a DR method, is time and power consuming, and requires user expertise in both the data domain and in DR algorithms. It could be extremely helpful to achieve better automatic hyperparameter selection for DR algorithms.

    \item \textbf{Can the DR transformation from high- to low-dimensions be made more interpretable or explainable?} The DR mapping itself -- that is, the transformation from high to low dimensions -- typically is complex. There are some cases in which insight into this mapping can be gained, for instance, if PCA is used as the DR method, we may be able to determine which of the original dimensions are dominant in the first few principle components. 
    %While the clusters and manifold structures revealed by DR algorithms in the low-dimensional space can be illuminating, the mapping itself to that space is often complicated. 
    It may be useful to design modern approaches to help users understand how the final two or three dimensions are defined in terms of the high-dimensional features. This may take the form of explanatory post-hoc visualizations or constrained DR methods.

\end{enumerate}

%Scientific discovery application  - don't want to examine a lot of spurious clusters.

%These two questions can potentially create many issues over the application of DR algorithms. 
\noindent\textbf{Example:} Computational biologists often apply DR methods to single-cell RNA sequence data to understand the cell differentiation process and discover previously-unknown subtypes of cells. Without a suitable way to tune parameters, they may be misled by a DR method into thinking that a spurious cluster from a DR method is actually a new subtype of cell, when it is simply a failure of the DR method to capture local or global structure (Challenge \ChDR.2 above). Since tuning hyperparameters in a high-dimensional space is difficult (without the ground truth afforded to supervised methods), the researchers have no way to see whether this cluster is present in the high-dimensional data or not (Challenge \ChDR.1 above).
%If they are not able to capture information in the original data, we may  can lead to inaccurate trajectories in the DR results, therefore harming the understanding of the differentiation process. Unsuitable hyperparameters, on the other hand, cause a lot of spurious clusters to be created, and 
Scientists could waste a lot of time examining each such spurious cluster. If we were able to solve the problems with DR tuning and structure preservation discussed above, it will make DR methods more reliable, leading to potentially increased understanding of many datasets. 

%%%%%%%%%%%%%%
\section{Machine learning models that incorporate physics and other generative or causal constraints}\label{sec:physics}

There is a growing trend towards developing machine learning models that incorporate physics (or other) constraints. These models are not purely data-driven, in the sense that their training may require little data or no data at all \citep[e.g.,][]{rao2020physics}. Instead, these models are trained to observe physical laws, often in the form of ordinary (ODEs) and partial differential equations (PDEs). These physics-guided models provide alternatives to traditional numerical methods (e.g., finite element methods) for solving PDEs, and are of immense interest to physicists, chemists, and materials scientists. The resulting models are interpretable, in the sense that they are constrained to follow the laws of physics that were provided to them. (It might be easier to think conversely: physicists might find that a standard supervised machine learning model that is trained on data from a known physical system -- but that does not follow the laws of physics -- would be uninterpretable.)

%Let us explain how we could possibly have a machine learning model trained with no data but only with a set of PDEs. Let us say (for sake of example) that $f'(x) = f(x)$ with initial condition $f(0) = 1$. We aim to approximate $f(x)$ with $g(x)$,  $\sum (g'(x_i) - g(x_i))^2 + (g(0) - 1)^2$.

\begin{figure}[th]
    \centering
    \includegraphics[width=.9\linewidth]{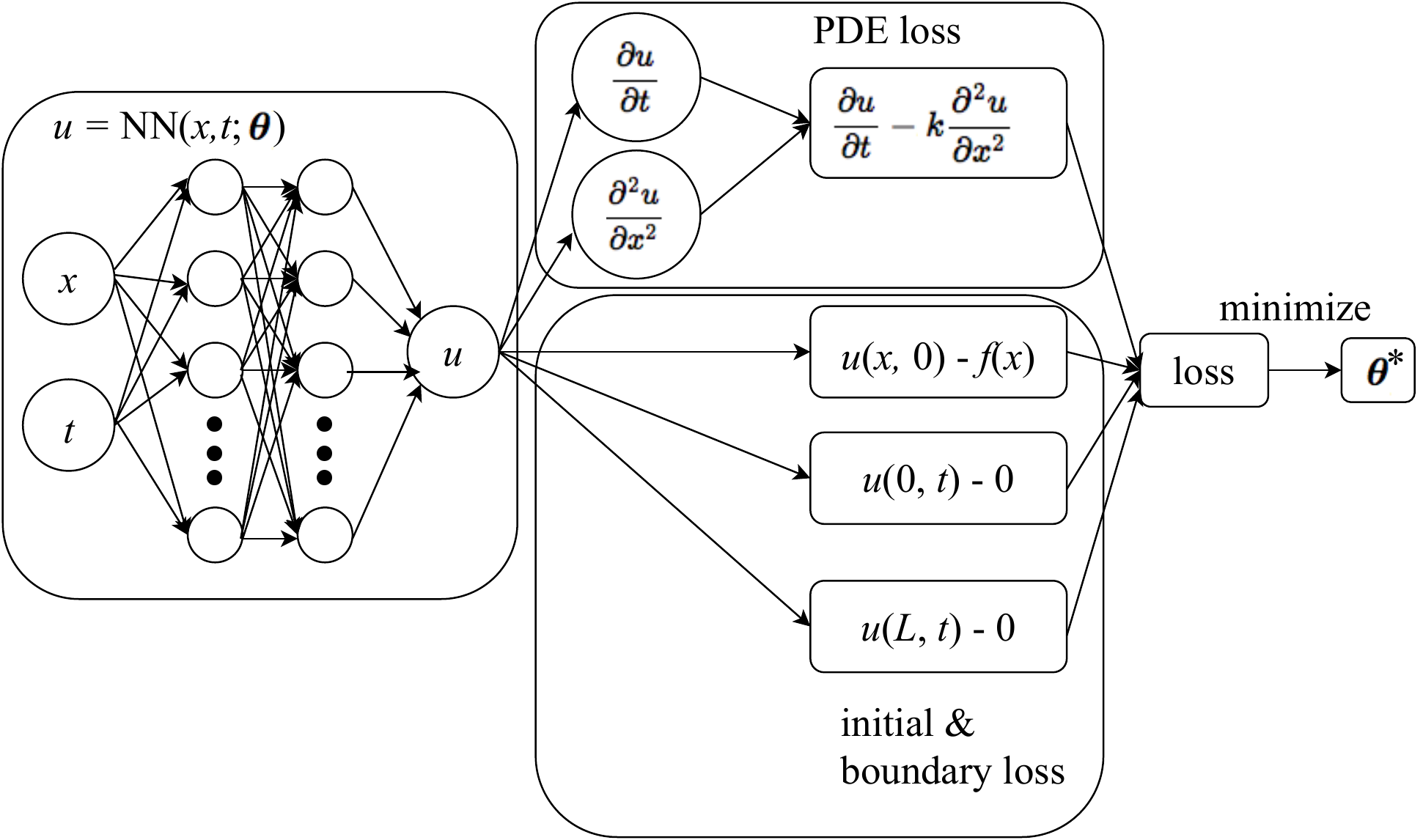}
    \caption{Physics-informed neural network (PINN) framework for solving the one-dimensional heat equation $\frac{\partial u}{\partial t}=k\frac{\partial^2 u}{\partial x^2}$ with initial condition $u(x, 0) = f(x)$ and Dirichlet boundary conditions $u(0, t)=0$ and $u(L,t)=0$. Here, $u$ is the model, which is the output of a neural network (left), obeys the heat equation (upper right) and initial and boundary conditions (lower right). The degree to which $u$ does not obey the heat equation or initial or boundary conditions is reflected in the loss function.}
    \label{fig:pinn}
\end{figure}

The idea of using machine learning models to approximate ODEs and PDEs solutions is not new. \citet{lee1990neural} developed highly parallel algorithms, based on neural networks, for solving finite difference equations (which are themselves approximations of original differential equations). \citet{psichogios1992hybrid} created a hybrid neural network-first principles modeling scheme, in which neural networks are used to estimate parameters of differential equations. \citet{lagaris1998artificial,lagaris2000neural} explored the idea of using neural networks to solve initial and boundary value problems.
%The idea of combining a neural network and a physical model was explored by \cite{psichogios1992hybrid}, who developed a hybrid neural network-first principles modeling scheme. The idea of using a neural network to approximate the solution of an initial/boundary value problem was explored by \cite{lagaris1998artificial}.
More recently, \cite{raissi2019physics} extended the earlier works and developed the general framework of a \textit{physics-informed neural network} (PINN). In general, a PINN is a neural network that approximates the solution of a set of PDEs with initial and boundary conditions. The training of a PINN minimizes the residuals from the PDEs as well as the residuals from the initial and boundary conditions. In general, physics-guided models (neural networks) can be trained without supervised training data. Let us explain how this works. Given a differential equation, say, $f'(t) = af(t) + bt + c$, where $a$, $b$ and $c$ are known constants, we could train a neural network $g$ to approximate $f$, by minimizing $(g'(t) - ag(t) - bt - c)^2$ at finitely many points $t$. Thus, no labeled data in the form $(t, f(t))$ (what we would need for conventional supervised machine learning) is needed. The derivative $g'(t)$ with respect to input $t$ (at each of those finitely many points $t$ used for training) is found by leveraging the existing network structure of $g$ using back-propagation. Figure \ref{fig:pinn} illustrates the training process of a PINN for approximating the solution of one-dimensional heat equation $\frac{\partial u}{\partial t}=k\frac{\partial^2 u}{\partial x^2}$ with initial condition $u(x, 0) = f(x)$ and Dirichlet boundary conditions $u(0, t)=0$ and $u(L,t)=0$. If observed data are available, a PINN can be optimized with an additional mean-squared-error term to encourage data fit. Many extensions to PINNs have since been developed, including fractional PINNs \citep{pang2019fpinns} and parareal PINNs \citep{meng2020ppinn}. PINNs have been extended to convolutional \citep{zhu2019physics} and graph neural network \citep{seo2019differentiable} backbones. They have been used in many scientific applications, including fluid mechanics modeling \citep{raissi2020hidden}, cardiac activation mapping \citep{sahli2020physics}, stochastic systems modeling \citep{yang2019conditional}, and discovery of differential equations \citep{raissi2018multistep,raissi2018deep_hidden_physics_models}.

In addition to neural networks, Gaussian processes are also popular models for approximating solutions of differential equations. For example, \cite{archambeau2007gaussian} developed a variational approximation scheme for estimating the posterior distribution of a system governed by a general stochastic differential equation, based on Gaussian processes. \cite{zhao2011pde} developed a PDE-constrained Gaussian process model, based on the global Galerkin discretization of the governing PDEs for the wire saw slicing process. More recently, \cite{pang2019neural} used the neural-network-induced Gaussian process (NNGP) regression for solving PDEs.

Despite recent success of PINNs or physics-guided machine learning in general, challenges still exist, including:

\begin{enumerate}[\thesection .1]
    \item \textbf{How to improve training of a PINN?} 
    \cite{wang2020understanding} discussed the problem of numerical stiffness in gradient backpropagation during training of a PINN. To mitigate the problem, they proposed an algorithm to dynamically balance various loss terms during training. How to further improve training of a PINN is still an open challenge that needs to be addressed.
    
    \item \textbf{How do we combine PINNs with optimal experimental/simulation design to quickly reduce uncertainty in approximating physical models?} Currently, data that might be used for training PINNs are often collected (from experiments or simulations) beforehand (i.e., prior to training of PINNs). This could lead to large uncertainty bands on areas of the input domain where data are scarce, and when the parameters of the PDE need to be learned from these data. Integrating experimental/simulation design with training of PINNs could lead to a more efficient data collection process and could reduce uncertainty of PINNs in approximating physical models. However, how to achieve this is still an open question.
    
    \item \textbf{Are there other ways of incorporating PDE (or other) constraints into models other than neural networks?} Neural networks can be easily differentiated using the chain rule. This makes them ideal for approximating solutions to PDEs. Other models, especially non-parametric regression models cannot be differentiated easily -- even though there have been some works that used Gaussian processes for approximating solutions to PDEs, these Gaussian-process solutions often rely on discrete (finite-difference) approximations for the original PDEs \citep[see][]{zhao2011pde}, or they could be non-differentiable at times where there are observations  \citep[see][]{archambeau2007gaussian}. How one can incorporate differential constraints (in terms of differential equations) into non-parametric regression models remains an open question.
    
    Going beyond PDEs to other types of complex constraints that encode prior knowledge, \cite{Kursuncu20} proposed the concept of \textit{knowledge infused} deep learning, where the learning of a deep neural network is guided by domain knowledge represented as a knowledge graph. Knowledge graphs encode relationships between entities in the world, including common sense knowledge. Thus, being able to incorporate information from a knowledge graph could be very helpful if we can figure out ways to do it effectively.
\end{enumerate}

\noindent\textbf{Example:} Suppose that a team of aerospace engineers wants to study the aerodynamics for a new design of aircraft wing. They want to use a physics-guided machine learning model to approximate a solution to the physical model (a set of PDEs), because numerically computing the PDE solution is computationally expensive. They want to try different machine learning models, but they need to first think about how they can integrate the PDE constraints into machine learning models that are not easily differentiable with respect to input (Challenge \ChPhysics.3). If they use a PINN, they need to ensure that training does not suffer from gradient pathologies (Challenge \ChPhysics.1). In addition, collecting data to support training of a PINN by conducting experiments or running simulations can be expensive. They want to know how to integrate experimental/simulation design with training of PINNs to achieve the best model (Challenge \ChPhysics.2). How to integrate constraints into various models, and how to improve training and to co-design experiments/simulations with model training are challenges that burden not only aerospace engineers but any scientist working with physics-guided machine learning models.

%%%%%%%%%%%%%%%%%%%%%% SECTION 8
\section{Characterization of the ``Rashomon'' set of good models}\label{sec:Rashomon} 
In many practical machine learning problems, there is a multiplicity of almost-equally-accurate models. This set of high performing models is called the \textit{Rashomon set}, based on an observation of the \textit{Rashomon effect}  by the statistician Leo Breiman. The Rashomon effect occurs when there are multiple descriptions of the same event \citep{breiman2001statistical} with possibly no ground truth. The Rashomon set includes these multiple descriptions of the same dataset  \citep[and again, none of them are assumed to be the truth, see][]{semenova2019study,fisher2018model,dong2020exploring,marx2020predictive}.
Rashomon sets occur in multiple domains, including credit score estimation, medical imaging, natural language processing, health records analysis, recidivism prediction, and so on \citep{d2020underspecification, marx2020predictive}. 
We have discussed in Challenges \ChCaseBased{} and \ChSupDis{} that even deep neural networks for computer vision exhibit Rashomon sets, because neural networks that perform case-based reasoning on disentanglement still yielded models that were equally accurate to their unconstrained values; thus, these interpretable deep neural models are within the Rashomon set.
Rashomon sets present an opportunity for data scientists: if there are many equally-good models, we can choose one that has desired properties that go beyond minimizing an objective function. In fact, the model that optimizes the training loss might not be the best to deploy in practice anyway due to the possibilities of poor generalization, trust issues, or encoded inductive biases that are undesirable \citep{d2020underspecification}. More careful approaches to problem formulation and model selection could be taken that include the possibility of model multiplicity in the first place. Simply put -- we need ways to explore the Rashomon set, particularly if we are interested in model interpretability.

%A multiplicity of good models occurs often in practical problems, where different machine learning algorithms achieve approximately the same level of performance \citep{breiman2001statistical, Rudin19, semenova2019study}. In this case, many almost-equally-accurate models exist, and we call this set of good models the \textit{Rashomon set}, based on an observation by the statistician Leo Breiman on its existence. The model that optimizes the training loss might not be the best to deploy in practice due to the possibilities of poor generalization, trust issues, or encoded inductive biases that are undesirable \citep{d2020underspecification}. These can occur in multiple domains, especially including credit score estimation, medical imaging, natural language processing, health records analysis, recidivism prediction, so on \citep{d2020underspecification, marx2020predictive}. Thus, more careful approaches to problem formulation and model selection have to be taken that include the possibility of the model multiplicity in the first place.
%Considering the set of good models (\textit{the Rashomon set}) helps with the modeling challenges described above. First, Knowing the Rashomon set or its properties can help to narrow down the search for good hypotheses from a model class. Instead of looking for one model from a model class, one can search for a model from this set that would be almost as accurate as the training loss minimizer. Second, 

\begin{figure}[th]
    \centering
    \includegraphics[width=\textwidth]{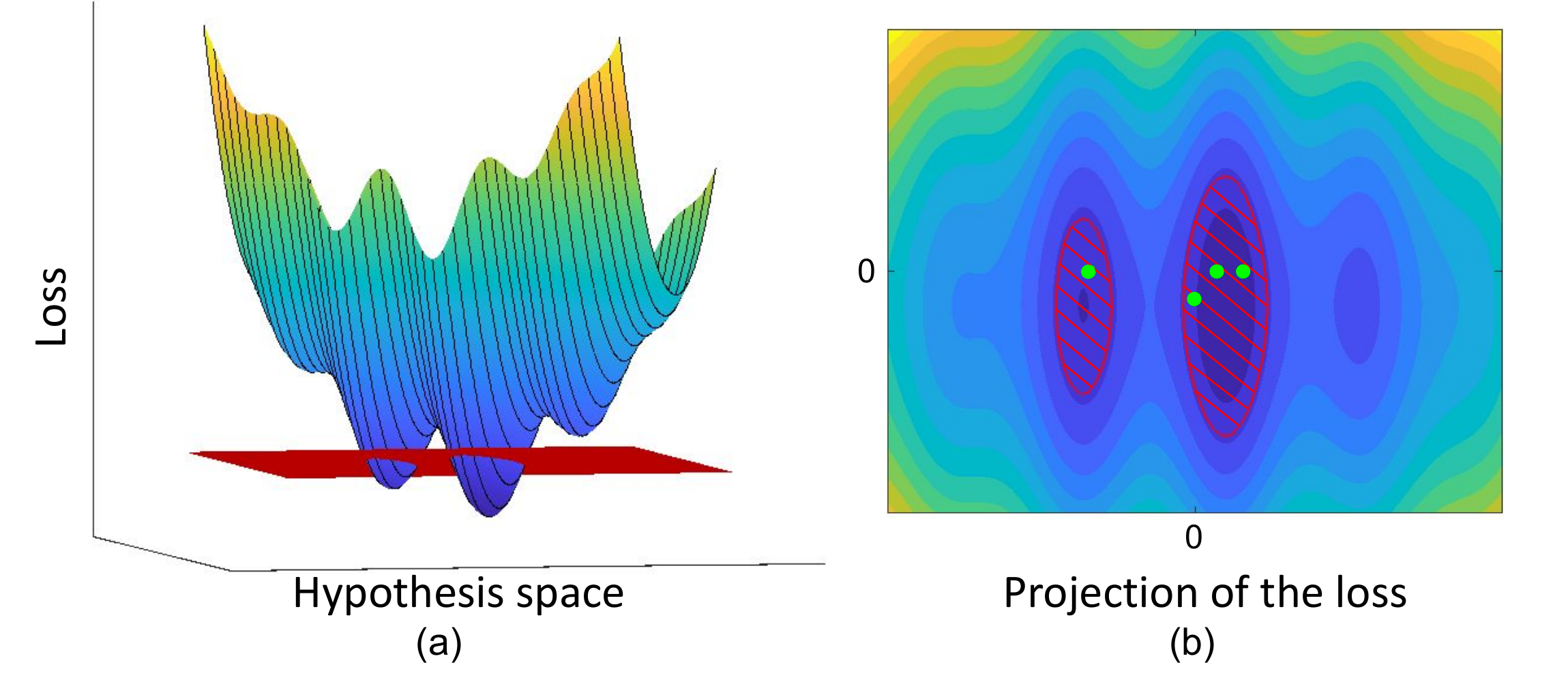}
	\caption{ (a) An illustration of a possible Rashomon set in two-dimensional hypothesis space. Models from two local minima that are below the red plane belong to the Rashomon set. (b) An illustration of a possible visualization of the Rashomon set. Models inside the shaded red regions belong to the Rashomon set. The plot is created as a contour plot of the loss over the hypothesis space. Green dots represent a few simpler models inside the Rashomon set. These models are simpler because they are sparse: they depend on one less dimension than other models in the Rashomon set.} 
	\label{fig:rset_example}
\end{figure}

Formally, the Rashomon set is the set of models whose training loss is below a specific threshold, as shown in Figure \ref{fig:rset_example} (a). Given a loss function and a model class $\mathcal{F}$, the Rashomon set can be written as
\[R(\mathcal{F}, f^*, \epsilon) = \{f \in \mathcal{F} \text{ such that } Loss(f) \leq Loss (f^*) + \epsilon\},\]
where $f^*$ can be an empirical risk minimizer, optimal model or any other reference model. We would typically choose $\mathcal{F}$ to be complex enough to contain models that fit the training data well without overfitting. The threshold $\epsilon$, which is called the Rashomon parameter, can be a hard hyper-parameter set by a machine learning practitioner or a percentage of the loss (i.e., $\epsilon$ becomes $\epsilon' Loss (f^*)$). We would typically choose $\epsilon$ or $\epsilon'$ to be small enough that suffering this additional loss would have little to no practical significance on predictive performance. For instance, we might choose it to be much smaller than the (generalization) error between training and test sets. We would conversely want to choose $\epsilon$ or $\epsilon'$ to be as large as permitted so that we have more flexibility to choose models within a bigger Rashomon set.

It has been shown by \citet{semenova2019study} that when the Rashomon set is large, under weak assumptions, it must contain a simple (perhaps more interpretable) model within it. The argument goes as follows: assume the Rashomon set is large, so that it contains a ball of functions from a complicated function class $\mathcal{F}_{\textrm{complicated}}$ (think of this as high-dimensional polynomials that are complex enough to fit the data well without overfitting). If a set of simpler functions $\mathcal{F}_{\textrm{simpler}}$ could serve as approximating set for $\mathcal{F}_{\textrm{complicated}}$ (think decision trees of a certain depth approximating the set of polynomials), it means that each complicated function could be well-approximated by a simpler function (and indeed, polynomials can be well-approximated by decision trees). By this logic, the ball of $\mathcal{F}_{\textrm{complicated}}$ that is within the Rashomon set must contain at least one function within $\mathcal{F}_{\textrm{simpler}}$, which is the simple function we were looking for. 

\citet{semenova2019study} also suggested a useful rule of thumb for determining whether a Rashomon set is large: run many different types of machine learning algorithms (e.g., boosted decision trees, support vector machines, neural networks, random forests, logistic regression) and if they generally perform similarly, it correlates with the existence of a large Rashomon set (and thus the possibility of a simpler function also achieving a similar level of accuracy). The knowledge that there might exist a simple-yet-accurate function before finding it could be very useful, particularly in the cases where finding an optimal sparse model is NP-hard, as in Challenge \ChLogical. Here, the user would run many different algorithms to determine whether it would be worthwhile to solve the hard problem of finding an interpretable sparse model.

%Assume that the model class contains models of different complexity. For a function from a more complicated class that is in the Rashomon set, if there exists a simpler, possibly interpretable, function from a simpler function class, then this interpretable function also is contained within the Rashomon set. This argument allows us to prove that the interpretable model exists before finding it \citep{semenova2019study}, that in a lot of cases can be NP-hard as discussed in Section \ref{sec:sparse}. Finally, 

The knowledge of the functions that lie within Rashomon sets sees value in multiple interesting use cases. 
Models with various important properties besides interpretability can exist in the Rashomon set, including fairness \citep{coston2021characterizing} and monotonicity. Thus, understanding the properties of the Rashomon set could be pivotal for analysis of a complex machine learning problem and its possible modeling choices.

The \textit{size of the Rashomon set} can be considered as a way of measuring the \textit{complexity of a learning problem}. Problems with large Rashomon sets are less complicated problems, since more good solutions exist to these problems. The Rashomon set is a property of the model class and dataset. 
The size of the Rashomon set differs from other known characterizations of the complexity in machine learning. Ways that complexity of function classes, algorithms, or learning problems are typically measured in statistical learning theory include the VC dimension, covering numbers, algorithmic stability, Rademacher complexity, or flat minima \citep[see, for instance,][]{srebro2010smoothness, kakade2009complexity, zhou2002covering, schapire1998boosting, koltchinskii2002empirical, bousquet2002stability, vapnik1971uniform}; the size of the Rashomon set differs from all of these quantities in fundamental ways, and it is important in its own right for showing the existence of simpler models.   
%of the learning problem as those are mostly based on worst-case analysis 
% or are data agnostic \citep{vapnik1971uniform}. If the Rashomon set is small it could be possible that it contains models from complex model classes that are not inherently interpretable \citep{semenova2019study}  and, therefore, requires a more complicated, possibly even black-box solution. 

A useful way to represent the hypothesis for a problem is by projecting it to variable importance space, where each axis represents the importance of a variable. That way, a single function is represented by a point in this space (i.e., a vector of coordinates), where each coordinate represents how important a variable is to that model. (Here, variable importance for variable $v$ is measured by \textit{model reliance} or \textit{conditional model reliance}, which represents the change in loss we would incur if we scrambled variable $v$.) The Rashomon set can be represented as a subset of this variable importance space. 
\citet{fisher2018model} used this representation to create a \textit{model independent} notion of variable importance: specifically, they suggest to consider the maximum and minimum of variable importance across all models in the Rashomon set. This is called the Model Class Reliance (MCR). For any user who would pick a ``good'' model (i.e., a model in the Rashomon set), then the importance of the variable is within the range given by the MCR. \citet{dong2020exploring} expand on this by suggesting to visualize the ``cloud'' of variable importance values for models within the Rashomon set. This cloud helps us understand the Rashomon set well enough to choose a model we might find interpretable.
 %Rashomon sets can help in characterizing variable importance by defining the model class reliance as a range of how important a variable can be for any model from the Rashomon set \citep{fisher2018model}. The model class reliance is a more general and robust measure of variable importance as it is less affected by specific modeling choices. 
% Analogously to variable importance, Rashomon sets are used in order to compute the predictive fairness properties \citep{coston2021characterizing} by minimizing or maximizing the disparity between values of the protected or sensitive attribute over the Rashomon set. There are multiple works \citep{TulabandhulaRu13, TulabandhulaRu14MLJ, tulabandhula2014robust} that leverage the Rashomon set for assisting in the decision-making process, including constructing the uncertainty set for optimization over worst-case scenarios (that is, to perform robust optimization).

\begin{figure}[th]
    \centering
    \includegraphics[width=0.5\textwidth]{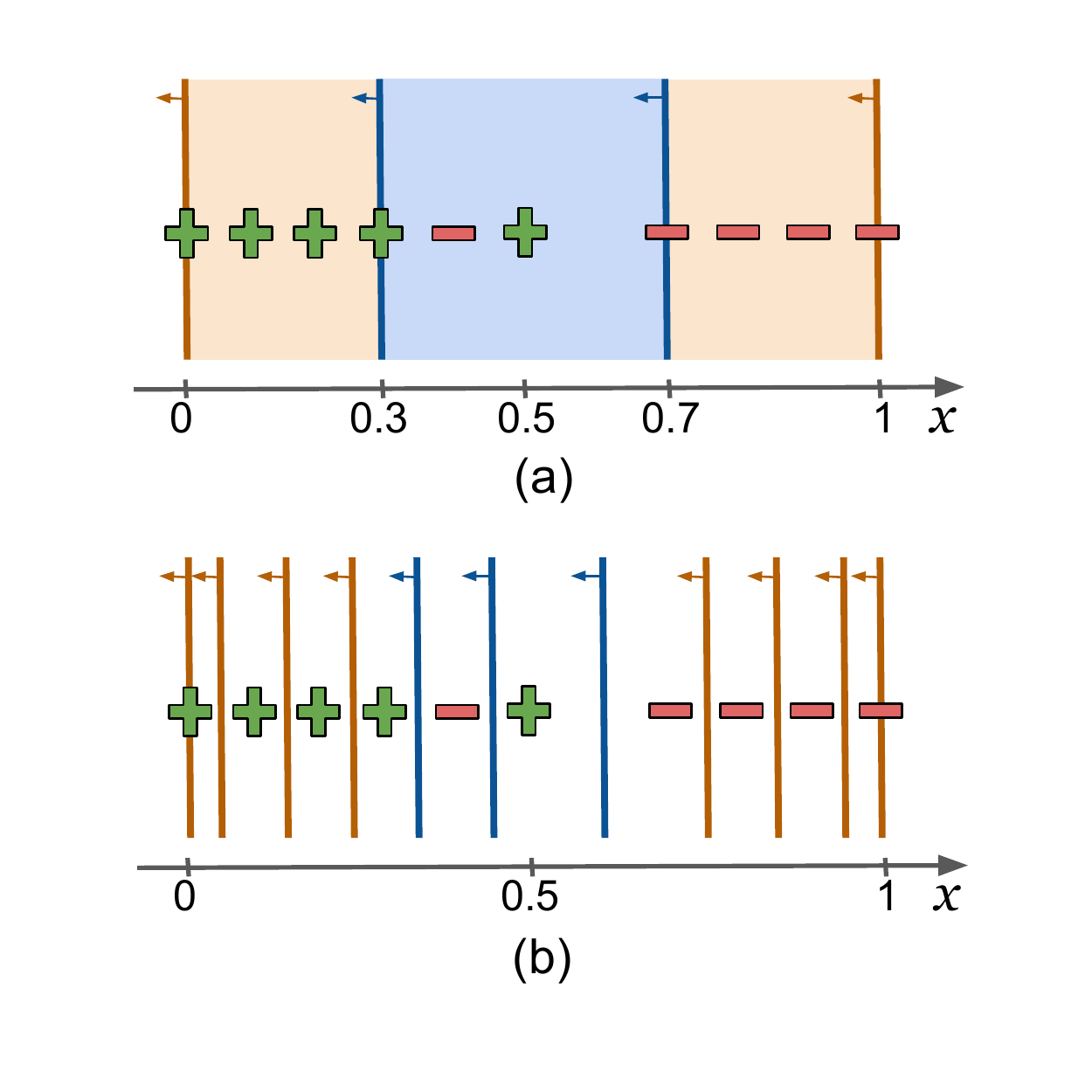}
	\caption{A one-dimensional example of computation of the Rashomon ratio. The hypothesis space consists of decision stumps $f(x) = \{\mathbbm{1}_{[x\leq a]}\}, a\in [0, 1]$. (The function is $1$ if $x\leq a$ and $0$ otherwise.) The loss function is the zero-one loss: $loss(f, x, y) = \mathbbm{1}_{[f(x) = y]}$, which is 1 if the model makes a mistake and zero otherwise. A model $f$ belongs to the Rashomon set if $\sum_i loss(f, x_i, y_i) \leq 0.2$, that is, the model made $2$ mistakes. The Rashomon ratio in (a) is equal to $0.4$ and is computed as the ratio of volumes of decision stumps in the Rashomon set (blue shaded region) to the volume of all possible decision stumps in the area where data reside (blue and orange shaded region). The pattern Rashomon ratio in (b) is equal to $3/11$ and is computed as the ratio of classifiers that belong to the Rashomon set (blue stumps, of which there are 3 unique stumps with respect to the data) to the number of all possible classifiers (orange and blue stumps, of which there are 11). This figure shows that there are multiple ways to compute the size of the Rashomon set and it is not clear which one to choose.
	} 
	\label{fig:rratio_example}
\end{figure}

Current research works have barely scratched the characterization and usage of model multiplicity and Rashomon sets in machine learning. Exploring the Rashomon set is critical if we hope to find interpretable models within the set of accurate models.
We ask the following questions about the Rashomon set:
%Among different questions about the Rashomon set, answering the following are the most important for the understanding of the set's properties: 
%\textbf{Measure: How to characterize the Rashomon set? How to measure its size?} \textbf{Visualize: What techniques can be used to visualize the Rashomon set?} \textbf{Choose: What model to choose from the Rashomon set?}

\begin{enumerate}[\thesection .1]
    \item \textbf{How can we characterize the Rashomon set?} As discussed above, the volume of the Rashomon set is a useful indirect indicator of the existence of interpretable models. There have been several works that suggest computing statistics of the Rashomon set in parameter space, such as the volume in parameter space, which is called the \textit{Rashomon volume}. The Rashomon volume or other statistics that characterize the Rashomon set can be computed more easily if the volume in parameter space has a single global minimum \citep{hochreiter1997flat, dinh2017sharp, keskar2016large, chaudhari2016entropy, keskar2016large, chaudhari2019entropy}; these variations include $\epsilon$-flatness \citep{hochreiter1997flat,dinh2017sharp} and $\epsilon$-sharpness \citep{keskar2016large, dinh2017sharp}. 
The \textit{Rashomon ratio} provides some perspective on the size of the Rashomon set, defined as the ratio of the Rashomon volume to the volume of the whole hypothesis space \citep{semenova2019study}, assuming that both are bounded. 
%\end{itemize}

%Comparing to the flat minima, this notion works well in discrete hypothesis spaces, where the volume function can be defined as a simple count of models. 

There are several problems that arise when computing the size of the Rashomon set in parameter space. First, the volume of a set of functions can be hard to compute or estimate, but there are ways to do it, including sampling (or analytical computation in some simple cases when there is one minimum).
%For very simple cases, these volumes can be computed analytically. In cases where the parameter space does not have too many dimensions, one could use an algorithm like WhiM \citep{lavine2019whim} to compute the volume. It is also possible to use sampling to estimate the volume. 
%For instance,  designed an algorithm called WhIM (Where it Matters) that uses axis-parallel boxes to approximate a region in a parameter space of functions. WHIM could potentially be used to approximate the Rashomon set. It is also possible to use sampling to estimate the volume of the Rashomon set.
Second, the choice of parameterization of the hypothesis space matters when working in parameter space. Overparametrization or underparametrization can cause volumes in parameter space to be artificially made larger or smaller without a change in the actual space of functions \citep{dinh2017sharp}. For instance, if we make a copy of a variable and include it in the datasets, it can change the Rashomon ratio. A third but separate problem with the Rashomon ratio is its denominator: the denominator is the volume of all models we might reasonably consider before seeing the data. If we change the set of models we might consider, we change the denominator, and the ratio changes. And, because different model spaces are different, there might not be a way to directly compare the Rashomon ratios computed on different hypothesis spaces to each other. 
%     Moreover, changes in the hypothesis space can significantly change the denominator of the ratio, making it hard to compare the ratios computed over hypothesis spaces of different complexity. 
% approximates function on important regions set up by constraints. WHIM could potentially be used to approximate the Rashomon set in restricted cases, as the rectangles that WHIM uses must be parallel to the axes, which is a potential limitation when the Rashomon set has smooth boundaries. 
    
This brings us to another way to measure the Rashomon set: The \textit{pattern Rashomon ratio}  \citep{semenova2019study,marx2020predictive}, which could potentially be used to handle some of these problems. The pattern Rashomon ratio considers unique predictions on the data (called ``patterns'') rather than the count of functions themselves. In other words, the pattern Rashomon ratio considers the count of predictions that could be made rather than a count of functions. Figure \ref{fig:rratio_example} illustrates the difference between the Rashomon ratio and the pattern Rashomon ratio. The pattern Rashomon ratio has a fixed denominator (all possible patterns one could produce on the data using functions from the class). A direct comparison of the pattern Rashomon ratio over function classes is meaningful. However, it is still difficult to compute. 
    %While an improvement over the Rashomon ratio, the computation of the pattern ratio is hard due to the huge search space of possible labelings achievable by the hypothesis space. 
    It is also more sensitive to small shifts and changes in data. 

    Despite that multiple measures, including $\epsilon$-flatness, $\epsilon$-sharpness, Rashomon ratio, and pattern ratio, have been proposed to measure the size of the Rashomon set, they do not provide a universal and simple-to-calculate way of characterization the size of the Rashomon set. Many challenges to remain open:
    \begin{enumerate}
        \item \textbf{What is a good space in which to measure the size of the Rashomon set? What is a good measure of the size of the Rashomon set?} As discussed above, the parameter space suffers from issues with parameterization, but the pattern Rashomon ratio has its own problems.
        %the function space depends on the volume definition and computation, the output space depends too much on the layout of data. 
       % Perhaps there are some ways to fix existing problems (for example, force the parameter space to allow only a hypothesis space with a parameterization that is unique for each model). Or, we could design a measure that computes the size of the Rashomon space in a new space. What should the properties of this new space be? 
       Would variable importance space be suitable, where each axis represents the importance of a raw variable, as in the work of \citet{dong2020exploring}? Or is there an easier space and metric to work with?
        %Can we design an effective and easy-to-compute metric for the  size of the Rashomon set or the Rashomon ratio in this space?
        %\item \textbf{What is a good measure of the size of the Rashomon set?} Given the new space and a new metric, how can we use it to compute the size of the Rashomon set? Should it still be a ratio over the hypothesis space or a new characteristic that is hypothesis-space-independent and easier to compute? 
        %Both the Rashomon ratio and pattern Rashomon ratio depend on the size of the hypothesis space. While this gives a nice grounding and understanding of how large the Rashomon set is in the percentage, the computation of the size of the hypothesis space is much more complicated than the computation of the Rashomon set.
        \item \textbf{Are there approximation algorithms or other techniques that will allow us to efficiently compute or approximate the size of the Rashomon set?} At worst, computation over the Rashomon set requires a brute force calculation over a large discrete hypothesis space. In most cases, the computation should be much easier. Perhaps we could use dynamic programming or branch and bound techniques to reduce the search space so that it encompasses the Rashomon set but not too much more than that? In some cases, we could actually compute the size of the Rashomon set analytically. For instance, for linear regression, a closed-form solution in parameter space for the volume of the Rashomon set has been derived based on the singular values of the data matrix \citep{semenova2019study}.
        %Most of the approaches to compute the Rashomon ratio requires a lot of computation and often brute force. For example, to compute the Rashomon ratio for a discrete space with volume function as count, one needs to check every model for its containment in the Rashomon set. However, models that achieve some level of performance can have some similar properties. Perhaps some greedy approaches can be designed that cut at a time chunks of the hypothesis space with models that satisfy these properties. Perhaps an approximation algorithm can be designed that takes advantage of a participation oracle (whether a model is in the Rashomon set). For example, if the loss is convex, given the participation oracle a randomized algorithm could be used to bound the volume of the Rashomon set \citep{semenova2019study}. 
       % \item \textbf{Can domain-specific constraints help in measuring the size of the Rashomon set?} For example, for linear regression, a closed-form solution in the parameter space for the volume of the Rashomon set has been derived based on the singular values of the data matrix \citep{semenova2019study}. Perhaps considering other constraints such as loss smoothness
        %, a specific property of the hypothesis space, or constraints that follow from the problem domain 
       % will help design better characterizations of the size of the Rashomon set. 
    \end{enumerate} 
    
    \item \textbf{What techniques can be used to visualize the Rashomon set?} Visualization of the Rashomon set can potentially help us to understand its properties, issues with the data, biases, or underspecification of the problem. To give an example of how visualization can be helpful in troubleshooting models generally, \citet{li2017visualizing} 
    %introduced a 
    %filter normalization method that helped to 
    visualized the loss landscape of neural networks and, as a result, found answers to questions about the selection of the batch size, optimizer, and network architecture. \citet{kissel2021forward} proposed a tree-based graphical visualization to display outputs of the \textit{model path selection procedure} that finds models from the Rashomon set based on a forward selection of variables. The visualization helps us to understand the stability of the model selection process, as well as the richness of the Rashomon set, since wider graphical trees imply that there are more models available for the selection procedure. To characterize the Rashomon set in variable importance space, \textit{variable importance diagrams} have been proposed \citep{dong2020exploring}, which are 2-dimensional projections of the \textit{variable importance cloud}. The \textit{variable importance cloud} is created by mapping every variable to its importance for every good predictive model (every model in the Rashomon set). 
    Figure \ref{fig:rset_example}(b) uses a similar technique of projection in two-dimensional space and depicts the visualization of the example of the Rashomon set of Figure \ref{fig:rset_example}(a). This simple visualization allows us to see the Rashomon set's layout, estimate its size or locate sparser models within it. 
    Can more sophisticated approaches be developed that would allow good visualization? The success of these techniques most likely will have to depend on whether we can design a good metric for the model class. After the metric is designed, perhaps we might be able to utilize techniques from Challenge \ChDR{} for the visualization.

    \item \textbf{What model to choose from the Rashomon set?}  %When the Rashomon set is small, the machine learning problem becomes less underspecified. In this case, the goal is to find a model from narrow local minima, and the modeling is focused on developing an algorithm for the model class. However, 
    When the Rashomon set is large, it can contain multiple accurate models with different properties. Choosing between them might be difficult, particularly if we do not know how to explore the Rashomon set. Interactive methods might rely on dimension reduction techniques (that allow users to change the location of data on the plot or to change the axis of visualization), weighted linear models (that allow the user to compute weights on specific data points), continuous feedback from the user (that helps to continuously improve models prediction in changing environments, for example, in recommender systems) in order to interpret or choose a specific model with desired property. \citet{das2019beames} design a system called BEAMES that allows users to interactively select important features, change weights on data points, visualize and select a specific model or even an ensemble of models. BEAMES searches the hypothesis space for models that are close to the practitioners' constraints and design choices. The main limitation of BEAMES is that it works with linear regression classifiers only. Can a similar framework that searches the Rashomon set, instead of the whole hypothesis space, be designed? What would the interactive specification of constraints look like in practice to help the user choose the right model? 
   % If there is a desirable constraint such as interpretability or fairness, then the modeler can focus on optimizing for this constrain. However, in cases when there is no additional specification and the modeler is not aware of the large Rashomon set, the model obtained as the best train performer may generalize poorly or be not trustworthy. Rashomon set even might contain models of the same complexity with the same level of performance, but with contradicting predictions \citep{marx2020predictive}. What can be a good model to choose in such cases? 
%    Can a framework that will allow interactive specification of constraints 
%    % (similar to Challenge 10) 
%    be helpful in choosing the right model? 
    Can collaboration with domain experts in other ways be useful to explore the Rashomon set?
\end{enumerate}

\noindent \textbf{Example}: Suppose that a financial institution would like to make data-driven loan decisions. The institution must  have a model that is as accurate as possible, and must provide reasons for loan denials. In practice, loan decision prediction problems have large Rashomon sets, and many machine learning methods perform similarly despite their different levels of complexity. To check whether the Rashomon set is indeed large, the financial institution wishes to measure the size of the Rashomon set (Challenge \ChRashomon.1) or visualize the layout of it (Challenge \ChRashomon.2) to understand how many accurate models exist, and how many of them are interpretable. %, do they need to trade-off any accuracy for the ability to get an interpretable model, so on. 
If the Rashomon set contains multiple interpretable models, the financial institution might wish to use an interactive framework (Challenge \ChRashomon.3) to navigate the Rashomon set and design constraints that would help to locate the best model for their purposes. For example, the institution could additionally optimize for fairness or sparsity.

\section{Interpretable reinforcement learning}\label{sec:rein}

%https://arxiv.org/abs/1901.00188  <-- useful?

Reinforcement learning (RL) methods determine what actions to take at different states in order to maximize a cumulative numerical reward \citep[e.g., see][]{sutton2018reinforcement}. In RL, a learning agent interacts with the environment by a trial and error process, receiving signals (rewards) for the actions it takes. % until the goal is achieved.
%A reinforcement learning (RL) system is provided with a set of rules and rewards by which the RL system should act \citep{sutton1998introduction}. The system interacts and explores the environment described by these rules until it learns good behaviors according to the rewards. 
%The goal is to optimize for long-term outcomes by learning a sequence of decisions that the system needs to make to receive the highest reward. 
Recently, deep reinforcement learning algorithms have achieved state-of-the-art performance in  mastering the game of Go and various Atari games \citep{silver2016mastering, mnih2013playing} and have been actively used in robotic control \citep{kober2013reinforcement, tai2016survey}, simulated autonomous navigation and self-driving cars \citep{kiran2021deep, sallab2017deep}, dialogue summarization, and question answering \citep{li2016deep}. RL has also been applied to high-stakes decisions, such as healthcare and personalized medicine, including approaches for dynamic treatment regimes \citep{liu2017deep, yu2019reinforcement}, treatment of sepsis \citep{komorowski2018artificial}, treatment of chronic disorders including epilepsy \citep{guez2008adaptive}, for HIV therapy \citep{parbhoo2017combining}, and management of Type 1 diabetes \citep{javad2019reinforcement}. 

\begin{figure}[th]
    \centering
    \includegraphics[width=0.6\textwidth]{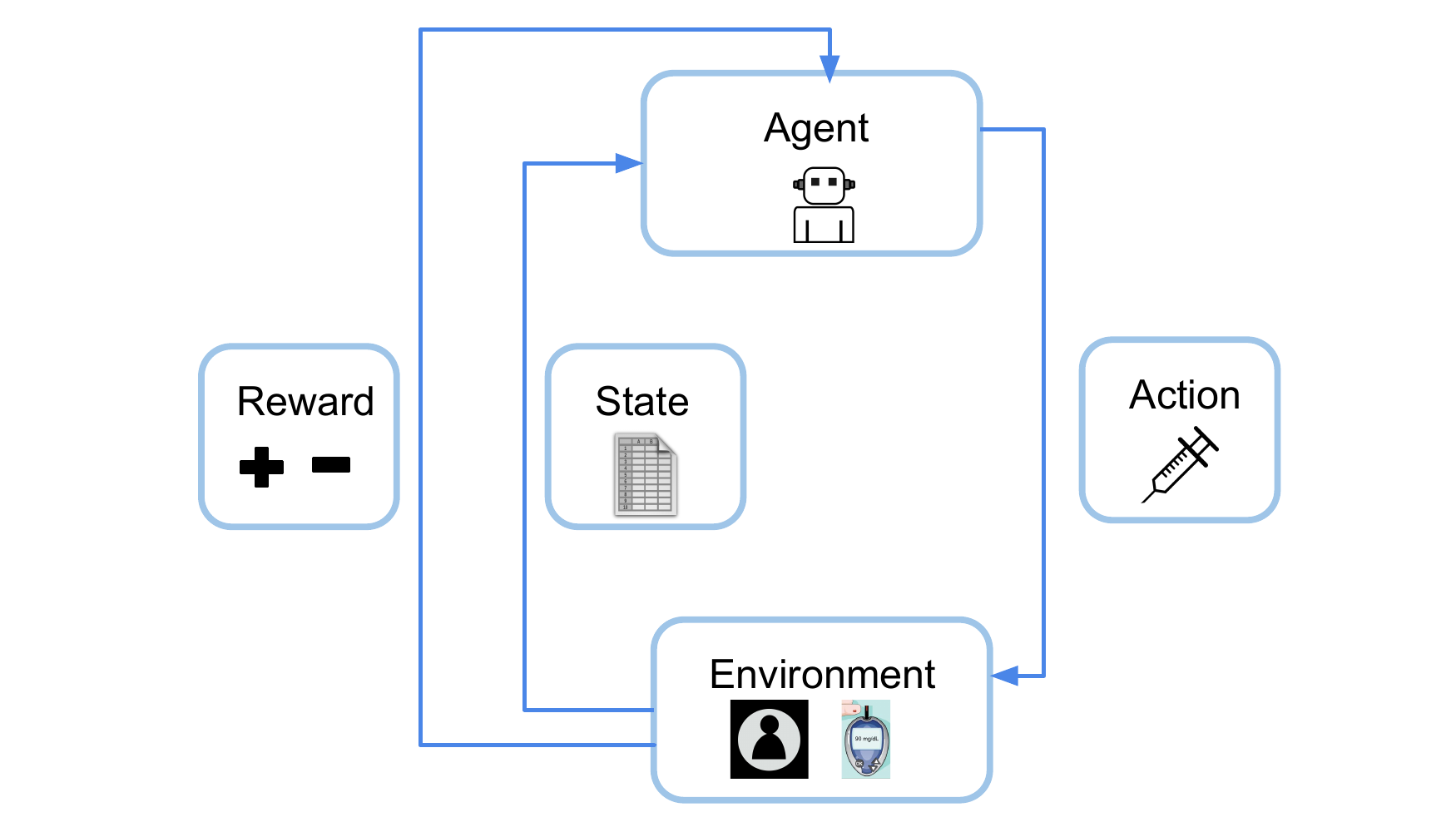}
	\caption{An illustration of an RL system based on the example of Type 1 diabetes management. This is theoretically a closed-loop system, where the agent observes the state from the environment, makes an action, and receives a reward based on the action. For diabetes control, the state consists of measurements of glucose and other patient data, the actions are injections of insulin, and the rewards are based on sugar levels.} 
	\label{fig:rlexample}
\end{figure}

In reinforcement learning, at each timestep, an agent observes the state (situation) that it is currently in, chooses an action according to a policy (a learned mapping from states to actions), and receives an intermediate reward that depends on the chosen action and the current state, which leads the agent to the next state, where the process repeats. For example, let us consider an RL system that is used for the management of glucose levels of Type 1 diabetes patients by regulating the amount of synthetic insulin in the blood. 
%Type 1 diabetes is caused by high levels of glucose in the blood because of little or no production of insulin by the body. This condition has no cure and treatment is focused on maintaining the blood sugar level by injecting the right amount of synthetic insulin, which is challenging to do at all times due to constant changes in the patient's diet, or daily habits. RL system can be used to control the amount of insulin that has to be injected. 
%The environment, in this case, will be represented by the patient's information (vitals, diet, exercise) and insulin interaction;
The state is represented by the patient's information, including diet, exercise level, weight, so on; actions include the amount of insulin to inject; rewards are positive if the blood sugar level is within a healthy range and negative otherwise. Figure \ref{fig:rlexample} shows the schematic diagram of a reinforcement learning system based on the diabetes example. 

Reinforcement learning is more challenging than a standard supervised setting: the agent does not receive all the data at once (in fact, the agent is responsible for collecting data by exploring the environment); sometimes besides immediate rewards, there are delayed rewards when feedback is provided only after some sequence of actions, or even after the goal is achieved (for example, the positive reward on a self-driving car agent is provided after it reaches a destination). Because reinforcement learning involves long-term consequences to actions, because actions and states depend on each other, and because agents need to collect data through exploration, interpretability generally becomes more difficult than in a standard supervised learning. % it can be more difficult to  It is easier to explain a single point classification results in the supervised learning comparing to the reinforcement learning, that is spanned over some period of time. There are more long term consequences that one needs to account for in the RL setting. 
%Additionally, there might be confounders that are difficult to pinpoint, for example, if the agent learns unexpected shortcuts while interacting with the environment that lead to a high reward.
Search spaces for RL can be massive since we need to understand which action to take in each state based on an estimate of future rewards, uncertainty of these estimates, and exploration strategy. In deep reinforcement learning \citep{mnih2013playing, silver2016mastering}, policies are defined by deep neural networks, which helps to solve complex applied problems, but typically the policies are essentially impossible to understand or trust.

Interpretability could be particularly valuable in RL. Interpretability might help to reduce the RL search space; if we understand the choices and intent of the RL system, we can remove actions that might lead to harm. 
%A key example of this is in sepsis management, where \citet{gottesman2019guidelines} show that a standard search of the state space would miss important parts of it that could be associated with poor outcomes. 
Interpretability could make it easier to troubleshoot (and use) RL for long-term medical treatment or to train an agent for human-robot collaboration.

%The process is restarted again. 
%The RL system can be formulated using a Markov Decision Process (MDP). An MDP is a 5-tuple $(S, A, P, R, \gamma)$, where $S=\{s\}$ is a state space, $A= \{a\}$ is an action space, $P$ are transition probabilities, where $p(s'|s,a)$ denotes a probability of transitioning from state $s$ to state $s'$ while taking an action $a$, $R$ is a reward function, where $R(s,a)$ is the intermediate reward for taking an action $a$ in state $s$, and $\gamma \in [0,1)$ is a discount factor for the future rewards. Given the discount factor $\gamma$, the total accumulated reward $R$  is typically set to be the sum of discounted future rewards $R = \sum_{y=0}^{\infty} \gamma^t r_{t}$. The policy $\pi: S \rightarrow A$, according to which the agent is acting, is a mapping from states to actions, where $\pi(s)$ denotes what action to take in state $s$. In deep reinforcement learning, policies $\pi$ are defined by deep neural networks, which helps tremendously to solve complex applied problems, but also makes it close to impossible to understand, trust or interpret the policy.

%\citep{kersting2008non} - policy gradients, where policy is represented by a sum of regression trees in relational domain
%\citep{roth2019conservative} - decision tree policy, propositional domain, have to settle for lower reward to get more compact tree - need to set a threshold that compromises between size of the tree and perfomance

\begin{figure}[th]
    \centering
    \includegraphics[width=0.9\textwidth]{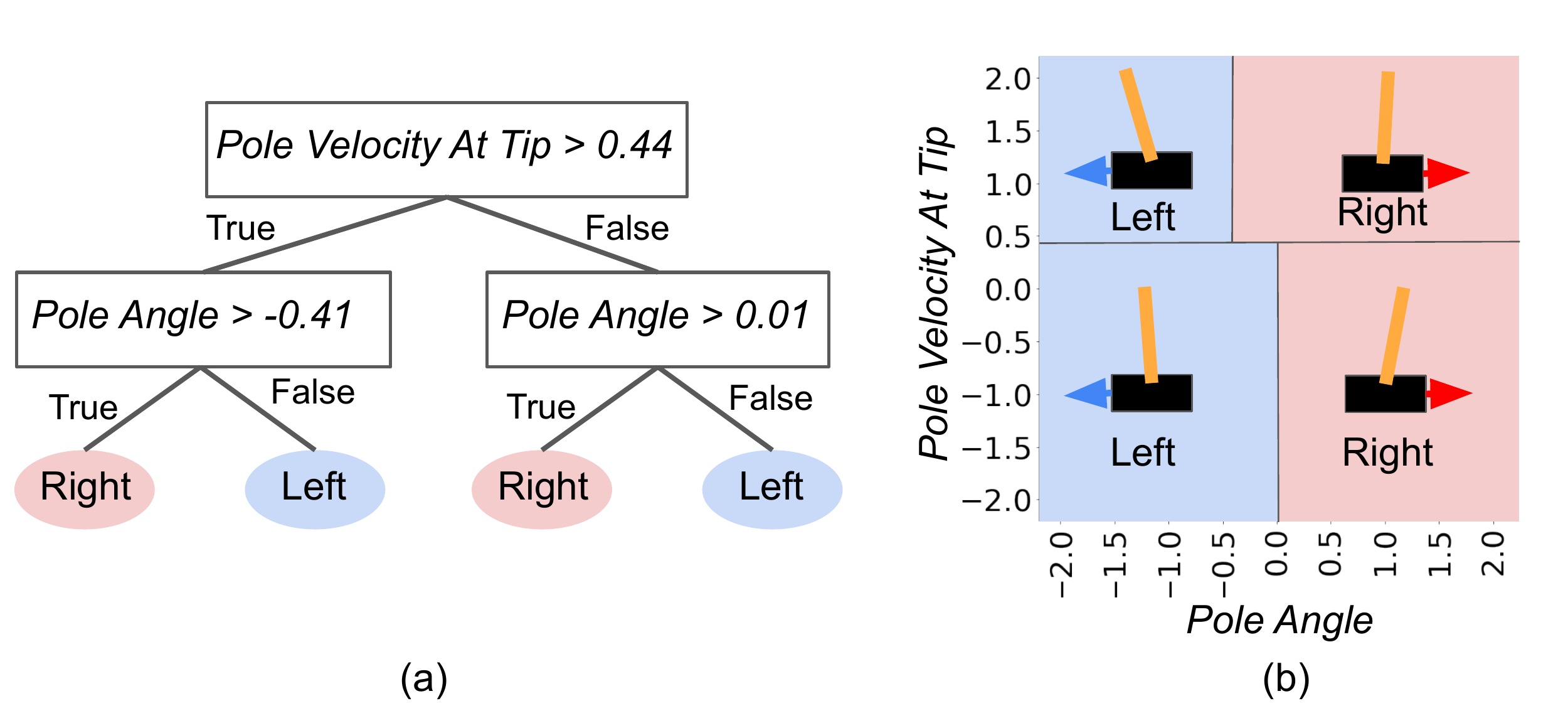}
	\caption{(a) An example of a learned decision tree policy for the Cart-Pole balancing problem adapted from \citet{silva2020optimization}. In the Cart-Pole environment, a pole is attached to a cart by a joint. The cart moves along the horizontal axis. The goal is to keep the pole balanced upright, preventing it from falling. The state consists of four features: \textit{Cart Position, Cart Velocity, Pole Angle and Pole Velocity At Tip}. Here, angle 0$^{\circ}$ means straight upwards. There are two possible actions at each state, including \textit{push cart to the left} or \textit{push cart to the right}. (b) is a visualization of the policy from \citet{silva2020optimization} based on two features: \textit{Pole Angle} and \textit{Pole Velocity At Tip}. Interestingly, it is easy to see from this figure that the policy is not left-right symmetric. Note that there are many reasonable policies for this problem, including many that are asymmetric.}
	\label{fig:policy_tree}
\end{figure}

One natural way to include interpretability in a reinforcement learning system is by representing the policy or value function by a decision tree or rule list \citep{roth2019conservative, silva2020optimization}. In a policy tree representation, the nodes might contain a small set of logical conditions, and leaves might contain an estimate of the optimal probability distribution over actions (see Figure \ref{fig:policy_tree}). A policy tree can be grown, for example, in a greedy incremental way by updating the tree only when the estimated discounted future reward would increase sufficiently \citep{roth2019conservative}; in the future, one could adapt methods like those from Challenge \ChLogical{} to build such trees. %Other literature considers representing the policy by a differential decision tree \citep{suarez1999globally}, which is learned with gradient descent \citep{silva2020optimization}. 
%A policy tree can be learned in a supervised manner, using a dataset of state-action pairs, to approximate the expected rewards for an action taken in a given state (the Q-function) at each time step \citep{ernst2005tree,roth2019conservative,silva2020optimization}. 
%These trees could be learned in an additive way by creating new branches instead of previous leaf nodes to represent a policy \citep{roth2019conservative}, by gradient update for differential decision tree (it allows gradient update by replacing boolean decisions in tree nodes with sigmoid activation) \citep{silva2020optimization}.

%Other way  as additive tress to  or trained  used as a supervised state aggregation method to Other tree-type methods (such as CART, UTree, tree bagging) have been also applied in reinforcement learning as a supervised state aggregation \citep{ernst2005tree, finney2012thing}. 
% Recently, \citep{silva2020optimization} proposed to use a differential decision tree (this tree allows gradient update by replacing the boolean decisions in tree nodes with sigmoid activation) as a function approximator in RL. As a result, after discretization, the policy is represented by a decision tree or a rule list, where nodes contain simple branching rules and leaves contain an estimate of the optimal probability distribution over actions for the policy gradient method.

 Other methods make specific assumptions about the domain or problem that allows for interpretable policies. One of the most popular simplifications is to consider a symbolic or relational state representation. In this case, the state is a combination of logical or relational statements. For example, in the classic Blocks World domain, where the objective is to learn how to stack blocks on top of each other to achieve a target configuration, these statements comprising the state might look like \textit{on(block1, block2)}; \textit{on(block1, table)}; \textit{free(block2)}. %Based on a relational representation, as a field that combines inductive logic programming and reinforcement learning, relational reinforcement learning \citep{dvzeroski1998relational, dvzeroski2001relational} was introduced over twenty years ago. 
 Typically, in relational reinforcement learning, policies are represented as relational regression trees \citep{dvzeroski1998relational,dvzeroski2001relational, kersting2008non, finney2012thing, das2020fitted} that are more interpretable than neural networks.

A second set of assumptions is based on a natural decomposition of the problem for multi-task or skills learning. For example, \citet{shu2017hierarchical} use hierarchical policies that are learned over multiple tasks, where each task is decomposed into multiple sub-tasks (skills). The description of the skills is created by a human, so that the agent learns these understandable skills (for instance, task \textit{stack blue block} can be decomposed into the skills \textit{find blue block}, \textit{get blue block}, and \textit{put blue block}).

%\begin{figure}[th]
%    \centering
%    \includegraphics[width=\textwidth]{figures/PIRL.png}
%	\caption{ An example of an interpretable policy from \citet{verma2018programmatically} for acceleration task in the  TORCS car racing environment. $h_{\textit{RPM}}$ and $h_{\textit{TP}}$ are information from the sensors used in the environment. \textcolor{dark-green}{peek}$(x, t)$ provides an observation from $x$ at time $t$ and \textcolor{dark-green}{fold}$(f,x)$ is a function $f$ combination over sequence $x$.}
%	\label{fig:PIRL}
%\end{figure}

% add a paragraph on DTR

As far as we know, there are currently no general interpretable well-performing methods for deep reinforcement learning that allow transparency in the agent's actions or intent. Progress has been made instead on explainable deep reinforcement learning (posthoc approximations), including tree-based explanations \citep{coppens2019distilling, liu2018toward, dhebar2020interpretable, bastani2018verifiable, dahlin2020designing}, reward decomposition \citep{juozapaitis2019explainable},  and attention-based methods \citep{zambaldi2018relational, mott2019towards}. %The explainability of real-world policies and actions of the controllers are one of the key challenges of real-world reinforcement learning \citep{dulac2019challenges}. 
An interesting approach is that of \citet{verma2018programmatically}, who define rule-based policies through a domain-specific, human-readable programming language that 
%is verifiably correct and 
generalizes to unseen environments, but shows slightly worse performance than neural network policies it was designed to explain.
%trained on. To find a rule-based policy, a deep reinforcement learning system is used to compute an oracle that guides the local search algorithm to find the rule-based policy that will minimize the distance to the oracle. The resulting rule-based policy is an approximation of the neural network rather than an interpretable policy learned directly from data. 
The approach works for deterministic policies and symbolic domains only and will not work for the domains where the state is represented as a raw data image, unless an additional logical relation extractor is provided.

Based on experience from supervised learning (discussed above), we know that post-hoc explanations suffer from multiple problems, in that explanations are often incorrect or incomplete. \citet{Atrey2020Exploratory} have argued that for deep reinforcement learning, saliency maps should be used for exploratory, and not explanatory, purposes.
%For example, saliency maps suffer from unfalsiability and cognitive bias and thus should be used for exploratory and not explanatory purposes \citep{Atrey2020Exploratory}. 
Therefore, developing interpretable reinforcement learning policies and other interpretable RL models is important. The most crucial challenges for interpretable reinforcement learning area include:
\begin{enumerate}[\thesection .1]
    \item \textbf{What constraints would lead to an accurate interpretable policy?} 
    In relational methods, policies or value functions could be represented by decision or regression trees, however, since these methods work for symbolic domains only, they have limited practical use without a method of extracting interpretable relations from the domain. Can interpretable deep neural network methods similar to those in Challenge \ChSupDis{} provide an interpretable policy without the performance compromise?
    Are there other ways of adding constraints or structure to the policy that will ensure human-understandable actions and intent? Will adding other sparsity or interpretability constraints, as in Challenges \ChLogical{} and \ChScoring{}, to the policy or value function help improve interpretability without sacrificing accuracy?  %how do we design models that not only explain the policy and provide estimated rewards for each possible action, but also provide some insight into the calculation of the value function?
    Can we design interpretable models whose explanations can capture longer term consequences of actions?
    
    %In PIRL \citep{verma2018programmatically}, policies were represented by programming languages.  
   % Understanding a policy decision in RL may be significantly more complicated than in classification because one needs to consider the full horizon of possible future actions and states. 
    %It is possible in these complex settings that there is a more severe tradeoff in performance for increased regularization \citep{roth2019conservative}. 

   % While tree-based representations provide interpretability for RL systems, these methods may need to be constantly retrained to adapt to changing data distributions \citep{ernst2005tree}, or they may have a trade-off between performance and regularization on tree size \citep{roth2019conservative}, or they may perform significantly worse than deep learning alternatives \citep{finney2012thing}.  

    %Another related question to the interpretability constraints is how to evaluate the level of interpretability of the RL system?  \citet{dulac2019challenges} advocates for the human to determine how well the policy's intent is understandable through A/B testing of agent's policies. 
    
    \item \textbf{Under what assumptions does there exist an interpretable policy that is as accurate as the more complicated black-box model?} Reinforcement learning systems are tricky to train, understand and debug. It is a challenge in itself to decide what constraints will lead to an interpretable policy without loss of performance. For example, for PIRL \citep{verma2018programmatically}, there was a drop in performance of the resulting rule-based policy. Of course, one of the possible reasons for this drop could be the fact that it used posthoc approximations rather than inherently interpretable models; another reason is that there might not be a well-performing sparse rule-based policy. So, the question is: how would we know if it is useful to optimize for an interpretable policy? A simple test that tells the modeler if such a model exists (similar to the characterizations of the Rashomon sets in Challenge \ChRashomon) would be helpful before trying to develop an algorithm to find this well-performing model (if it exists).

    %REMOVE THIS CHALLENGE
%    \item \textbf{Can we build upon existing attention (saliency map) based methods to optimize for interpretability in the policy?}  In order to learn some relations and mimic memory interaction between relevant problem entities, Google Brain proposed an attention mechanism \citep{vaswani2017attention, palm2018recurrent, santoro2017simple}. This mechanism projects query, key, and value vectors to a lower dimensional space and then computes some correspondences between them using dot-products. %For sequential data, such as text, the attention mechanism allows determining the most relevant elements through the entire sequence. %This property makes the attention mechanism a successful substitution of recurrent neural networks in some cases.
%    Attention mechanisms was also used to provide understandability for the agent's policy and actions \citep{zambaldi2018relational, mott2019towards}. However, the relations that the system learns are more like a relevance score that does not have an interpretation as a logical relation. Additionally this relevance score is as good as the input data representation. If the input are images with complicated objects the explanations of the attention-based methods might not be transparent anymore. However, given that the attention allows to focus on the relevance regions, can it be modified to ensure better understandability of the agent's actions, perhaps similar to case-by-case reasoning in Challenge \ChSupDis? 

    \item \textbf{Can we simplify or interpret the state space?} Deep reinforcement learning often operates on complex data (raw data) or huge, if not continuous state spaces. The agent might be required to do a lot of exploration to learn useful policies in this case. Simplifying the state space might not only make the RL system learn more efficiently, but might also help humans to understand the decisions it is making. For example, by applying t-SNE to the last layer activation data of \citet{mnih2013playing}'s deep Q-learning  model, \citet{zahavy2016graying} showed that the network automatically discovers hierarchical structures. Previously, these structures had not been used to simplify the state space.  \citet{zhang2021identifying} improve over batch RL by computing the policy only at specific decision points where clinicians treat patients differently, instead of at every timestep. This resulted in a significantly smaller state space and faster planning.
    
\end{enumerate}

\noindent \textbf{Example:} Consider a machine learning practitioner that is working on planning for assistive robots that help the elderly to perform day-to-day duties. In cases when the robot does some damage during its task execution or chooses a unusual plan to execute, it might be hard to understand the logic behind the robot's choices. If the practitioner can check that an interpretable model exists (Challenge \ChRL.2), optimize for it (Challenge \ChRL.1), and interpret the state space (Challenge \ChRL.3), debugging the robot's failed behavior and correcting it can be much faster than without these techniques. Further, interpretability will help to strengthen trust between the robot and the human its assisting, and help it to generalize to new unforeseen cases. %If we see a future, in which assistive robotics are team players, transparency and interpretability are core principles to strive for.

% \section{Interactive interpretable machine learning and counterfactual explanations}\label{sec:interactive} 

% Use counterfactual explanations to design better ML models. If the counterfactual explanation is bad, how do we use that information to improve the model?

% Counterfactual review: %https://arxiv.org/abs/2010.10596

% counterfactual explanations - they require a cost function and it can be difficult to elicit these. 

% \citep{Schramowski20,Laugel2019}

% Use this reference:
% \citep{SokolFlach20}
% \url{https://link.springer.com/content/pdf/10.1007/s13218-020-00637-y.pdf}

% Interesting discussion of issues with posthoc explanation
% \citep{Laugel2019,Barocas20}

% - Exploration of the Rashomon set

% - Seamless addition of domain constraints

% \citet{ChangEtAl2012} aimed to reverse-engineer a black box scoring model for product rankings and aimed to determine what the most cost-effective change to a product could be made to boost the product into the top 10. They found that the black box model was difficult to reverse-engineer because the product rating scores were rounded before being reported. They concluded that black box rating systems cause disadvantages for both companies that build products and consumers, because companies could not actually conduct effective counterfactual reasoning to improve their products, and consumers are blindfolded the definition of product quality and it may differ from their own priorities (e.g., safety features of a car vs$.$ sleekness of dashboard display).  

\section{Problems that weren't in our top 10 but are really important}\label{sec:additional}

We covered a lot of ground in the 10 challenges above, but we certainly did not cover all of the important topics related to interpretable ML. Here are a few that we left out:

\begin{itemize}

\item \textbf{Can we improve preprocessing to help with both interpretability and test accuracy?} As we discussed, some interpretable modeling problems are computationally difficult (or could tend to overfit) without preprocessing. For supervised problems with tabular data, popular methods like Principal Component Analysis (PCA) would generally transform the data in a way that damages interpretability of the model. This is because each transformed feature is a combination of all of the original features. Perhaps there are other general preprocessing tools that would preserve predictive power (or improve it), yet retain interpretability.

\item \textbf{Can we convey uncertainty clearly?} Uncertainty quantification is always important.  \citet{Tomsett20} discusses the importance of uncertainty quantification and interpretability, and \citet{antoran2020getting} discusses interpretable uncertainty estimation.

\item \textbf{Can we divide the observations into easier cases and harder cases, assigning more interpretable models to the easier cases?}
Not all observations are equally easy to classify. As noted in the work of \citet{TongWang2019,TongWangEtAl2021}, it is possible that easier cases could be handled by simpler models, leaving only the harder cases for models with less interpretability.

\item \textbf{Do we need interpretable neural networks for tabular data?} There have been numerous attempts to design interpretable neural networks for tabular data, imposing sparsity, monotonicity, etc. However, it is unclear whether there is motivation for such formulations, since neural networks do not generally provide much benefit for tabular data over other kinds of models. One nice benefit of the ``optimal'' methods discussed above is that when we train them we know how far we are from a globally optimal solution; the same is not true for neural networks, rendering them not just less interpretable but also harder to implement reliably.

\item \textbf{What are good ways to co-design models with their visualizations?} Work in data and model visualization will be important in conveying information to humans. Ideally, we would like to co-design our model in a way that lends itself to easier visualization. One particularly impressive interactive visualization was done by \citet{Steffen2018} for a model on loan decisions. Another example of an interactive model visualization tool is that of \citet{chen2018interpretable}. These works demonstrate that as long as the model can be visualized effectively, it can be non-sparse but still be interpretable. In fact, these works on loan decisions constitute an example of when a sparse model might not be interpretable, since people might find a loan decision model unacceptable if it does not include many types of information about the applicant's credit history. Visualizations can allow the user to hone in on the important pieces of the model for each prediction.

\item \textbf{Can we do interpretable matching in observational causal inference?} In the classical potential outcomes framework of causal inference, matching algorithms are often used to match treatment units to control units. Treatment effects can be estimated using the matched groups. Matching in causal inference can be considered a form of case-based reasoning (Challenge \ChCaseBased{}). Ideally, units within each matched group would be similar on important variables. One such framework for matching is the Almost Matching Exactly framework  \citep[also called ``learning to match,'' see][]{FLAME}, which learns the important variables using a training set, and prioritizes matching on those important variables. 

Matching opens the door for interpretable machine learning in causal inference because it produces  treatment and control data whose distributions overlap. 
%It can allow us to use other interpretable machine learning tools for observational causal inference and/or policy design. 
Interpretable machine learning algorithms for supervised classification discussed in Challenges \ChLogical{}, \ChScoring{}, and \ChGAM{} can be directly applied to matched data; for instance, a sparse decision tree can be used on matched data to obtain an interpretable model for individual treatment effects, or an interpretable policy (i.e., treatment regimes).

\item \textbf{How should we measure variable importance?} There are several types of variable importance measures: those that measure how important a variable is to a specific prediction, those that measure how important a variable is generally to a model, and those that measure how important a variable is independently of a model. Inherently interpretable models should generally not need the first one, as it should be clear how much a variable contributed to the prediction; for instance, for decision trees and other logical models, scoring systems, GAMs, case-based reasoning models, and disentangled neural networks, the reasoning process that comes with each prediction tells us explicitly how a variable (or concept) contributed to a given prediction. There are many explainable ML tutorials to describe how to estimate the importance of variables for black box models. For the other two types of variable importance, we refer readers to the work of \citet{fisher2018model}, which has relevant references. If we can decide on an ideal variable importance measure, we may be able to create interpretable models with specific preferences on variable importance.

% \item \textbf{Can we determine what level of complexity of model is actually needed before solving that problem?} In Challenge \ChRashomon, we discussed how one could determine whether simpler models exist without finding them. But this approach does not specify exactly what level of complexity might be required for a given model to be within the Rashomon set.

%\item Troubleshooting data (Lesia - look for literature)
\item \textbf{What are the practical challenges in deploying interpretable ML models?} There are many papers that provide useful background about the use of interpretability and explainability in practice. For instance, an interesting perspective is provided by \citet{BhattEtAl} who conduct an explainability survey showing that explanations often used only internally for troubleshooting rather than being shown to users. \citet{Kaur2020,Pour18} (among others) have performed extensive human experiments on interpretable models and posthoc explanations of black box models, with some interesting and sometimes nonintuitive results.

%\item \textbf{Counterfactual reasoning} Our ability to perform counterfactual reasoning is important. Scoring systems and decision trees make it easy to perform counterfactual reasoning, but counterfactual reasoning may not come so naturally with other types of models. This topic is heavily covered in XAI reviews, 

\item \textbf{What type of explanation is required by law?} The question of what explanations are legally required is important and interesting, but we will leave those to legal scholars such as \citet{BibalEtal20,WachterEtAl17}. Interestingly, scholars have explicitly stated that a ``Right to Explanation'' for automated decision-making does not actually exist in the European Union's General Data Protection Regulation, despite the intention \citep{WachterEtAl17}. \citet{Rudin19} proposes that for a high-stakes decision that deeply affects lives, no black box model should be used unless the deployer can prove that no interpretable model exists with a similar accuracy. If enacted, this would likely mean that black box models rarely (if ever) would be deployed for high-stakes decisions.

\item \textbf{What are other forms of interpretable models?} It is not possible to truly review all interpretable ML models. One could argue that most of applied Bayesian statistics fits into our definition of interpretable ML because the models are constrained to be formed through an interpretable generative process. This is indeed a huge field, and numerous topics that we were unable to cover also would have deserved a spot in this survey.

\end{itemize}

%- Adrian Weller
%- Jenn Wortman Vaughn
%- Geoff Hinton - Haiyang

%\begin{figure}[h!]
%\centering
%\includegraphics[scale=1.7]{universe}
%\caption{The Universe}
%\label{fig:universe}
%\end{figure}

\section{Conclusion}
In this survey, we hoped to provide a pathway for readers into important topics in interpretable machine learning. The literature currently being generated on interpretable and explainable AI can be downright confusing. 
%We cannot imagine that there is a more confusing literature right now than that of interpretable machine learning. 
The sheer diversity of individuals weighing in on this field includes not just statisticians and computer scientists but legal experts, philosophers, and graduate students, many of whom have not either built or deployed a machine learning model ever. It is easy to underestimate how difficult it is to convince someone to use a machine learning model in practice, and interpretability is a key factor.
Many works over the last few years have contributed new terminology, mistakenly subsumed the older field of interpretable machine learning into the new field of ``XAI,'' and review papers have universally failed even to truly  distinguish between the basic concepts of explaining a black box and designing an interpretable model. Because of the misleading terminology, where papers titled ``explainability'' are sometimes about ``interpretability'' and vice versa, it is very difficult to follow the literature (even for us). At the very least, we hoped to introduce some fundamental principles, and
cover several important areas of the field and show how they relate to each other and to real problems.
Clearly this is a massive field that we cannot truly hope to cover, but we hope that the diverse areas we covered and problems we posed might be useful to those needing an entrance point into this maze. 

Interpretable models are not just important for society, they are also beautiful. One might also find it absolutely magical that simple-yet-accurate models exist for so many real-world datasets. We hope this document allows you to see not only the importance of this topic but also the elegance of its mathematics and the beauty of its models.

\section*{Acknowledgments}
We thank Leonardo Lucio Custode for pointing out several useful references to Challenge \ChRL{}. Thank you to David Page for providing useful references on early explainable ML. Thank you to the anonymous reviewers that made extremely helpful comments.

\bibliography{references}
\end{document}